\documentclass[11pt]{article}
\usepackage[T1]{fontenc}
\usepackage[utf8]{inputenc}
\usepackage{lmodern}
\usepackage[margin=1in]{geometry} 
\usepackage{amsmath,amssymb,amsthm,mathtools}
\usepackage{microtype}
\usepackage{array,longtable}
\usepackage[authoryear,round]{natbib}
\usepackage{xcolor}
\usepackage{xurl}
\usepackage[colorlinks=true,allcolors=blue]{hyperref}

\allowdisplaybreaks[2]
\newtheorem{theorem}{Theorem}
\newtheorem{proposition}{Proposition}
\newtheorem{lemma}{Lemma}
\newtheorem{corollary}{Corollary}
\theoremstyle{definition}

\newtheorem{question}{Question}
\theoremstyle{remark}
\newtheorem{remark}{Remark}
\numberwithin{equation}{section}
\DeclareMathOperator{\Var}{Var}

\title{Algorithmic Stability I: Concentration in Hilbert Space}
\author{Thanh Nguyen-Cung\thanks{VinUniversity. Emails:
\href{mailto:24thanh.nc@vinuni.edu.vn}{24thanh.nc@vinuni.edu.vn},
\href{mailto:binh.nt2@vinuni.edu.vn}{binh.nt2@vinuni.edu.vn}}
\and Binh T. Nguyen\footnotemark[1]}
\date{}

\begin{document}
\maketitle

\begin{abstract}
We study concentration for sums of $n$ dependent random variables taking values in a separable Hilbert space and constructed from independent observations. Under two conditional centering conditions, we show that the $L_p$ norm of their sum is bounded by $15p\|\Omega\|_F$ for every $p\ge2$, where $\Omega$ is the matrix of coordinate sensitivities. We establish lower bounds matching this estimate up to universal constants for every sensitivity matrix at $p=2$ and for uniform sensitivities throughout $2\le p\le n$. For deterministic uniformly stable algorithms with bounded losses, we establish matching moment and tail bounds. We derive generalization bounds for losses and gradients that remove the factor $\log n$ from the stability term of previous bounds and replace population second moments by variances in the sampling term. For randomized algorithms under mean square uniform stability, the same improvements hold for the conditional root mean square of the generalization gap, with high probability over the training sample. For Lipschitz losses, we remove the factor $\log n$ from existing excess-risk bounds for approximate strongly convex empirical minimization and extend these bounds to empirical quadratic growth relative to the minimizer set, without assuming stability of the selected minimizers. Separately, under empirical Polyak--{\L}ojasiewicz conditions, we establish bounds for all smooth empirical minimizers without requiring population strong convexity, population gradient domination, or stability of the selected minimizers. These bounds yield an excess-risk rate of $O(1/n^2)$ at fixed confidence in the low-noise regime.
\end{abstract}

\begingroup
\small
\setcounter{tocdepth}{3}
\tableofcontents
\endgroup
\clearpage
\section{Introduction}\label{sec:introduction}
A central question in statistical learning is how well a model trained on a finite sample performs on unseen data. Although the empirical risk measures the performance of a learned model on the training sample, it does not necessarily reflect its population risk. A key difficulty is that the model is selected using the same observations on which its empirical risk is evaluated. Understanding how this data dependence affects generalization is therefore essential.

Let $S=(Z_1,\ldots,Z_n)$ consist of $n$ independent observations drawn from an unknown distribution $\mathrm{P}$ on a measurable space $\mathcal Z$. A learning algorithm $A$ returns a decision $A(S)$ in a hypothesis set $\mathcal W$, whose error on an observation $z$ is measured by a loss $\ell:\mathcal W\times\mathcal Z\to[0,\infty)$. Let $Z\sim\mathrm{P}$ be independent of $S$. The population and empirical risks are defined as
\[
R(w)=\mathbb E_Z\ell(w,Z),\qquad R_{\text{emp}}(w)=\frac{1}{n}\sum_{i=1}^n\ell(w,Z_i).
\]
Our object of interest is the generalization gap $R(A(S))-R_{\text{emp}}(A(S))$ which measures the difference between the expected loss on new observations and the average training loss. If $w$ is fixed independently of the sample, then $\ell(w,Z_1),\ldots,\ell(w,Z_n)$ are independent random variables, and standard concentration inequalities can be applied under suitable tail assumptions. 

However, this argument does not directly extend to $A(S)$, since the learned decision depends on the same observations used to compute its empirical risk. Algorithmic stability offers a way to address this difficulty by controlling the sensitivity of the learned loss to changes in the training sample. More broadly, it has been used to analyze nearest-neighbor methods \citep{devroye1979}, regularized learning algorithms \citep{bousquet2002}, and stochastic gradient descent \citep{hardt2016}. Its connections to cross-validation, learnability, and differential privacy are also discussed in \citet[Section~1.4]{feldman2019} and the references therein. To make the notion of stability precise, we recall the definition of uniform stability from \citet{bousquet2002}. A learning algorithm $A$ is said to be uniformly $\gamma$-stable for some $\gamma\ge0$ if, for every pair of samples $s,\tilde s$ differing in exactly one observation,
\[
\sup_{z\in\mathcal Z}|\ell(A(s),z)-\ell(A(\tilde s),z)|\le\gamma.
\]
Uniform stability then guarantees that $\left| \mathbb E_S\left[ R(A(S))-R_{\mathrm{emp}}(A(S)) \right] \right| \leq \gamma$. However, this bound only controls the generalization gap in expectation and does not guarantee a small generalization gap for single training samples. Establishing high-probability bounds therefore requires a concentration argument that accounts for the dependence of the learned decision on the training data. Similar difficulties arise in gradient generalization, where the goal is to control the discrepancy between the population and empirical gradients at the learned decision. In this paper, we study how stability can be used to establish concentration bounds for both losses and gradients. 

Throughout Sections~\ref{sec:introduction} and~\ref{sec:main-results}, we take $0<\delta\le1/e$ as the failure probability. Unless stated otherwise, high-probability bounds hold with probability at least $1-\delta$ over $S$. For nonnegative quantities $a$ and $b$, we write $a\lesssim b$ if $a\le Cb$ for a universal constant $C$, $a\gtrsim b$ if $b\lesssim a$, and $a\asymp b$ if both inequalities hold.

\subsection{High-probability generalization under uniform stability}

Suppose that $0\le\ell\le M$. For each sample $s$, let $f_s(z)=\ell(A(s),z)$ and define $\bar f(z)=\mathbb E_{S'}f_{S'}(z)$, where $S'\sim\mathrm P^n$ is independent of $S$. The generalization gap then admits the decomposition
\[
\begin{array}{@{}r@{\,}c@{\,}c@{\,}c@{\,}c@{}}
\displaystyle R(A(S))-R_{\mathrm{emp}}(A(S))
&=&\displaystyle\left(\mathbb E_Z(f_S-\bar f)(Z)-\frac1n\sum_{i=1}^n(f_S-\bar f)(Z_i)\right)
&+&\displaystyle\left(\mathbb E_Z\bar f(Z)-\frac1n\sum_{i=1}^n\bar f(Z_i)\right)\\[0.7ex]
&=&\text{stability error}&+&\text{sampling error}.
\end{array}
\]
The first term captures the dependence of the learned loss on the training sample, which we control using stability. The second is the sampling error of $\bar f$, which we bound using standard concentration inequalities. The bounded differences argument of \citet{bousquet2002} gives
\[
|R(A(S))-R_{\text{emp}}(A(S))|\lesssim \underbrace{\gamma\sqrt{n\log(1/\delta)}}_{\text{stability error}}+\underbrace{M\sqrt{\frac{\log(1/\delta)}{n}}}_{\text{sampling error}}.
\]
At fixed confidence, the stability term is larger by a factor of order $\sqrt n$ than the bound in expectation. In particular, when $\gamma\asymp1/\sqrt n$, the expectation bound vanishes, whereas the stability term above remains of constant order. This limitation is particularly relevant to iterative algorithms, whose stability can deteriorate as the number of iterations increases~\citep{hardt2016}. 

Reducing the extra factor of $\sqrt n$ in the classical stability term motivated several improvements. \citet{feldman2018} reduced the stability term to order $\sqrt{M\gamma\log(1/\delta)}$. They also established a bound on the second moment of order $\gamma^2+M^2/n$, suggesting that the factor $\sqrt n$ in the classical stability term could be avoided. Subsequently, \citet{feldman2019} obtained a high-probability bound with a stability term of order $\gamma\log n\log(n/\delta)$. Building on these advances, \citet{bkz2020} further improved the dependence on the confidence level, showing that
\begin{equation}\label{eq:intro-bkz}
|R(A(S))-R_{\text{emp}}(A(S))|\lesssim\gamma\log n\log(1/\delta)+M\sqrt{\frac{\log(1/\delta)}{n}}.
\end{equation}
The stability term in~\eqref{eq:intro-bkz} still contains a factor of $\log n$, leading \citet[Section~6]{bkz2020} to ask whether it can be removed.

\begin{question}\label{q:generalization}
For a uniformly $\gamma$-stable algorithm with loss $0\le\ell\le M$, does the generalization gap satisfy
\[
|R(A(S))-R_{\text{emp}}(A(S))|\lesssim\gamma\log(1/\delta)+M\sqrt{\frac{\log(1/\delta)}{n}} \, ?
\]
\end{question}

We show that the proposed bound holds, removing the extra $\log n$ factor in \eqref{eq:intro-bkz}. Moreover, we show that the resulting bound is optimal in the worst case. For every $n\ge1$, $M>0$, $\gamma\ge0$, and $p\ge1$, the largest $L_p$ generalization gap among deterministic uniformly $\gamma$-stable learning algorithms with losses in $[0,M]$ is of order $\min\{M,\gamma p+M\sqrt{p/n}\}$. Matching upper and lower tail bounds hold at probability scale $e^{-p}$, with a learning problem attaining the lower bounds for all $p$ (Proposition~\ref{prop:learning-lower} and Corollary~\ref{cor:learning-sharp-order}).

The finite-moment framework of \citet[Theorem~3.4]{lei2026} permits random replacement envelopes with finite $L_p$ moments and gives a sum of exponential and polynomial tail bounds. It covers unbounded losses, whereas Question~\ref{q:generalization} concerns bounded losses under deterministic uniform stability.

Although this bound is optimal under uniform stability, its sampling term remains of order $1/\sqrt n$ at fixed confidence. For excess risk, the rates can be improved under additional assumptions. In particular, \citet{shalevshwartz2009} established an expected excess-risk bound of order $1/n$ for strongly convex empirical risk minimization. This raises a natural question: can we achieve the same rate with high probability? 

\subsection{High-probability excess risk bounds} 
Assume that a population minimizer $w^\star\in\arg\min_{w\in\mathcal W}R(w)$ exists, and write $R^\star=R(w^\star)$. The excess risk of $A(S)$ is $R(A(S))-R^\star$. Empirical risk minimization returns $\widehat w_S\in\arg\min_{w\in\mathcal W}R_{\text{emp}}(w)$. For a general algorithm, define its empirical optimization error by 
\[\Delta_{\mathrm{opt}}(S)=R_{\text{emp}}(A(S))-\inf_{w\in\mathcal W}R_{\text{emp}}(w).
\]
With this notation, we can express the excess risk as
\[
\begin{array}{@{}r@{\,}c@{\,}c@{\,}c@{\,}c@{}}
\displaystyle R(A(S))-R^\star
&=&\displaystyle\Delta_{\mathrm{opt}}(S)
&+&\displaystyle\left(R(A(S))-R_{\text{emp}}(A(S))
+\inf_{w\in\mathcal W}R_{\text{emp}}(w)-R^\star\right)\\[0.7ex]
&=&\text{optimization error}&+&\text{statistical error}.
\end{array}
\]
For an empirical minimizer, the optimization error is zero. The statistical error accounts for the difference between empirical and population risks and can be positive or negative for a general learning algorithm.

The question of achieving a high-probability rate of $1/n$ had already arisen in the study of stochastic convex optimization. For losses that are $\lambda$-strongly convex and $L_\ell$-Lipschitz on a closed convex subset of a Hilbert space, \citet{shalevshwartz2009} used stability to establish $\mathbb E[R(\widehat w_S)-R^\star]\lesssim L_\ell^2/(\lambda n)$. Applying Markov's inequality gives $R(\widehat w_S)-R^\star\lesssim L_\ell^2/(\lambda n\delta)$ with probability at least $1-\delta$. While the rate $1/n$ is preserved, the confidence dependence is $1/\delta$ rather than $\log(1/\delta)$. In Appendix~A, they noted that logarithmic confidence dependence could be obtained under an additional smoothness assumption on the population risk, but with a worse dependence on the strong convexity parameter. This led to the following question.

\begin{question}\label{q:erm}
Can the empirical minimizer $\widehat w_S$ satisfy
\[
R(\widehat w_S)-R^\star\lesssim\frac{L_\ell^2}{\lambda n}\log(1/\delta)
\]
without a smoothness assumption or dependence on the dimension?
\end{question}

\citet{klochkov2021} combined stability with a generalized Bernstein condition and obtained $R(\widehat w_S)-R^\star\lesssim L_\ell^2\log n\log(1/\delta)/(\lambda n)$ for strongly convex Lipschitz losses.
Their result answered Question~\ref{q:erm} up to a factor $\log n$. The generalized Bernstein condition allows the sampling error to decrease with the excess risk, leading to the faster rate $1/n$. Our goal is to remove the remaining $\log n$ factor while preserving this faster rate. 

\subsection{Gradient generalization and optimization} 
For nonconvex problems, an optimization algorithm may find a point with a small empirical gradient without minimizing the empirical risk~\citep{ghadimi2013}. However, a small empirical gradient does not necessarily imply a small population gradient. Suppose $\mathcal W\subseteq\mathbb R^d$, the loss is differentiable, and differentiation commutes with expectation. At the learned decision,
\[
\begin{array}{@{}r@{\,}c@{\,}c@{\,}c@{\,}c@{}}
\displaystyle \nabla R(A(S))
&=&\displaystyle\nabla R_{\text{emp}}(A(S))
&+&\displaystyle\left(\nabla R(A(S))-\nabla R_{\text{emp}}(A(S))\right)\\[0.7ex]
&=&\text{empirical gradient}&+&\text{gradient generalization error}.
\end{array}
\]
Taking Euclidean norms and applying the triangle inequality gives
\[
\|\nabla R(A(S))\|_2
\le\|\nabla R_{\text{emp}}(A(S))\|_2
+\|\nabla R(A(S))-\nabla R_{\text{emp}}(A(S))\|_2.
\]
Here $\|\cdot\|_2$ denotes the Euclidean norm on $\mathbb R^d$.

\citet{foster2018} studied this transfer through uniform convergence of gradients. A complementary approach, developed by \citet{lei2023}, measures stability of the gradients themselves. \citet{fanlei2024} established high-probability bounds for gradient generalization using concentration in Hilbert spaces, allowing different sensitivities across training observations. If replacing the $j$th training observation changes the gradient by at most $\gamma_j$, their stability term depends on $\gamma_{\mathrm{rms}}=\left(\frac{1}{n}\sum_{j=1}^n\gamma_j^2\right)^{1/2}$ rather than $\max_j\gamma_j$. In Section~7, they explicitly asked whether the factor $\log n$ multiplying this quantity can be removed. \citet{zhu2025} subsequently retained the population second moment of the gradient. When $\|\nabla\ell(w,z)\|_2\le L_\ell$ and $\gamma_j=\gamma$ for every $j$, their bound for a deterministic algorithm has the form
\[
\|\nabla R(A(S))-\nabla R_{\text{emp}}(A(S))\|_2\lesssim\gamma\log n\log(1/\delta)+\sqrt{\frac{\log(1/\delta)}{n}\mathbb E_Z\|\nabla\ell(A(S),Z)\|_2^2}+\frac{L_\ell\log(1/\delta)}{n}.
\]
The population second moment in this bound can be much smaller than $L_\ell^2$, leading to a tighter sampling term. However, the stability term still contains the factor $\log n$ that \citet{fanlei2024} asked to remove. We therefore seek a bound with stability term of order $\gamma_{\mathrm{rms}}\log(1/\delta)$ for unequal sensitivities $\gamma_1,\ldots,\gamma_n$, with a sampling term depending on the population gradient variance and constants independent of the dimension.

A further question arises when the learning algorithm is randomized, since its output depends not only on the training sample but also on its internal randomness. Stability for randomized algorithms was studied by \citet{elisseeff2005}. Let $\xi\sim\mathrm Q$ be a random seed independent of $S$, and write $A(S,\xi)$ for the output. Comparing neighboring samples using the same seed allows us to study sensitivity to the training data separately from the algorithm's randomness. \citet[Section~1.1]{yuanli2023} asked whether exponential generalization bounds could be obtained under assumptions weaker than stability for each seed. They used stability in mean square over the seed to bound the expected absolute generalization gap. To obtain a high-probability guarantee over both sources of randomness, they used an additional algorithm based on bagging. We therefore ask whether bounded gradients and mean square gradient stability suffice to control
\[
\left(\mathbb E_\xi\|\nabla R(A(S,\xi))-\nabla R_{\mathrm{emp}}(A(S,\xi))\|_2^2\right)^{1/2}
\]
with high probability over $S$, with a stability term of order $\gamma\log(1/\delta)$ and a sampling term depending on the population gradient variance, without requiring stability for every seed or modifying the original algorithm.

\subsection{Reduction to a concentration problem}
The problems discussed above concern different statistical quantities, but they share a common challenge: the training sample is used both to construct the predictor and to evaluate it. To study these problems in a common framework, let $\mathcal H$ be a real separable Hilbert space with norm $\|\cdot\|_{\mathcal H}$, and let $F:\mathcal Z^n\times\mathcal Z\to\mathcal H$ be jointly measurable. The choices $F(S,z)=\ell(A(S),z)$, $F(S,z)=\ell(A(S),z)-\ell(w^\star,z)$, and $F(S,z)=\nabla\ell(A(S),z)$ correspond to losses, excess losses, and gradients, respectively. In each case, define the corresponding generalization gap by 
\[
\Delta(S)=\mathbb E_ZF(S,Z)-\frac{1}{n}\sum_{i=1}^nF(S,Z_i).
\]
Assume $\|F(s,z)\|_{\mathcal H}\le M$ for every $s\in\mathcal Z^n$ and $z\in\mathcal Z$, and $\|F(s,z)-F(\tilde s,z)\|_{\mathcal H}\le\gamma$ for all neighboring samples $s,\tilde s\in\mathcal Z^n$ and every $z\in\mathcal Z$. For a strongly measurable random variable $X$ taking values in $\mathcal H$, write $\|X\|_{L_p(\mathcal H)}=(\mathbb E\|X\|_{\mathcal H}^p)^{1/p}$ for every $p\in[1,\infty)$.

For a square-integrable Hilbert-valued random variable $f(Z)$, write $\Var_Z(f(Z))=\mathbb E_Z\|f(Z)-\mathbb E_Z f(Z)\|_{\mathcal H}^2$. Equivalently, $\Var_Z(f(Z))=\mathbb E_Z\|f(Z)\|_{\mathcal H}^2-\|\mathbb E_Z f(Z)\|_{\mathcal H}^2$. When $f$ depends on the sample or a random seed, the variance is taken over $Z$ with these arguments fixed.

We use the reduction of \citet[Lemma~7]{bkz2020} to study $\Delta(S)$ through a sum of conditionally centered functions. For each $i\in[n]$, write $S_{-i}$ for the observations other than $Z_i$, and let $S^{(i)}$ be obtained by replacing $Z_i$ with an independent copy $Z_i'$. Define
\[
g_i(S)=\mathbb E_{Z_i'}\left[\mathbb E_ZF(S^{(i)},Z)-F(S^{(i)},Z_i)\right].
\]
The same argument applies to Hilbert-valued functions: $\mathbb E[g_i(S)\mid S_{-i}]=0$, and changing coordinate $j\ne i$ changes $g_i(S)$ by at most $2\gamma$. The assumed sensitivity bound also gives
\[
\left\|n\Delta(S)-\sum_{i=1}^n g_i(S)\right\|_{\mathcal H}\le2n\gamma.
\]
We provide the Hilbert-valued details in Appendix~\ref{app:statistics}.

To separate the independent and dependent components of this sum, define $k_i(Z_i)=\mathbb E[g_i(S)\mid Z_i]$ and $h_i(S)=g_i(S)-k_i(Z_i)$. Then
\[
\sum_{i=1}^n g_i(S)=\underbrace{\sum_{i=1}^n h_i(S)}_{\text{dependent component}}+\underbrace{\sum_{i=1}^n k_i(Z_i)}_{\text{independent component}}.
\]
The variables $k_i(Z_i)$ are independent and centered, whereas the random variables $h_i(S)$ satisfy both $\mathbb E[h_i(S)\mid Z_i]=0$ and $\mathbb E[h_i(S)\mid S_{-i}]=0$. We refer to these two conditions as double centering. 
The functions $h_i$ retain the sensitivity bound $2\gamma$ for changes in coordinate $j\ne i$, since $k_i(Z_i)$ depends only on $Z_i$.

In the scalar case, \citet[Theorem~4]{bkz2020} obtained an $L_p$ bound of order $pn\gamma\log n+M\sqrt{pn}$ for $\sum_{i=1}^n g_i(S)$, with $p\ge2$. They asked whether the $\log n$ factor could be removed for functions satisfying the same centering and sensitivity conditions \citep[Section~6]{bkz2020}. We study this question in Hilbert spaces with unequal coordinate sensitivities, covering gradients and randomized algorithms.

\begin{question}\label{q:moment}
Let $Z_1,\ldots,Z_n$ be independent, and let $g_i(S)$ be Bochner-integrable random variables taking values in $\mathcal H$. Suppose that, for every $i\in[n]$, both $\mathbb E[g_i(S)\mid S_{-i}]=0$ and $\|\mathbb E[g_i(S)\mid Z_i]\|_{\mathcal H}\le M$ hold almost surely. Assume also that changing coordinate $j$ changes $g_i$ by at most $\gamma$ in Hilbert norm for every pair of distinct indices $i,j\in[n]$. Under these assumptions, does the inequality
\[
\left\|\sum_{i=1}^n g_i(S)\right\|_{L_p(\mathcal H)}\lesssim pn\gamma+M\sqrt{pn}
\]
hold for every $p\ge2$, with a constant independent of $n$ and the dimension of $\mathcal H$?
\end{question}

\section{Our contribution}\label{sec:main-results}

\subsection{Main theorem}\label{sec:main}
For the concentration theorem, the coordinates need not have the same distribution. Let $n\ge1$ and write $[n]=\{1,\ldots,n\}$. For each $i\in[n]$, let $(\mathcal Z_i,\mathcal A_i)$ be a measurable space and let $Z_i$ take values in $\mathcal Z_i$. Assume that $Z_1,\ldots,Z_n$ are independent, and equip $\mathcal X=\prod_{i=1}^n\mathcal Z_i$ with the product $\sigma$-algebra. Write $S=(Z_1,\ldots,Z_n)$ and $S_{-i}=(Z_j)_{j\in[n]\setminus\{i\}}$. For a fixed point, write $s=(z_1,\ldots,z_n)\in\mathcal X$ and $s_{-i}=(z_j)_{j\in[n]\setminus\{i\}}$. Let $\mathcal H$ be a real separable Hilbert space. For a matrix $\Omega=(\gamma_{ij})\in\mathbb R^{n\times n}$ with nonnegative entries and zero diagonal, its Frobenius norm is $\|\Omega\|_F=\left(\sum_{\substack{i,j=1\\i\ne j}}^n\gamma_{ij}^2\right)^{1/2}$.

\begin{theorem}\label{thm:main}
Let $h_i:\mathcal X\to\mathcal H$, $i\in[n]$, be measurable, with $h_i(S)$ Bochner-integrable. Suppose that, for all distinct indices $i,j\in[n]$ and every pair $s,\tilde s\in\mathcal X$ differing only in coordinate $j$, we have $\|h_i(s)-h_i(\tilde s)\|_{\mathcal H}\le\gamma_{ij}$. Assume also that, for every $i\in[n]$, the following identities hold almost surely:
\begin{equation}\label{eq:double-centering} \mathbb E[h_i(S)\mid Z_i]=0,\qquad\mathbb E[h_i(S)\mid S_{-i}]=0. \end{equation}
Then, for every $p\ge2$,
\begin{equation}\label{eq:pointwise-moment} \left\|\sum_{i=1}^n h_i(S)\right\|_{L_p(\mathcal H)}\le15p\|\Omega\|_F. \end{equation}
\end{theorem}

By Markov's inequality, for every $\delta\in(0,1/e]$, the norm of the sum is at most a universal constant times $\|\Omega\|_F\log(1/\delta)$ with probability at least $1-\delta$.

When $\gamma_{ij}=\gamma$ for every pair of distinct indices $i,j\in[n]$, we have $\|\Omega\|_F=\sqrt{n(n-1)}\,\gamma\le n\gamma$, so the theorem gives, for every $p\ge2$,
\begin{equation}\label{eq:core-overview} \left\|\sum_{i=1}^n h_i(S)\right\|_{L_p(\mathcal H)}\le15pn\gamma. \end{equation}
For the full sum, suppose that, for every $i\in[n]$, both $\mathbb E[g_i(S)\mid S_{-i}]=0$ and $\|\mathbb E[g_i(S)\mid Z_i]\|_{\mathcal H}\le M$ hold almost surely. Assume also that changing coordinate $j$ changes $g_i$ by at most $\gamma$ for every pair of distinct indices $i,j\in[n]$. Corollary~\ref{cor:uniform-moment} then gives, for every $p\ge2$,
\[
\left\|\sum_{i=1}^n g_i(S)\right\|_{L_p(\mathcal H)}\le\underbrace{15pn\gamma}_{\text{dependent component}}+\underbrace{2M\sqrt{pn}}_{\text{independent component}}.
\]
At $p=2$, the estimate following Corollary~\ref{cor:general-moment} uses orthogonality of the two components to bound the squared norm of their sum by $\|\Omega\|_F^2+\sum_{i=1}^n\mathbb E\|\mathbb E[g_i(S)\mid Z_i]\|_{\mathcal H}^2$. Thus Theorem~\ref{thm:main} and Corollary~\ref{cor:uniform-moment} answer Question~\ref{q:moment}, removing $\log n$ from the dependent component while preserving the independent term. More generally, if $\gamma_{ij}=\gamma_j$ for every pair of distinct indices $i,j\in[n]$, the theorem uses $\sqrt{n-1}\left(\sum_{j=1}^n\gamma_j^2\right)^{1/2}$ instead of $n\max_j\gamma_j$. Our bound accounts for individual coordinate sensitivities rather than bounding them all by the largest one.

The dependence on $\|\Omega\|_F$ is optimal at $p=2$ for every prescribed sensitivity matrix, as shown by Proposition~\ref{prop:sharper-L2} and Proposition~\ref{prop:matrix-lower}. For uniform sensitivities, Corollary~\ref{cor:sharp-order} establishes that the worst-case $L_p$ norm is of order $\gamma n\min\{p,n\}$. Thus, the linear dependence on $p$ in \eqref{eq:core-overview} is necessary throughout $2\le p\le n$, while the moment remains of order $n^2\gamma$ for $p\ge n$. This dependence on $p$ need not be sharp for sparse sensitivity matrices. Corollaries~\ref{cor:tail-lower} and~\ref{cor:independent-lower} further establish a tail lower bound for the dependent sum and a moment lower bound for the full sum, respectively.

To show that the same rates can be attained by learning algorithms, Proposition~\ref{prop:learning-lower} constructs a deterministic uniformly stable learning algorithm with bounded loss whose generalization gap has a matching lower tail at probability scale $e^{-p}$ for every $p\ge1$. The same learning problem works for all $p$, and Corollary~\ref{cor:learning-sharp-order} establishes matching worst-case moment and tail bounds.

\subsection{Statistical consequences}
We return to the learning setting $S\sim\mathrm{P}^n$. The same concentration bound gives guarantees for losses, excess losses, and gradients. Faster rates follow when the population variance is small or can be controlled by the quantity being estimated.

\paragraph{Generalization with a population variance.}
For the function $F$ and generalization gap $\Delta$ defined above, Corollary~\ref{cor:statistic} gives
\begin{equation}\label{eq:intro-statistic} \|\Delta(S)\|_{\mathcal H}\lesssim\underbrace{\gamma\log(1/\delta)}_{\text{stability error}}+\underbrace{\sqrt{\frac{\log(1/\delta)}{n}\Var_Z(F(S,Z))}+\frac{M\log(1/\delta)}{n}}_{\text{sampling error}}. \end{equation}
For bounded losses, Corollary~\ref{cor:uniform} answers Question~\ref{q:generalization}. Compared with \eqref{eq:intro-bkz}, the stability error has no factor $\log n$, and the sampling error depends on the population variance of the fitted function rather than its second moment or uniform bound. For a nonnegative loss bounded by $M$, this variance is at most $R(A(S))[M-R(A(S))]$. The estimate therefore improves when the population risk is small; if both $\gamma$ and $R(A(S))$ are of order $1/n$, the whole bound is of order $1/n$ at fixed confidence. Corollary~\ref{cor:statistic} allows coordinate sensitivities $\gamma_j$ and uses $\gamma_{\mathrm{rms}}$ in place of $\gamma$.

\paragraph{Randomization and conditional root mean squares.}
For gradients, the corresponding stability condition is
\[
\sup_{z\in\mathcal Z}\mathbb E_\xi\|\nabla\ell(A(s,\xi),z)-\nabla\ell(A(\tilde s,\xi),z)\|_2^2\le\gamma^2
\]
for all neighboring samples $s,\tilde s$. The gradient theorem of \citet{zhu2025} assumes this condition but bounds the expected norm over the seed. Such a bound on the first moment does not, in general, imply a bound of the same order for the conditional root mean square. This control of the second moment is required to derive excess risk bounds through gradient domination. Their proof of Lemma~13 in Appendix~C.3 also treats the conditional second moment; our improvement concerns the stability factor and the sampling variance.

Let $\Upsilon(S,z,\xi)$ take values in a real separable Hilbert space $\mathcal K$. Regard $F(S,z)=\Upsilon(S,z,\cdot)$ as an element of $L_2(\mathrm{Q};\mathcal K)$, with norm $(\mathbb E_\xi\|\Upsilon(S,z,\xi)\|_{\mathcal K}^2)^{1/2}$. Under the measurability, envelope, and mean square stability conditions in Section~\ref{subsec:randomization}, \eqref{eq:intro-statistic} controls the conditional root mean square of the generalization gap. The population variance becomes $\mathbb E_\xi\Var_Z(\Upsilon(S,Z,\xi))$. The probability is over $S$; stability need not hold separately for every seed. The resulting bound, stated in \eqref{eq:random-statistic}, provides the randomized gradient guarantee described above when $\Upsilon(S,z,\xi)=\nabla\ell(A(S,\xi),z)$. It also treats randomized loss estimates and the squared quantities used in excess risk analysis.

\paragraph{Excess risk under a generalized Bernstein condition.}
To obtain faster excess risk rates, we use a centered generalized Bernstein condition that controls the variance of the excess loss by the excess risk. Related conditions are used to obtain fast rates through local complexity~\citep{bartlett2005} and smoothness with small optimal risk~\citep{srebro2010}. Let $\mathcal W^\star=\arg\min_{w\in\mathcal W}R(w)$ be nonempty. For some $B\ge0$, assume that each $w\in\mathcal W$ admits a minimizer $u\in\mathcal W^\star$ such that
\[
\Var_Z(\ell(w,Z)-\ell(u,Z))\le B[R(w)-R^\star].
\]

Under this condition, Corollary~\ref{cor:generalized-bernstein-risk} shows that a deterministic uniformly $\gamma$-stable algorithm with $0\le\ell\le M$ satisfies
\[
R(A(S))-R^\star\lesssim
\underbrace{\Delta_{\mathrm{opt}}(S)+\mathbb E_S\Delta_{\mathrm{opt}}(S)}_{\text{optimization error}}
+\underbrace{\left(\gamma+\frac{M+B}{n}\right)\log(1/\delta)}_{\text{statistical error}}.
\]
For exact empirical risk minimization, both optimization errors vanish, and the resulting bound removes the factor $\log n$ from the stability term in \citet{klochkov2021}. Under the same assumptions, we also give a bound using the realized optimization error, with a larger coefficient on that error. Corollary~\ref{cor:approximate} instead uses Lipschitz continuity and empirical quadratic growth relative to the minimizer set. For strongly convex Lipschitz losses, it gives $R(\widehat w_S)-R^\star\lesssim L_\ell^2\log(1/\delta)/(\lambda n)$, answering Question~\ref{q:erm} without a smoothness assumption or dependence on the dimension. More generally, it allows multiple empirical minimizers, requires no population quadratic growth, and accounts for the realized optimization error.

\paragraph{Population gradients and excess risk.}
For gradients, \eqref{eq:gradient-refined} provides the gradient generalization bound sought above, with stability term $\gamma_{\mathrm{rms}}\log(1/\delta)$ and a sampling term depending on the population gradient variance. It also controls the conditional root mean square for randomized algorithms. Under the strong growth condition, \eqref{eq:strong-growth} and~\eqref{eq:strong-growth-mean} remove the factor $\log n$ from the stability term in bounds on population gradient norms.

For excess risk, consider a common gradient stability parameter $\gamma$ and assume that, for some $v_0,v_1\ge0$ and $\mu>0$, the bounds $\Var_Z(\nabla\ell(w,Z))\le v_0+v_1[R(w)-R^\star]$ and $R(w)-R^\star\le\|\nabla R(w)\|_2^2/(2\mu)$ hold at every algorithm output $w$. In the unconstrained deterministic case, Corollary~\ref{cor:risk} gives
\[
R(A(S))-R^\star\lesssim\frac{\|\nabla R_{\text{emp}}(A(S))\|_2^2}{\mu}+\frac{v_0\log(1/\delta)}{\mu n}+\frac{L_\ell^2\log^2(1/\delta)}{\mu n^2}+\frac{\gamma^2\log^2(1/\delta)}{\mu},
\]
provided $n\gtrsim(v_1/\mu)\log(1/\delta)$. The full statement includes constraints through a stationarity residual and allows randomization. The stability term has no factor $(\log n)^2$. At fixed confidence and fixed structural constants, the bound has order $1/n^2$ when $v_0=O(1/n)$, $\gamma=O(1/n)$, and the squared empirical stationarity residual is $O(1/n^2)$. Without a restriction on the noise, the rate $1/n^2$ cannot hold uniformly, even for smooth strongly convex losses and exact empirical minimization (Appendix~\ref{app:fixed-noise}).

\paragraph{Empirical minimization and gradient methods.}
Concentration bounds for stable loss and gradient functions do not by themselves establish stability of a learning algorithm or control its empirical optimization error. \citet[Theorems~3, 5, 7 and~8]{charles2018} study stability of algorithms that converge to global minimizers under the empirical Polyak--\L{}ojasiewicz (PL) or quadratic growth condition. Their uniform stability bounds further require compatibility of the selected empirical minimizers with projections onto neighboring minimizer sets (Assumption~1). In Section~\ref{sec:algorithms}, we use empirical quadratic growth to control mixed differences of the empirical optimal value and apply concentration without assuming stability of a selected minimizer. For smooth empirical minimization, we compare population risks across empirical minimizers under empirical PL.

In the strongly convex Lipschitz case, an optimization error $\epsilon_{\mathrm{opt}}$ contributes at most $L_\ell\sqrt{2\epsilon_{\mathrm{opt}}/\lambda}$, without a confidence factor. A weighted projected subgradient method consequently has excess risk of order $(L_\ell^2/\lambda)\left(\log(1/\delta)/n+1/\sqrt T\right)$ after $T$ steps. Under the empirical PL and smoothness assumptions stated in Section~\ref{subsec:smooth-erm}, the smooth empirical minimization bounds hold for every empirical minimizer, without assuming stability of a selected minimizer. The unconstrained gradient descent bound uses empirical PL to obtain geometric decay of the optimization term without assuming contraction of the update.

\paragraph{Summary of application results and comparison with prior work.}
Appendix~\ref{app:comparison-stability} summarizes the generalization bounds for Hilbert-valued statistics, randomized algorithms, and bounded losses, including worst-case moment bounds and excess risk under a generalized Bernstein condition.
Appendix~\ref{app:comparison-gradient} presents the bounds for gradient generalization, population gradients, and excess risk under gradient domination. Appendix~\ref{app:comparison-optimization} compares excess-risk bounds for approximate empirical risk minimization, projected subgradient descent, and smooth empirical risk minimization. 

\subsection{Proof outline and organization}
We outline the four steps in the proof of Theorem~\ref{thm:main}. Let $\Omega=(\gamma_{ij})$ be the sensitivity matrix and let $\nu$ be the law of $S$. The argument combines the contributions of all coordinate pairs at the Frobenius scale without the logarithmic dependence on $n$ in the earlier dyadic approach.

\paragraph{Step 1: Reconstruct the dependent sum from coordinate differences.}
Following the coordinate averaging and difference operators in \citet[Section~1.3]{naor2016} and the product-space decomposition in \citet[Section~8.3]{odonnell2014}, let $\mathsf E_j$ denote averaging over coordinate $j$, and set $\mathsf D_j=\mathrm{Id}-\mathsf E_j$ and $\mathcal L=\sum_{j=1}^n\mathsf D_j$. The Hoeffding projections $\mathsf P_J$ from Section~\ref{subsec:coordinate-averages} diagonalize these operators, with $\mathcal L\mathsf P_J=|J|\mathsf P_J$. The two conditional centering assumptions imply that $\mathsf P_Jh_i=0$ unless $i\in J$ and $|J|\ge2$. For each such set $J$, there are $|J|-1$ coordinates $j\ne i$ for which $\mathsf D_j$ retains the corresponding component. Assigning each contribution the factor $1/(|J|-1)$ recovers the original component.

Let $S'$ be an independent copy of $S$, and let $S^{(i)}$ and $S^{(ij)}$ be obtained by replacing coordinate $i$, or both coordinates $i,j$, using $S'$. For $f\in L_2(\nu;\mathcal H)$, define the replacement differences by
\[
\partial_i f=f(S)-f(S^{(i)}),\qquad \partial_i\partial_j f=f(S)-f(S^{(i)})-f(S^{(j)})+f(S^{(ij)}).
\]
Define $K_{ij}f=\frac12\partial_i\partial_j\sum_{\substack{J\subseteq[n]\\|J|\ge2}}\mathsf P_Jf/(|J|-1)$. The reconstruction above, self-adjointness of the Hoeffding projections, and integration in the replacement coordinates yield
\begin{equation}\label{eq:outline-product-duality}
\mathbb E\left\langle\sum_{i=1}^n h_i(S),f(S)\right\rangle=\sum_{\substack{i,j=1\\i\ne j}}^n\mathbb E\langle\partial_jh_i,K_{ij}f\rangle.
\end{equation}
The sensitivity assumption gives $\|\partial_jh_i\|_{\mathcal H}\le\gamma_{ij}$. Steps 2--3 establish bounds for mixed replacement operators with simpler multipliers, and Step 4 uses these bounds to control $K_{ij}f$ and hence the pairing in~\eqref{eq:outline-product-duality}.

\paragraph{Step 2: Combine coordinate pairs using squared-sensitivity weights.}
Define $\mathcal L^{-1}$ to vanish on constants and to multiply each component of order $|J|\ge1$ by $1/|J|$. To control $K_{ij}f$, we first estimate the mixed replacement operators $\mathsf R_{ij}f=\frac12\partial_i\partial_j\mathcal L^{-1}f$. Assume $\|\Omega\|_F>0$; the zero case follows from sensitivity and centering. Put $\mathcal I=\{(i,j):i\ne j,\ \gamma_{ij}>0\}$ and assign probability $\theta_{ij}=\gamma_{ij}^2/\|\Omega\|_F^2$ to each pair $(i,j)\in\mathcal I$. On $\widehat{\mathcal X}=\mathcal X\times\mathcal X\times\mathcal I$ with probability measure $\widehat{\mathbb P}=\nu\otimes\nu\otimes\theta$, define $T_\Omega$ by
\[
(T_\Omega f)(s,s',i,j)=\frac{\|\Omega\|_F}{\gamma_{ij}}(\mathsf R_{ij}f)(s,s').
\]
The squared weights cancel the normalization when taking the $L_2$ norm. The identity $\mathbb E\|\partial_i\partial_j u\|_{\mathcal H}^2=4\|\mathsf D_i\mathsf D_j u\|_{L_2(\nu;\mathcal H)}^2$ and orthogonality give the contraction $\|T_\Omega f\|_{L_2}\le\|f\|_{L_2}$, proved in Lemma~\ref{lem:product-L2}. The $L_1$ norm, on the other hand, retains one factor of each sensitivity:
\[
\|T_\Omega f\|_{L_1}=\frac{1}{\|\Omega\|_F}\sum_{\substack{i,j=1\\i\ne j}}^n\gamma_{ij}\|\mathsf R_{ij}f\|_{L_1(\nu\otimes\nu;\mathcal H)}.
\]
Thus an $L_q$ estimate for $T_\Omega$, together with this identity, will control the mixed replacement operators with $\|\Omega\|_F$ as the scale.

\paragraph{Step 3: Obtain a weighted $L_q$ estimate through weak type and interpolation.}
Step 4 uses the conjugate exponent $q=p/(p-1)$. For large $p$, this exponent is close to one, so the $L_2$ contraction alone is insufficient. For every $f\in L_2(\nu;\mathcal H)$ and every $t>0$, we establish the weak type $(1,1)$ estimate
\[
\widehat{\mathbb P}\{\|T_\Omega f\|_{\mathcal H}>t\}\le\frac{5}{t}\|f\|_{L_1(\nu;\mathcal H)}.
\]
For $t>\|f\|_{L_1}$, we construct a decomposition $f=\psi+\mathcal Lu$ through a convex minimization problem on $L_2(\nu;\mathcal H)$. Lemma~\ref{lem:product-decomposition} gives $\|\psi\|_{L_2}^2\le t\|f\|_{L_1}$ and $\nu\{u\ne0\}\le\|f\|_{L_1}/t$. The $L_2$ contraction controls $T_\Omega\psi$, whereas the mixed replacement difference of $u$ vanishes unless $u$ is nonzero at one of $S,S^{(i)},S^{(j)},S^{(ij)}$. Conditional on the sampled pair $(i,j)$, each of these four samples has law $\nu$. The union bound is therefore over four samples, and averaging over the pair distribution gives an exceptional probability at most $4\nu\{u\ne0\}$, with no factor for the number of coordinate pairs. Chebyshev's inequality adds at most $\|f\|_{L_1}/t$, giving the weak type constant five; see Lemma~\ref{lem:product-weak}.

By Marcinkiewicz interpolation~\citep[Section~1.3]{grafakos2014}, the weak type estimate and the $L_2$ contraction give $\|T_\Omega f\|_{L_q}\le C_q\|f\|_{L_q}$ for $1<q<2$, with $C_q=\left(\frac{15q}{2(q-1)}+\frac{9q}{2-q}\right)^{1/q}$; see Lemma~\ref{lem:product-interpolation}. Since the enlarged space has probability measure one, $\|T_\Omega f\|_{L_1}\le\|T_\Omega f\|_{L_q}$. The dependence of $C_q$ on $q-1$ will yield linear growth in $p$.

\paragraph{Step 4: Return to the reconstruction and prove the moment bound.}
We relate the operators $K_{ij}$ from Step 1 to the operators $\mathsf R_{ij}$ estimated in Step 3. Averaging over the coordinates other than $i,j$ gives the identity~\eqref{eq:product-shift}. The averages are $L_1$ contractions by Jensen's inequality, and the identity gives $\|K_{ij}f\|_{L_1(\nu\otimes\nu;\mathcal H)}\le2\|\mathsf R_{ij}f\|_{L_1(\nu\otimes\nu;\mathcal H)}$. Combining the estimate in Lemma~\ref{lem:product-K} with~\eqref{eq:outline-product-duality}, we obtain
\[
\left|\mathbb E\left\langle\sum_{i=1}^n h_i(S),f(S)\right\rangle\right|\le\sum_{\substack{i,j=1\\i\ne j}}^n\gamma_{ij}\|K_{ij}f\|_{L_1(\nu\otimes\nu;\mathcal H)}\le2C_q\|\Omega\|_F\|f\|_{L_q(\nu;\mathcal H)}.
\]
For $p\ge3$ and $q=p/(p-1)$, the estimate $C_q\le(15/2)p$ and $L_p$--$L_q$ duality prove the bound $15p\|\Omega\|_F$. For $2\le p<3$, monotonicity of $L_p$ norms and the unsimplified estimate at $p=3$ give $\|\sum_{i=1}^n h_i(S)\|_{L_p(\mathcal H)}\le2(99/2)^{2/3}\|\Omega\|_F<30\|\Omega\|_F\le15p\|\Omega\|_F$. Unlike the approach of \citet{bkz2020}, which sums over dyadic coordinate blocks, our proof uses a single weighted operator on the product space, thereby avoiding an additional $\log n$ factor.

\paragraph{Organization of the paper.} The rest of the paper is organized as follows. 
Section~\ref{sec:preliminaries} introduces the notation and preliminary results.
Section~\ref{sec:proof} proves the concentration theorem and derives its consequences.
Section~\ref{sec:sharpness} establishes sharpness.
Section~\ref{sec:applications} develops the statistical and algorithmic applications where their proofs are given in Appendix~\ref{app:applications}.
Auxiliary concentration and optimization results are proved in Appendix~\ref{app:auxiliary}.

\section{Preliminaries}\label{sec:preliminaries}

\subsection{Probability spaces and norms}\label{subsec:notation}
\paragraph{Notation.}
All logarithms are natural unless stated otherwise. The confidence parameter $\delta$ lies in $(0,1)$ unless a smaller range is specified. We write $\mathbb{P}\{E\}$ for the probability of an event $E$ under the law of the random variables involved, and $\mathbf1_\mathcal{B}$ for the indicator of a measurable set $\mathcal{B}$. The letter $C$ denotes a universal constant that may change between occurrences. We write $a\lesssim b$ if $a\le Cb$, $a\gtrsim b$ if $b\lesssim a$, and $a\asymp b$ if both hold. In sample size conditions, $n\gtrsim b$ requires a sufficiently large constant. We use $O(\cdot)$ and $\Theta(\cdot)$ for asymptotic upper and matching bounds, with constants independent of the asymptotic variable. Unless stated otherwise, probability bounds in Sections~\ref{sec:introduction} and~\ref{sec:main-results} hold with probability at least $1-\delta$ over $S$. For rate comparisons, we assume $n\ge2$ and $0<\delta\le 1/e$, so $\log(e^2/\delta)\asymp\log(1/\delta)$ and we use the latter throughout these sections.

For each $i\in[n]$, let $\nu_i$ be the law of $Z_i$ on $(\mathcal Z_i,\mathcal A_i)$. We write
\[
\nu=\bigotimes_{i=1}^n\nu_i,\qquad
\mathcal X_{-i}=\prod_{j\in[n]\setminus\{i\}}\mathcal Z_j,\qquad
\nu_{-i}=\bigotimes_{j\in[n]\setminus\{i\}}\nu_j.
\]
For $s\in\mathcal X$ and $y\in\mathcal Z_i$, let $s^{(i\leftarrow y)}$ denote the point obtained by replacing coordinate $i$ of $s$ with $y$. We call two samples neighboring if they differ in exactly one coordinate.

For every $p\in[1,\infty)$ and every $f\in L_p(\nu;\mathcal H)$, the norm is
\[
\|f\|_{L_p(\nu;\mathcal H)}
=\left(\int_{\mathcal X}\|f(s)\|_{\mathcal H}^p\,\mathrm{d}\nu(s)\right)^{1/p}.
\]
Since $S$ has law $\nu$, we have $\|f\|_{L_p(\nu;\mathcal H)}=\|f(S)\|_{L_p(\mathcal H)}$. The norm $\|f\|_{L_\infty(\nu;\mathcal H)}$ is the essential supremum of $\|f(s)\|_{\mathcal H}$ with respect to $\nu$. For scalar random variables, $\|X\|_{L_p}=(\mathbb E|X|^p)^{1/p}$ whenever $1\le p<\infty$ and the moment is finite. For a random variable $X$ taking values in a Hilbert space $\mathcal K$, $\|X\|_{L_\infty(\mathcal K)}$ is the essential supremum of $\|X\|_{\mathcal K}$. For vectors in $\mathbb R^d$, $\|\cdot\|_2$ denotes the Euclidean norm rather than a moment norm. Inner products $\langle\cdot,\cdot\rangle$ are taken in the Hilbert space containing their arguments. Expectations of random variables taking values in a Hilbert space are understood as Bochner integrals.

\subsection{Coordinate averages and Hoeffding projections}\label{subsec:coordinate-averages}
For each $j\in[n]$ and $1\le p\le\infty$, define the linear operators $\mathsf E_j,\mathsf D_j:L_p(\nu;\mathcal H)\to L_p(\nu;\mathcal H)$ by $(\mathsf E_j f)(S)=\mathbb E[f(S)\mid S_{-j}]$ almost surely and $\mathsf D_j=\mathrm{Id}-\mathsf E_j$, where $\mathrm{Id}$ denotes the identity operator. These operators extend the coordinate averages and differences on the discrete cube in \citet[Section~1.3]{naor2016} and the scalar coordinate projections in \citet[Definition~8.17 and Remark~8.18]{odonnell2014} to Hilbert-valued functions on product spaces. On the uniform cube, $\mathsf D_j$ is one half of the sign-reversal difference in \citet[Eq.~(7)]{naor2016}.

By Jensen's inequality, $\|\mathsf E_j f\|_{L_p(\nu;\mathcal H)}\le\|f\|_{L_p(\nu;\mathcal H)}$ for every $1\le p\le\infty$. Averaging twice in the same coordinate gives $\mathsf E_j^2=\mathsf E_j$, and Fubini's theorem gives $\mathsf E_j\mathsf E_k=\mathsf E_k\mathsf E_j$ for all $j,k\in[n]$. It follows that $\mathsf E_j\mathsf D_j=0$. On $L_2(\nu;\mathcal H)$, the operator $\mathsf E_j$ is the orthogonal projection onto the functions that do not depend on coordinate $j$, whereas $\mathsf D_j$ is the orthogonal projection onto the functions with zero average in that coordinate. Thus $\mathsf E_j^*=\mathsf E_j$, $\mathsf D_j^*=\mathsf D_j$, and $\mathsf D_j^2=\mathsf D_j$ on $L_2(\nu;\mathcal H)$.

For $J\subseteq[n]$, define $\mathsf P_J:L_2(\nu;\mathcal H)\to L_2(\nu;\mathcal H)$ by
\[
\mathsf P_J=\prod_{j\in J}\mathsf D_j\prod_{j\in[n]\setminus J}\mathsf E_j.
\]
These are the orthogonal Hoeffding projections~\citep{hoeffding1948}; see also \citet[Theorem~8.35]{odonnell2014} for their scalar counterparts on product spaces. The function $\mathsf P_J f$ depends only on coordinates in $J$ and has zero average in each of those coordinates. In particular, $\mathsf P_\varnothing f$ is the constant function with value $\mathbb E[f(S)]$. Since all factors commute and are orthogonal projections, $\mathsf P_J^*=\mathsf P_J=\mathsf P_J^2$ and $\mathsf P_J\mathsf P_K=0$ whenever $J\ne K$. Expanding $\prod_{j=1}^n(\mathsf E_j+\mathsf D_j)=\mathrm{Id}$ gives the following identities, the second of which holds for every $j\in[n]$:
\[
\sum_{J\subseteq[n]}\mathsf P_J=\mathrm{Id},\qquad
\sum_{\substack{J\subseteq[n]\\j\in J}}\mathsf P_J=\mathsf D_j.
\]

\subsection{Concentration for independent sums}\label{subsec:independent-concentration}

The following moment bound follows from bounded differences and orthogonality in a Hilbert space; related moment estimates can be found in \citet{boucheron2013}. See also the Marcinkiewicz--Zygmund argument in \citet[Lemma~14]{fanlei2024}.

\begin{lemma}\label{lem:independent}
Let $X_1,\ldots,X_n$ be independent centered random variables in a real separable Hilbert space $\mathcal H$. Suppose $\|X_i\|_{\mathcal H}\le M$ almost surely for every $i\in[n]$, where $M\ge0$. Then, for every $p\ge2$,
\[
\left\|\sum_{i=1}^nX_i\right\|_{L_p(\mathcal H)}\le2M\sqrt{pn}.
\]
\end{lemma}

We also use a Bernstein inequality of the type established for martingales in smooth Banach spaces by \citet{pinelis1994}.

\begin{lemma}\label{lem:bernstein}
Let $X_1,\ldots,X_n$ be independent centered random variables in a real separable Hilbert space $\mathcal H$. Suppose $\|X_i\|_{\mathcal H}\le M$ almost surely for every $i\in[n]$, where $M\ge0$, and set $\sigma^2=\sum_{i=1}^n\mathbb E\|X_i\|_{\mathcal H}^2$. Then, for every $t>0$,
\begin{equation}\label{eq:bernstein} \mathbb{P} \left\{\left\|\sum_{i=1}^nX_i\right\|_{\mathcal H}>2\sigma\sqrt{t}+2Mt\right\}\le2e^{-t}. \end{equation}
\end{lemma}

Proofs with these constants are given in Appendix~\ref{app:auxiliary}.

\subsection{Learning and optimization notation}\label{subsec:learning-notation}

For the learning applications, we specialize the product space $\mathcal X$ from Section~\ref{sec:main} to independent and identically distributed observations on a common space $\mathcal Z$ with distribution $\mathrm{P}$. Thus $\mathcal Z_i=\mathcal Z$ for every $i\in[n]$, $\mathcal X=\mathcal Z^n$, and $S=(Z_1,\ldots,Z_n)\sim\mathrm{P}^n$. The hypothesis or parameter set is $\mathcal W$, and the loss is $\ell:\mathcal W\times\mathcal Z\to\mathbb R$. Following the standard stability terminology of \citet{bousquet2002}, the population and empirical risks are
\[
R(w)=\mathbb E_Z\ell(w,Z),\qquad
R_{\text{emp}}(w)=\frac{1}{n}\sum_{i=1}^n\ell(w,Z_i).
\]
When the sample must be explicit, we write $R_{\text{emp}}(w;s)=\frac{1}{n}\sum_{i=1}^n\ell(w,z_i)$ for $s=(z_1,\ldots,z_n)$. Write $R^\star=\inf_{w\in\mathcal W}R(w)$ whenever this infimum is finite. If a population minimizer $w^\star$ exists, then $R^\star=R(w^\star)$. The excess risk of $w$ is $R(w)-R^\star\ge0$; it measures statistical suboptimality relative to the best population decision in $\mathcal W$. For an algorithm $A$, the generalization gap and empirical optimization error are, respectively,
\[
R(A(S))-R_{\text{emp}}(A(S))\qquad\text{and}\qquad
R_{\text{emp}}(A(S))-\inf_{w\in\mathcal W}R_{\text{emp}}(w).
\]

When $\mathcal W$ is a nonempty closed convex subset of a Hilbert space $\mathcal H$, $\Pi_{\mathcal W}(x)$ denotes its metric projection: the unique point in $\mathcal W$ minimizing $\|x-w\|_{\mathcal H}$. A function $\Psi$ is $\mu$-strongly convex on $\mathcal W$ if $w\mapsto\Psi(w)-(\mu/2)\|w\|_{\mathcal H}^2$ is convex there. An $L$-smooth function is differentiable with an $L$-Lipschitz gradient. For a convex objective, $g\in\partial\Psi(w)$ means that $\Psi(u)\ge\Psi(w)+\langle g,u-w\rangle$ for every $u\in\mathcal W$~\citep{nesterov2004}.

For the differentiable applications, $\mathcal W\subseteq\mathbb R^d$ is nonempty, closed, and convex. For every nonempty set $\mathcal{C}\subseteq\mathbb R^d$ and every $x\in\mathbb R^d$, define $\operatorname{dist}(x,\mathcal{C})=\inf_{y\in\mathcal{C}}\|x-y\|_2$. Define the normal cone at $w\in\mathcal W$ by
\[
\mathcal{N}_{\mathcal W}(w)=\{u\in\mathbb R^d:\langle u,y-w\rangle\le0\text{ for every }y\in\mathcal W\}.
\]
For a differentiable function $\Psi$, define the stationarity residual by
\[
\mathcal{R}_\Psi(w)=\operatorname{dist}\left(0,\nabla\Psi(w)+\mathcal{N}_{\mathcal W}(w)\right).
\]
The residual vanishes at a differentiable constrained minimizer. If $\mathcal W=\mathbb R^d$, it equals $\|\nabla\Psi(w)\|_2$. The applications state the required quadratic growth and gradient domination assumptions explicitly; sufficient conditions are given in Lemmas~\ref{lem:optimization-facts} and~\ref{lem:strong-convexity-pl}.

\section{Proof of the Theorem~\ref{thm:main} and its consequences}\label{sec:proof}

\subsection{Concentration on arbitrary product spaces}\label{subsec:transfer}
The proof of the moment bound in \citet[Theorem~4]{bkz2020} uses a dyadic decomposition over $\lceil\log_2 n\rceil$ scales, building on the sample splitting argument of \citet{feldman2019}. We instead use the inverse of the product Laplacian and combine the mixed replacement differences in a single operator, assigning probabilities to ordered coordinate pairs in proportion to their squared sensitivities.

Let $\nu$ be the product law of $S$. For the auxiliary estimates below, assume $n\ge2$ and $\|\Omega\|_F>0$.

\paragraph{Mixed replacement operators.} Use the coordinate averages and Hoeffding projections from Section~\ref{subsec:coordinate-averages}, and set $\mathcal L=\sum_{i=1}^n\mathsf D_i$, extending the cube Laplacian in \citet[Section~3]{naor2016} to the present product space. For every $J\subseteq[n]$ and every $i\in[n]$, we have $\mathsf D_i\mathsf P_J=\mathsf P_J$ if $i\in J$, and $\mathsf D_i\mathsf P_J=0$ otherwise. Hence $\mathcal L\mathsf P_J=|J|\mathsf P_J$. Define $\mathcal L^{-1}=\sum_{\substack{J\subseteq[n]\\J\ne\varnothing}}|J|^{-1}\mathsf P_J$, which vanishes on constants. Both $\mathcal L$ and $\mathcal L^{-1}$ are bounded self-adjoint operators on $L_2(\nu;\mathcal H)$, since they are finite linear combinations of orthogonal projections with real coefficients. For every $u\in L_2(\nu;\mathcal H)$, using orthogonality of the Hoeffding projections and the identities $\mathsf P_{\varnothing}u=\mathbb E u(S)$ and $\sum_{J\subseteq[n]}\mathsf P_J=\mathrm{Id}$ gives
\begin{equation}\label{eq:product-gap}
\begin{aligned}
\mathbb E\langle u(S),(\mathcal Lu)(S)\rangle&=\sum_{J\subseteq[n]}|J|\|\mathsf P_Ju\|_{L_2(\nu;\mathcal H)}^2\\
&\ge\sum_{\substack{J\subseteq[n]\\J\ne\varnothing}}\|\mathsf P_Ju\|_{L_2(\nu;\mathcal H)}^2=\|u-\mathbb E u(S)\|_{L_2(\nu;\mathcal H)}^2.
\end{aligned}
\end{equation}
In addition, $\mathcal L^{-1}\mathcal Lu=\sum_{\substack{J\subseteq[n]\\J\ne\varnothing}}\mathsf P_Ju=u-\mathbb E u(S)$.

We use the independent-coordinate replacement construction underlying the Efron--Stein inequality~\citep[Chapter~3]{boucheron2013}. Let $S'$ be an independent copy of $S$. For $s=(z_1,\ldots,z_n)$ and $s'=(z'_1,\ldots,z'_n)$, write $s^{(i)}=s^{(i\leftarrow z'_i)}$ for each $i\in[n]$ and let $s^{(ij)}$ replace both coordinates $i,j$ using $s'$ for every pair of distinct indices $i,j\in[n]$. Define, for $f\in L_2(\nu;\mathcal H)$,
\[
\begin{aligned}
(\partial_i f)(s,s')&=f(s)-f(s^{(i)}),\\
(\partial_i\partial_j f)(s,s')&=f(s)-f(s^{(i)})-f(s^{(j)})+f(s^{(ij)}).
\end{aligned}
\]
The notation for random samples is $S^{(i)}$ and $S^{(ij)}$. Each of $S$, $S^{(i)}$, $S^{(j)}$, and $S^{(ij)}$ has law $\nu$. Thus these differences are well defined in $L_2(\nu\otimes\nu;\mathcal H)$ independently of the representative of $f$.

For $u,v\in L_2(\nu;\mathcal H)$ and each $k\in[n]$, conditional on $S_{-k}$ the random variables $Z_k$ and $Z'_k$ are independent with law $\nu_k$. Consequently,
\[
\begin{aligned}
\mathbb E\langle u(S),v(S^{(k)})\rangle&=\mathbb E\langle (\mathsf E_ku)(S),(\mathsf E_kv)(S)\rangle\\
&=\mathbb E\langle(\mathsf E_ku)(S),v(S)\rangle.
\end{aligned}
\]
The second equality follows by conditioning on $S_{-k}$, since $(\mathsf E_ku)(S)$ depends only on $S_{-k}$. Exchanging $Z_k$ and $Z'_k$ preserves their joint law and gives the same expectation for $\langle u(S^{(k)}),v(S)\rangle$. Expanding the inner product of the differences therefore gives
\begin{equation}\label{eq:product-replacement}
\begin{aligned}
\mathbb E\langle(\partial_k u)(S,S'),(\partial_k v)(S,S')\rangle&=\mathbb E\langle u(S),v(S)\rangle+\mathbb E\langle u(S^{(k)}),v(S^{(k)})\rangle\\
&\quad-\mathbb E\langle u(S),v(S^{(k)})\rangle-\mathbb E\langle u(S^{(k)}),v(S)\rangle\\
&=2\mathbb E\langle u(S),v(S)\rangle-2\mathbb E\langle(\mathsf E_ku)(S),v(S)\rangle\\
&=2\mathbb E\langle(\mathsf D_ku)(S),v(S)\rangle.
\end{aligned}
\end{equation}
Summing~\eqref{eq:product-replacement} over $k=1,\ldots,n$ gives the product-space Dirichlet identity
\begin{equation}\label{eq:product-dirichlet}
\mathbb E\langle(\mathcal Lu)(S),v(S)\rangle=\frac12\sum_{k=1}^n\mathbb E\langle(\partial_k u)(S,S'),(\partial_k v)(S,S')\rangle.
\end{equation}
For $v=u$, self-adjointness and idempotence of $\mathsf D_k$ give $\mathbb E\|\partial_k u\|_{\mathcal H}^2=2\|\mathsf D_ku\|_{L_2(\nu;\mathcal H)}^2$. To apply this identity twice, fix distinct indices $i,j\in[n]$. For a fixed replacement value $z'_j$, the function $s\mapsto(\partial_j u)(s,s')$ depends on $s'$ only through $z'_j$. Its replacement difference in coordinate $i$ is $\partial_i\partial_j u$. In this calculation, let $\mathsf D_i$ act on $s$ with $s'$ fixed. Since $i\ne j$, the average in coordinate $i$ commutes with replacement in coordinate $j$, so $\mathsf D_i(\partial_j u)=\partial_j(\mathsf D_i u)$. Applying the preceding identity first in coordinate $i$, integrating over $Z'_j$, and then applying it in coordinate $j$ gives
\begin{equation}\label{eq:product-mixed-difference}
\begin{aligned}
\mathbb E\|\partial_i\partial_j u\|_{\mathcal H}^2&=2\mathbb E\|\mathsf D_i(\partial_j u)\|_{\mathcal H}^2\\
&=2\mathbb E\|\partial_j(\mathsf D_i u)\|_{\mathcal H}^2=4\|\mathsf D_i\mathsf D_j u\|_{L_2(\nu;\mathcal H)}^2.
\end{aligned}
\end{equation}
Here and below, expectations of replacement differences are taken over both $S$ and $S'$.

Motivated by the discrete Riesz transforms in \citet[Section~1.3.1]{naor2016}, we use mixed differences of the inverse Laplacian. For every pair of distinct indices $i,j\in[n]$, define the operators from $L_2(\nu;\mathcal H)$ to $L_2(\nu\otimes\nu;\mathcal H)$ by
\[
\mathsf R_{ij}f=\frac12\partial_i\partial_j\mathcal L^{-1}f,\qquad K_{ij}f=\frac12\partial_i\partial_j\left(\sum_{\substack{J\subseteq[n]\\|J|\ge2}}\frac{\mathsf P_Jf}{|J|-1}\right).
\]
Let $\mathcal I=\{(i,j)\in[n]\times[n]:i\ne j,\ \gamma_{ij}>0\}$ and $\theta_{ij}=\gamma_{ij}^2/\|\Omega\|_F^2$ for each $(i,j)\in\mathcal I$. The weights define a probability distribution $\theta$ on $\mathcal I$, since $\sum_{(i,j)\in\mathcal I}\gamma_{ij}^2=\|\Omega\|_F^2$. In this subsection, equip $\widehat{\mathcal X}=\mathcal X\times\mathcal X\times\mathcal I$ with $\widehat{\mathbb P}=\nu\otimes\nu\otimes\theta$, and define
\[
(T_\Omega f)(s,s',i,j)=\frac{\|\Omega\|_F}{\gamma_{ij}}(\mathsf R_{ij}f)(s,s').
\]
All norms on $\widehat{\mathcal X}$ below use $\widehat{\mathbb P}$.

\begin{lemma}\label{lem:product-L2}
For every $f\in L_2(\nu;\mathcal H)$, we have $\|T_\Omega f\|_{L_2(\widehat{\mathcal X};\mathcal H)}\le\|f\|_{L_2(\nu;\mathcal H)}$.
\end{lemma}
\begin{proof}
Equation~\eqref{eq:product-mixed-difference} gives $\|\mathsf R_{ij}f\|_{L_2(\nu\otimes\nu;\mathcal H)}^2=\|\mathsf D_i\mathsf D_j\mathcal L^{-1}f\|_{L_2(\nu;\mathcal H)}^2$. Since $\mathsf D_i\mathsf D_j\mathsf P_J=\mathsf P_J$ when $\{i,j\}\subseteq J$ and vanishes otherwise, the expansion of $\mathcal L^{-1}$ and orthogonality of the projections give
\[
\begin{aligned}
\mathsf D_i\mathsf D_j\mathcal L^{-1}f&=\sum_{\substack{J\subseteq[n]\\\{i,j\}\subseteq J}}\frac{\mathsf P_Jf}{|J|},\\
\|\mathsf D_i\mathsf D_j\mathcal L^{-1}f\|_{L_2(\nu;\mathcal H)}^2&=\sum_{\substack{J\subseteq[n]\\\{i,j\}\subseteq J}}\frac{\|\mathsf P_Jf\|_{L_2(\nu;\mathcal H)}^2}{|J|^2}.
\end{aligned}
\]

The weights cancel the normalization in the squared norm, so
\[
\begin{aligned}
\|T_\Omega f\|_{L_2(\widehat{\mathcal X};\mathcal H)}^2&\mathrel{\mathmakebox[2.1em][c]{=}}\sum_{(i,j)\in\mathcal I}\theta_{ij}\frac{\|\Omega\|_F^2}{\gamma_{ij}^2}\|\mathsf R_{ij}f\|_{L_2(\nu\otimes\nu;\mathcal H)}^2\\
&\mathrel{\mathmakebox[2.1em][c]{=}}\sum_{(i,j)\in\mathcal I}\|\mathsf R_{ij}f\|_{L_2(\nu\otimes\nu;\mathcal H)}^2\\
&\mathrel{\mathmakebox[2.1em][c]{\overset{\eqref{eq:product-mixed-difference}}{=}}}\sum_{(i,j)\in\mathcal I}\|\mathsf D_i\mathsf D_j\mathcal L^{-1}f\|_{L_2(\nu;\mathcal H)}^2.
\end{aligned}
\]
Substituting the formula for the squared norm of $\mathsf D_i\mathsf D_j\mathcal L^{-1}f$ and extending the sum to all ordered pairs of distinct indices gives
\[
\|T_\Omega f\|_{L_2(\widehat{\mathcal X};\mathcal H)}^2\le\sum_{\substack{i,j=1\\i\ne j}}^n\sum_{\substack{J\subseteq[n]\\\{i,j\}\subseteq J}}\frac{\|\mathsf P_Jf\|_{L_2(\nu;\mathcal H)}^2}{|J|^2}.
\]
For each $J\subseteq[n]$ with $|J|\ge2$, the number of ordered pairs of distinct indices in $J$ is $|J|(|J|-1)$. Changing the order of summation and using orthogonality, we obtain
\begin{equation}\label{eq:product-T2}
\begin{aligned}
\|T_\Omega f\|_{L_2(\widehat{\mathcal X};\mathcal H)}^2&\le\sum_{\substack{J\subseteq[n]\\|J|\ge2}}\frac{|J|(|J|-1)}{|J|^2}\|\mathsf P_Jf\|_{L_2(\nu;\mathcal H)}^2\\
&\le\sum_{J\subseteq[n]}\|\mathsf P_Jf\|_{L_2(\nu;\mathcal H)}^2=\|f\|_{L_2(\nu;\mathcal H)}^2.
\end{aligned}
\end{equation}
Taking square roots gives the contraction.
\end{proof}

The variational argument in the next lemma follows the obstacle and partial balayage method of \citet[Theorem~1.2]{ouyang2026}. A formulation for functions with values in a finite-dimensional Hilbert space, based on a quadratic energy and an $L_1$ penalty, is given by \citet[Theorem~1.1]{linbalayage}. We use the product Laplacian in the quadratic energy to obtain a decomposition suitable for mixed replacement operators.
\begin{lemma}\label{lem:product-decomposition}
Let $f\in L_2(\nu;\mathcal H)$ and $t>\|f\|_{L_1(\nu;\mathcal H)}$. Then there exist $u,\psi\in L_2(\nu;\mathcal H)$ such that $f=\psi+\mathcal Lu$ and
\[
\nu\{s:u(s)\ne0\}\le\frac{\|f\|_{L_1(\nu;\mathcal H)}}{t},\qquad \|\psi\|_{L_2(\nu;\mathcal H)}^2\le t\|f\|_{L_1(\nu;\mathcal H)}.
\]
\end{lemma}
\begin{proof}
Set $m_f=\|f\|_{L_1(\nu;\mathcal H)}$. Consider the convex functional on $L_2(\nu;\mathcal H)$,
\[
\mathcal F(u)=\frac12\mathbb E\langle u(S),(\mathcal Lu)(S)\rangle+t\mathbb E\|u(S)\|_{\mathcal H}-\mathbb E\langle f(S),u(S)\rangle.
\]
Let $\bar u=\mathbb E u(S)$. Splitting the linear term and applying Cauchy--Schwarz give
\[
\begin{aligned}
\mathbb E\langle f(S),u(S)\rangle&=\mathbb E\langle f(S)-\mathbb E f(S),u(S)-\bar u\rangle+\langle\mathbb E f(S),\bar u\rangle\\
&\le\|f-\mathbb E f(S)\|_{L_2(\nu;\mathcal H)}\|u-\bar u\|_{L_2(\nu;\mathcal H)}+\|\mathbb E f(S)\|_{\mathcal H}\|\bar u\|_{\mathcal H}.
\end{aligned}
\]
By Jensen's inequality, $\mathbb E\|u(S)\|_{\mathcal H}\ge\|\bar u\|_{\mathcal H}$. Combining these two inequalities with~\eqref{eq:product-gap} and using $ab\le a^2/4+b^2$ gives
\[
\begin{aligned}
\mathcal F(u)&\ge\frac12\|u-\bar u\|_{L_2(\nu;\mathcal H)}^2-\|f-\mathbb E f(S)\|_{L_2(\nu;\mathcal H)}\|u-\bar u\|_{L_2(\nu;\mathcal H)}+\left(t-\|\mathbb E f(S)\|_{\mathcal H}\right)\|\bar u\|_{\mathcal H}\\
&\ge\frac14\|u-\bar u\|_{L_2(\nu;\mathcal H)}^2-\|f-\mathbb E f(S)\|_{L_2(\nu;\mathcal H)}^2+\left(t-\|\mathbb E f(S)\|_{\mathcal H}\right)\|\bar u\|_{\mathcal H}.
\end{aligned}
\] Since $t>m_f\ge\|\mathbb E f(S)\|_{\mathcal H}$ and $\|u\|_{L_2(\nu;\mathcal H)}^2=\|u-\bar u\|_{L_2(\nu;\mathcal H)}^2+\|\bar u\|_{\mathcal H}^2$, the lower bound tends to infinity as $\|u\|_{L_2(\nu;\mathcal H)}$ tends to infinity.

The quadratic term is continuous and convex because $\mathcal L$ is bounded, self-adjoint and nonnegative. The norm term is convex and continuous on $L_2(\nu;\mathcal H)$, since $|\mathbb E\|u(S)\|_{\mathcal H}-\mathbb E\|v(S)\|_{\mathcal H}|\le\|u-v\|_{L_2(\nu;\mathcal H)}$. The remaining term is continuous and linear. Thus $\mathcal F$ is continuous and convex, hence weakly lower semicontinuous. The lower bound shows that its infimum is finite and that a minimizing sequence is bounded. A bounded sequence in a Hilbert space has a weakly convergent subsequence. By weak lower semicontinuity, its weak limit attains the infimum of $\mathcal F$. We fix such a minimizer $u$ and set $\psi=f-\mathcal Lu$.

We next derive the pointwise optimality conditions. For every $v\in L_2(\nu;\mathcal H)$ and every $\alpha>0$, minimality of $u$, self-adjointness of $\mathcal L$, and the triangle inequality give
\[
\begin{aligned}
0&\le\mathcal F(u+\alpha v)-\mathcal F(u)\\
&=-\alpha\mathbb E\langle\psi(S),v(S)\rangle+\frac{\alpha^2}{2}\mathbb E\langle v(S),(\mathcal Lv)(S)\rangle+t\mathbb E\left[\|u(S)+\alpha v(S)\|_{\mathcal H}-\|u(S)\|_{\mathcal H}\right]\\
&\le-\alpha\mathbb E\langle\psi(S),v(S)\rangle+\frac{\alpha^2}{2}\mathbb E\langle v(S),(\mathcal Lv)(S)\rangle+\alpha t\mathbb E\|v(S)\|_{\mathcal H}.
\end{aligned}
\]
Divide by $\alpha$ and let $\alpha$ decrease to zero. Replacing $v$ by $-v$ in the resulting inequality gives
\[
|\mathbb E\langle\psi(S),v(S)\rangle|\le t\mathbb E\|v(S)\|_{\mathcal H}.
\]
Take $v(s)=\psi(s)/\|\psi(s)\|_{\mathcal H}$ on the set $\{s:\|\psi(s)\|_{\mathcal H}>t\}$ and $v(s)=0$ elsewhere. This is a bounded function in $L_2(\nu;\mathcal H)$. If that set had positive measure, its integral of $\|\psi(s)\|_{\mathcal H}$ would be strictly larger than $t$ times its measure, contrary to the last inequality. Hence $\|\psi(s)\|_{\mathcal H}\le t$ almost everywhere.

For $\alpha>-1$, we have $\|(1+\alpha)u(S)\|_{\mathcal H}=(1+\alpha)\|u(S)\|_{\mathcal H}$. Using the variation $(1+\alpha)u$ in the functional gives
\[
0\le\mathcal F((1+\alpha)u)-\mathcal F(u)=\alpha\left(t\mathbb E\|u(S)\|_{\mathcal H}-\mathbb E\langle\psi(S),u(S)\rangle\right)+\frac{\alpha^2}{2}\mathbb E\langle u(S),(\mathcal Lu)(S)\rangle.
\]
Letting $\alpha$ approach zero from the positive and negative sides yields $\mathbb E\langle\psi(S),u(S)\rangle=t\mathbb E\|u(S)\|_{\mathcal H}$. The bound on $\psi$ and Cauchy--Schwarz show that $t\|u(S)\|_{\mathcal H}-\langle\psi(S),u(S)\rangle$ is nonnegative almost surely. Its expectation is zero, so it vanishes almost surely. If $u(s)\ne0$, equality holds in both inequalities $\langle\psi(s),u(s)\rangle\le\|\psi(s)\|_{\mathcal H}\|u(s)\|_{\mathcal H}\le t\|u(s)\|_{\mathcal H}$. Hence, almost everywhere,
\[
\|\psi(s)\|_{\mathcal H}\le t,\qquad \psi(s)=t\frac{u(s)}{\|u(s)\|_{\mathcal H}}\quad\text{if }u(s)\ne0.
\]

We next show that $\psi$ retains the $L_1$ bound of $f$. Define $v(s)=\psi(s)/\|\psi(s)\|_{\mathcal H}$ when $\psi(s)\ne0$, and $v(s)=0$ otherwise. Then $\|v(s)\|_{\mathcal H}\le1$, $\langle\psi(s),v(s)\rangle=\|\psi(s)\|_{\mathcal H}$, and the preceding pointwise identity gives $\langle u(s),v(s)\rangle=\|u(s)\|_{\mathcal H}$ almost everywhere. For every pair of points $s,\tilde s$ satisfying these identities, Cauchy--Schwarz implies
\[
\begin{aligned}
\langle u(s)-u(\tilde s),v(s)-v(\tilde s)\rangle&=\|u(s)\|_{\mathcal H}+\|u(\tilde s)\|_{\mathcal H}-\langle u(s),v(\tilde s)\rangle-\langle u(\tilde s),v(s)\rangle\\
&\ge\|u(s)\|_{\mathcal H}+\|u(\tilde s)\|_{\mathcal H}-\|u(s)\|_{\mathcal H}-\|u(\tilde s)\|_{\mathcal H}=0.
\end{aligned}
\]
Both $S$ and $S^{(k)}$ have law $\nu$ for each $k\in[n]$, so the inequality holds almost surely at this pair of points. By~\eqref{eq:product-dirichlet},
\[
\mathbb E\langle(\mathcal Lu)(S),v(S)\rangle=\frac12\sum_{k=1}^n\mathbb E\langle(\partial_k u)(S,S'),(\partial_k v)(S,S')\rangle\ge0.
\]
Substituting $\psi=f-\mathcal Lu$ and using $\mathbb E\langle(\mathcal Lu)(S),v(S)\rangle\ge0$, $\|v(s)\|_{\mathcal H}\le1$, and Cauchy--Schwarz yields
\[
\begin{aligned}
\mathbb E\|\psi(S)\|_{\mathcal H}&=\mathbb E\langle\psi(S),v(S)\rangle\\
&=\mathbb E\langle f(S),v(S)\rangle-\mathbb E\langle(\mathcal Lu)(S),v(S)\rangle\\
&\le\mathbb E\|f(S)\|_{\mathcal H}=m_f.
\end{aligned}
\]
Let $E_0=\{s:u(s)\ne0\}$. Since $\|\psi(s)\|_{\mathcal H}=t$ almost everywhere on $E_0$, we have $t\nu(E_0)=\int_{E_0}\|\psi(s)\|_{\mathcal H}\,\mathrm{d}\nu(s)\le m_f$. Since $\|\psi(s)\|_{\mathcal H}\le t$ almost everywhere on $\mathcal X$, the same $L_1$ estimate controls the second moment. Thus
\[
\nu(E_0)\le\frac{m_f}{t},\qquad \|\psi\|_{L_2(\nu;\mathcal H)}^2=\mathbb E\|\psi(S)\|_{\mathcal H}^2\le t\mathbb E\|\psi(S)\|_{\mathcal H}\le tm_f.
\]
The definition of $\psi$ gives $f=\psi+\mathcal Lu$, and these are the stated bounds.
\end{proof}

In Gaussian space, \citet[Sections~3.1--3.4]{chen2026gaussian} combine a variational decomposition with a weak type estimate for a second-order inverse-generator transform, interpolation, and duality. On the product space, the mixed replacement difference need not vanish where $u$ vanishes. We instead use the support estimate of Lemma~\ref{lem:product-decomposition} at four replacement samples and average over the pair distribution $\theta$ to control the exceptional set in $\widehat{\mathcal X}$. The resulting weak type estimate is then used for interpolation; see \citet[Section~1.3]{grafakos2014}.

\begin{lemma}\label{lem:product-weak}
For every $f\in L_2(\nu;\mathcal H)$ and every $t>0$, we have
\begin{equation}\label{eq:product-weak}
\widehat{\mathbb P}\{\|T_\Omega f\|_{\mathcal H}>t\}\le\frac{5}{t}\|f\|_{L_1(\nu;\mathcal H)}.
\end{equation}
\end{lemma}
\begin{proof}
Set $m_f=\|f\|_{L_1(\nu;\mathcal H)}$. For $t\le m_f$, the right-hand side is at least $1$, so assume $t>m_f$. By Lemma~\ref{lem:product-decomposition}, there exist $u,\psi\in L_2(\nu;\mathcal H)$ such that $f=\psi+\mathcal Lu$, $\|\psi\|_{L_2(\nu;\mathcal H)}^2\le tm_f$, and $\nu(E_0)\le m_f/t$, where $E_0=\{s:u(s)\ne0\}$. Let $\bar u=\mathbb E u(S)$.

Since $f-\psi=\mathcal Lu$ and $\mathcal L^{-1}\mathcal Lu=u-\bar u$, the definition of $T_\Omega$ gives
\[
\begin{aligned}
(T_\Omega(f-\psi))(s,s',i,j)&=\frac{\|\Omega\|_F}{2\gamma_{ij}}\left(\partial_i\partial_j\mathcal L^{-1}\mathcal Lu\right)(s,s')\\
&=\frac{\|\Omega\|_F}{2\gamma_{ij}}\left(\partial_i\partial_j(u-\bar u)\right)(s,s')\\
&=\frac{\|\Omega\|_F}{2\gamma_{ij}}\left(u(s)-u(s^{(i)})-u(s^{(j)})+u(s^{(ij)})\right).
\end{aligned}
\]
Thus $T_\Omega(f-\psi)$ vanishes outside the exceptional set
\[
\widehat E_0=\{(s,s',i,j)\in\widehat{\mathcal X}:\{s,s^{(i)},s^{(j)},s^{(ij)}\}\cap E_0\ne\varnothing\}.
\]
For each fixed pair $(i,j)\in\mathcal I$, each of $S,S^{(i)},S^{(j)},S^{(ij)}$ has law $\nu$. Applying the union bound for this pair and averaging over the probability distribution $\theta$ give
\[
\begin{aligned}
\widehat{\mathbb P}(\widehat E_0)&\le\sum_{(i,j)\in\mathcal I}\theta_{ij}\left(\mathbb P\{S\in E_0\}+\mathbb P\{S^{(i)}\in E_0\}+\mathbb P\{S^{(j)}\in E_0\}+\mathbb P\{S^{(ij)}\in E_0\}\right)\\
&=4\nu(E_0)\sum_{(i,j)\in\mathcal I}\theta_{ij}=4\nu(E_0)\le\frac{4m_f}{t}.
\end{aligned}
\]
Outside this set, $T_\Omega f=T_\Omega\psi$ almost surely. Chebyshev's inequality and~\eqref{eq:product-T2} therefore yield
\[
\begin{aligned}
\widehat{\mathbb P}\{\|T_\Omega f\|_{\mathcal H}>t\}&\le\widehat{\mathbb P}(\widehat E_0)+\widehat{\mathbb P}\{\|T_\Omega\psi\|_{\mathcal H}>t\}\\
&\le\frac{4m_f}{t}+\frac{\|T_\Omega\psi\|_{L_2(\widehat{\mathcal X};\mathcal H)}^2}{t^2}\\
&\le\frac{4m_f}{t}+\frac{\|\psi\|_{L_2(\nu;\mathcal H)}^2}{t^2}\le\frac{5m_f}{t}.
\end{aligned}
\]
Substituting $m_f=\|f\|_{L_1(\nu;\mathcal H)}$ gives~\eqref{eq:product-weak}.
\end{proof}

\begin{lemma}\label{lem:product-interpolation}
Let $f\in L_2(\nu;\mathcal H)$ and $1<q<2$. Set $C_q=\left(\frac{15q}{2(q-1)}+\frac{9q}{2-q}\right)^{1/q}$. Then
\begin{equation}\label{eq:product-Tq}
\|T_\Omega f\|_{L_q(\widehat{\mathcal X};\mathcal H)}\le C_q\|f\|_{L_q(\nu;\mathcal H)}.
\end{equation}
\end{lemma}
\begin{proof}
We combine the weak type estimate and the $L_2$ contraction by the distribution-function argument of Marcinkiewicz interpolation~\citep[Section~1.3]{grafakos2014}. For each $t>0$, split $f=f_{>t}+f_{\le t}$, with
\[
f_{>t}(s)=f(s)\mathbf1_{\{\|f(s)\|_{\mathcal H}>t\}},\qquad f_{\le t}(s)=f(s)\mathbf1_{\{\|f(s)\|_{\mathcal H}\le t\}}.
\]
Both functions belong to $L_2(\nu;\mathcal H)$. By linearity of $T_\Omega$ and the triangle inequality, if $\|T_\Omega f\|_{\mathcal H}>t$, then either $\|T_\Omega f_{>t}\|_{\mathcal H}>2t/3$ or $\|T_\Omega f_{\le t}\|_{\mathcal H}>t/3$. Set $Y=\|f(S)\|_{\mathcal H}$. Applying~\eqref{eq:product-weak} to $f_{>t}$ at threshold $2t/3$, and~\eqref{eq:product-T2} with Chebyshev's inequality to $f_{\le t}$ at threshold $t/3$, gives
\[
\begin{aligned}
\widehat{\mathbb P}\{\|T_\Omega f\|_{\mathcal H}>t\}&\le\widehat{\mathbb P}\{\|T_\Omega f_{>t}\|_{\mathcal H}>2t/3\}+\widehat{\mathbb P}\{\|T_\Omega f_{\le t}\|_{\mathcal H}>t/3\}\\
&\le\frac{5}{2t/3}\|f_{>t}\|_{L_1(\nu;\mathcal H)}+\frac{1}{(t/3)^2}\|f_{\le t}\|_{L_2(\nu;\mathcal H)}^2\\
&=\frac{15}{2t}\mathbb E\left[Y\mathbf1_{\{Y>t\}}\right]+\frac9{t^2}\mathbb E\left[Y^2\mathbf1_{\{Y\le t\}}\right].
\end{aligned}
\]
For $1<q<2$, the distribution-function identity and the preceding bound give
\[
\begin{aligned}
\|T_\Omega f\|_{L_q(\widehat{\mathcal X};\mathcal H)}^q&=q\int_0^\infty t^{q-1}\widehat{\mathbb P}\{\|T_\Omega f\|_{\mathcal H}>t\}\,\mathrm{d}t\\
&\le\frac{15q}{2}\int_0^\infty t^{q-2}\mathbb E\left[Y\mathbf1_{\{Y>t\}}\right]\,\mathrm{d}t+9q\int_0^\infty t^{q-3}\mathbb E\left[Y^2\mathbf1_{\{Y\le t\}}\right]\,\mathrm{d}t.
\end{aligned}
\]
By Tonelli's theorem,
\[
\begin{aligned}
\int_0^\infty t^{q-2}\mathbb E\left[Y\mathbf1_{\{Y>t\}}\right]\,\mathrm{d}t&=\mathbb E\left[Y\int_0^Yt^{q-2}\,\mathrm{d}t\right]=\frac{\mathbb E Y^q}{q-1}=\frac{\|f\|_{L_q(\nu;\mathcal H)}^q}{q-1},\\
\int_0^\infty t^{q-3}\mathbb E\left[Y^2\mathbf1_{\{Y\le t\}}\right]\,\mathrm{d}t&=\mathbb E\left[Y^2\int_Y^\infty t^{q-3}\,\mathrm{d}t\right]=\frac{\mathbb E Y^q}{2-q}=\frac{\|f\|_{L_q(\nu;\mathcal H)}^q}{2-q}.
\end{aligned}
\]
The second integrand in the expectations is understood to vanish when $Y=0$. The restrictions $q>1$ and $q<2$ ensure convergence at $0$ and $\infty$, respectively. Substitution gives
\[
\|T_\Omega f\|_{L_q(\widehat{\mathcal X};\mathcal H)}^q\le\left(\frac{15q}{2(q-1)}+\frac{9q}{2-q}\right)\|f\|_{L_q(\nu;\mathcal H)}^q.
\]
Taking the $q$th root gives~\eqref{eq:product-Tq}.
\end{proof}

Let $f\in L_2(\nu;\mathcal H)$ and $1<q<2$. Since $\widehat{\mathbb P}$ is a probability measure, we have $\|T_\Omega f\|_{L_1(\widehat{\mathcal X};\mathcal H)}\le\|T_\Omega f\|_{L_q(\widehat{\mathcal X};\mathcal H)}$. On the other hand, the definition of $T_\Omega$ gives
\[
\begin{aligned}
\|T_\Omega f\|_{L_1(\widehat{\mathcal X};\mathcal H)}&=\sum_{(i,j)\in\mathcal I}\theta_{ij}\frac{\|\Omega\|_F}{\gamma_{ij}}\|\mathsf R_{ij}f\|_{L_1(\nu\otimes\nu;\mathcal H)}\\
&=\frac1{\|\Omega\|_F}\sum_{\substack{i,j=1\\i\ne j}}^n\gamma_{ij}\|\mathsf R_{ij}f\|_{L_1(\nu\otimes\nu;\mathcal H)}.
\end{aligned}
\]
Pairs outside $\mathcal I$ with $i\ne j$ contribute zero to the last sum. Multiplying by $\|\Omega\|_F$ and applying~\eqref{eq:product-Tq} yields
\begin{equation}\label{eq:product-R}
\sum_{\substack{i,j=1\\i\ne j}}^n\gamma_{ij}\|\mathsf R_{ij}f\|_{L_1(\nu\otimes\nu;\mathcal H)}\le C_q\|\Omega\|_F\|f\|_{L_q(\nu;\mathcal H)}.
\end{equation}

\begin{lemma}\label{lem:product-K}
For every $f\in L_2(\nu;\mathcal H)$ and every $1<q<2$, we have
\[
\sum_{\substack{i,j=1\\i\ne j}}^n\gamma_{ij}\|K_{ij}f\|_{L_1(\nu\otimes\nu;\mathcal H)}\le2C_q\|\Omega\|_F\|f\|_{L_q(\nu;\mathcal H)},
\]
with $C_q$ as in Lemma~\ref{lem:product-interpolation}.
\end{lemma}
\begin{proof}
We relate $K_{ij}$ to $\mathsf R_{ij}$ using the product noise operators of \citet[Definition~8.27 and Proposition~8.28]{odonnell2014}, restricted to the coordinates other than $i,j$. On functions of $(s,s')$, let each $\mathsf E_k$ in the following product act on coordinate $k$ of $s$ while leaving $s'$ fixed. For every pair of distinct indices $i,j\in[n]$, we claim that
\begin{equation}\label{eq:product-shift}
K_{ij}f=\mathsf R_{ij}f+\int_0^1\prod_{\substack{k=1\\k\ne i,j}}^n\left(\rho\mathrm{Id}+(1-\rho)\mathsf E_k\right)\mathsf R_{ij}f\,\mathrm{d}\rho.
\end{equation}
The product is the identity when $n=2$. To verify~\eqref{eq:product-shift}, consider the Hoeffding decomposition of $f$. Since $\mathsf P_Jf$ does not depend on the coordinates outside $J$, we have $\partial_i\partial_j\mathsf P_Jf=0$ whenever $\{i,j\}\nsubseteq J$. The definitions of the operators therefore give
\[
\mathsf R_{ij}f=\frac12\sum_{\substack{J\subseteq[n]\\\{i,j\}\subseteq J}}\frac{\partial_i\partial_j\mathsf P_Jf}{|J|},\qquad K_{ij}f=\frac12\sum_{\substack{J\subseteq[n]\\\{i,j\}\subseteq J}}\frac{\partial_i\partial_j\mathsf P_Jf}{|J|-1}.
\]
For each $J$ appearing in these sums and every $k\in[n]\setminus\{i,j\}$, the operator $\mathsf E_k$ commutes with $\partial_i\partial_j$. Hence $\mathsf E_k\partial_i\partial_j\mathsf P_Jf=0$ if $k\in J\setminus\{i,j\}$, while $\mathsf E_k\partial_i\partial_j\mathsf P_Jf=\partial_i\partial_j\mathsf P_Jf$ if $k\notin J$. It follows that the factor $\rho\mathrm{Id}+(1-\rho)\mathsf E_k$ multiplies the component by $\rho$ in the first case and by $1$ in the second. There are $|J|-2$ indices of the first type, so
\[
\prod_{\substack{k=1\\k\ne i,j}}^n(\rho\mathrm{Id}+(1-\rho)\mathsf E_k)\partial_i\partial_j\mathsf P_Jf=\rho^{|J|-2}\partial_i\partial_j\mathsf P_Jf.
\]
The right-hand side of~\eqref{eq:product-shift} is consequently equal to
\[
\begin{aligned}
\frac12\sum_{\substack{J\subseteq[n]\\\{i,j\}\subseteq J}}\frac{1}{|J|}\left(1+\int_0^1\rho^{|J|-2}\,\mathrm{d}\rho\right)\partial_i\partial_j\mathsf P_Jf&=\frac12\sum_{\substack{J\subseteq[n]\\\{i,j\}\subseteq J}}\frac1{|J|}\left(1+\frac1{|J|-1}\right)\partial_i\partial_j\mathsf P_Jf\\
&\quad=\frac12\sum_{\substack{J\subseteq[n]\\\{i,j\}\subseteq J}}\frac{\partial_i\partial_j\mathsf P_Jf}{|J|-1}=K_{ij}f.
\end{aligned}
\]
The integrand is a polynomial in $\rho$ with coefficients in $L_2(\nu\otimes\nu;\mathcal H)$, so the integral is well defined there and also in $L_1(\nu\otimes\nu;\mathcal H)$.

For $g\in L_1(\nu\otimes\nu;\mathcal H)$ and each $k\in[n]\setminus\{i,j\}$, Jensen's inequality for averaging over coordinate $k$ of $s$ gives
\[
\begin{aligned}
\|\mathsf E_kg\|_{L_1(\nu\otimes\nu;\mathcal H)}&=\mathbb E\|(\mathsf E_kg)(S,S')\|_{\mathcal H}\\
&\le\mathbb E(\mathsf E_k[\|g(\cdot,\cdot)\|_{\mathcal H}])(S,S')=\|g\|_{L_1(\nu\otimes\nu;\mathcal H)}.
\end{aligned}
\]
For $0\le\rho\le1$, the triangle inequality therefore gives
\[
\begin{aligned}
\|(\rho\mathrm{Id}+(1-\rho)\mathsf E_k)g\|_{L_1(\nu\otimes\nu;\mathcal H)}&\le\rho\|g\|_{L_1(\nu\otimes\nu;\mathcal H)}+(1-\rho)\|\mathsf E_kg\|_{L_1(\nu\otimes\nu;\mathcal H)}\\
&\le\|g\|_{L_1(\nu\otimes\nu;\mathcal H)}.
\end{aligned}
\]
The product in~\eqref{eq:product-shift} is a composition of such contractions. Using the triangle inequality in~\eqref{eq:product-shift} and then the contraction property gives
\[
\begin{aligned}
\|K_{ij}f\|_{L_1(\nu\otimes\nu;\mathcal H)}&\le\|\mathsf R_{ij}f\|_{L_1(\nu\otimes\nu;\mathcal H)}+\int_0^1\left\|\prod_{\substack{k=1\\k\ne i,j}}^n(\rho\mathrm{Id}+(1-\rho)\mathsf E_k)\mathsf R_{ij}f\right\|_{L_1(\nu\otimes\nu;\mathcal H)}\,\mathrm{d}\rho\\
&\le\|\mathsf R_{ij}f\|_{L_1(\nu\otimes\nu;\mathcal H)}+\int_0^1\|\mathsf R_{ij}f\|_{L_1(\nu\otimes\nu;\mathcal H)}\,\mathrm{d}\rho\\
&=2\|\mathsf R_{ij}f\|_{L_1(\nu\otimes\nu;\mathcal H)}.
\end{aligned}
\]
Multiplying by $\gamma_{ij}$, summing over distinct indices, and applying the weighted bound for $\mathsf R_{ij}$ yields, for every $1<q<2$,
\begin{equation}\label{eq:product-K}
\begin{aligned}
\sum_{\substack{i,j=1\\i\ne j}}^n\gamma_{ij}\|K_{ij}f\|_{L_1(\nu\otimes\nu;\mathcal H)}&\le2\sum_{\substack{i,j=1\\i\ne j}}^n\gamma_{ij}\|\mathsf R_{ij}f\|_{L_1(\nu\otimes\nu;\mathcal H)}\\
&\le2C_q\|\Omega\|_F\|f\|_{L_q(\nu;\mathcal H)}.
\end{aligned}
\end{equation}
This gives the stated estimate.
\end{proof}

\begin{proof}[\textnormal{\bfseries Proof of Theorem~\ref{thm:main}}]
Let $h_i$, $i\in[n]$, be the functions in Theorem~\ref{thm:main}, and let $\nu$ be the product law of $S$. If $n=1$, then $h_1(S)=\mathbb E[h_1(S)\mid Z_1]=0$ almost surely. We therefore assume $n\ge2$.

\paragraph{Bounded representatives.} Fix $i\in[n]$. By Bochner Fubini and $\mathbb E[h_i(S)\mid Z_i]=0$, there exists a measurable set $\mathcal Z_i^0\subseteq\mathcal Z_i$ with $\nu_i(\mathcal Z_i^0)=1$ such that for every $z_i\in\mathcal Z_i^0$, the section $y_{-i}\mapsto h_i(z_i,y_{-i})$ is Bochner-integrable and has integral zero. For $x_{-i},y_{-i}\in\mathcal X_{-i}$, replace the coordinates of $x_{-i}$ successively by those of $y_{-i}$. Each replacement in coordinate $j\in[n]\setminus\{i\}$ changes the function value by at most $\gamma_{ij}$. The triangle inequality therefore gives $\|h_i(z_i,x_{-i})-h_i(z_i,y_{-i})\|_{\mathcal H}\le\sum_{\substack{j=1\\j\ne i}}^n\gamma_{ij}$. For every $z_i\in\mathcal Z_i^0$ and every $x_{-i}\in\mathcal X_{-i}$, Jensen's inequality and the centering identity give
\[
\begin{aligned}
\|h_i(z_i,x_{-i})\|_{\mathcal H}&=\left\|\int_{\mathcal X_{-i}}(h_i(z_i,x_{-i})-h_i(z_i,y_{-i}))\,\mathrm{d}\nu_{-i}(y_{-i})\right\|_{\mathcal H}\\
&\le\int_{\mathcal X_{-i}}\|h_i(z_i,x_{-i})-h_i(z_i,y_{-i})\|_{\mathcal H}\,\mathrm{d}\nu_{-i}(y_{-i})\\
&\le\sum_{\substack{j=1\\j\ne i}}^n\gamma_{ij}.
\end{aligned}
\]
Replace $h_i(s)$ by $\mathbf1_{\mathcal Z_i^0}(z_i)h_i(s)$. This modification agrees with the original function almost everywhere, so both centering identities continue to hold almost surely. Changing a coordinate $j\ne i$ leaves the multiplier $\mathbf1_{\mathcal Z_i^0}(z_i)$ unchanged. Thus the sensitivity bound holds for the modified function at every pair of points differing only in coordinate $j$. We continue to write $h_i$ for the modified function. For every $i\in[n]$ and every $z_i\in\mathcal Z_i$, we have
\begin{equation}\sup_{s\in\mathcal X}\|h_i(s)\|_{\mathcal H}\le\sum_{\substack{j=1\\j\ne i}}^n\gamma_{ij},\qquad \int_{\mathcal X_{-i}} h_i(z_i,y_{-i})\,\mathrm{d}\nu_{-i}(y_{-i})=0. \label{eq:bounded-section} \end{equation}
The functions $h_i$ are therefore bounded and strongly measurable, so they belong to $L_p(\nu;\mathcal H)$ for every $1\le p\le\infty$. If $\|\Omega\|_F=0$, they vanish by the first inequality of~\eqref{eq:bounded-section} and the theorem follows, so assume $\|\Omega\|_F>0$.

\paragraph{Reconstruction of the dependent sum.} Fix $i\in[n]$. The condition $\mathbb E[h_i(S)\mid S_{-i}]=0$ gives $\mathsf E_ih_i=0$ and $\mathsf D_ih_i=h_i$. For every $J\subseteq[n]$ with $i\notin J$, the projection $\mathsf P_J$ contains the factor $\mathsf E_i$. Hence $\mathsf P_Jh_i=0$ for these sets. The other centering condition, $\mathbb E[h_i(S)\mid Z_i]=0$, gives $\prod_{\substack{k=1\\k\ne i}}^n\mathsf E_kh_i=0$. Since $\mathsf P_{\{i\}}=\mathsf D_i\prod_{\substack{k=1\\k\ne i}}^n\mathsf E_k$, we also have $\mathsf P_{\{i\}}h_i=0$. The only remaining components are indexed by sets containing $i$ and at least one other index. Each such set $J$ contains exactly $|J|-1$ indices distinct from $i$. Thus
\[
\begin{aligned}
h_i&=\sum_{\substack{J\subseteq[n]\\i\in J,\ |J|\ge2}}\mathsf P_Jh_i=\sum_{\substack{J\subseteq[n]\\i\in J,\ |J|\ge2}}\sum_{j\in J\setminus\{i\}}\frac{\mathsf P_Jh_i}{|J|-1}\\
&=\sum_{\substack{j=1\\j\ne i}}^n\sum_{\substack{J\subseteq[n]\\\{i,j\}\subseteq J}}\frac{\mathsf P_Jh_i}{|J|-1}.
\end{aligned}
\]

We next express the pairing of each component with $f$ in terms of replacement differences of $h_i$. For distinct indices $i,j\in[n]$ and every $g\in L_2(\nu;\mathcal H)$, the function $\partial_jh_i$ is independent of the replacement coordinate $Z'_i$. Averaging $\partial_i\partial_jg$ over $Z'_i$ while holding $S$ and $Z'_j$ fixed gives $(\mathsf D_i\partial_jg)(S,S')$. Moreover, since $i\ne j$, we have $\mathsf E_i\partial_jh_i=\partial_j\mathsf E_ih_i=0$, where $\mathsf E_i$ acts on coordinate $i$ of $s$. By self-adjointness of $\mathsf E_i$,
\[
\begin{aligned}
\mathbb E\langle\partial_jh_i,\partial_i\partial_jg\rangle&=\mathbb E\langle\partial_jh_i,\mathsf D_i\partial_jg\rangle\\
&=\mathbb E\langle\partial_jh_i,\partial_jg\rangle-\mathbb E\langle\mathsf E_i\partial_jh_i,\partial_jg\rangle\\
&=\mathbb E\langle\partial_jh_i,\partial_jg\rangle.
\end{aligned}
\]
Applying~\eqref{eq:product-replacement} in coordinate $j$ gives
\[
\begin{aligned}
\frac12\mathbb E\langle\partial_jh_i,\partial_i\partial_jg\rangle&\mathrel{\mathmakebox[3.6em][c]{=}}\frac12\mathbb E\langle\partial_jh_i,\partial_jg\rangle\\
&\mathrel{\mathmakebox[3.6em][c]{=}}\mathbb E\langle h_i(S),(\mathsf D_jg)(S)\rangle\\
&\mathrel{\mathmakebox[3.6em][c]{\overset{\mathsf D_ih_i=h_i}{=}}}\mathbb E\langle(\mathsf D_ih_i)(S),(\mathsf D_jg)(S)\rangle\\
&\mathrel{\mathmakebox[3.6em][c]{\overset{\mathsf D_i^*=\mathsf D_i}{=}}}\mathbb E\langle h_i(S),(\mathsf D_i\mathsf D_jg)(S)\rangle.
\end{aligned}
\]
For $g=\sum_{\substack{J\subseteq[n]\\|J|\ge2}}\mathsf P_Jf/(|J|-1)$, the identity $\mathsf D_i\mathsf D_j\mathsf P_J=\mathsf P_J$ when $\{i,j\}\subseteq J$, and zero otherwise, gives
\[
\mathbb E\langle\partial_jh_i,K_{ij}f\rangle=\sum_{\substack{J\subseteq[n]\\\{i,j\}\subseteq J}}\frac{1}{|J|-1}\mathbb E\langle h_i(S),(\mathsf P_Jf)(S)\rangle.
\]
Using the reconstruction of $h_i$ and self-adjointness of the Hoeffding projections, we obtain
\begin{equation}\label{eq:product-duality}
\begin{aligned}
\mathbb E\left\langle\sum_{i=1}^nh_i(S),f(S)\right\rangle&=\sum_{i=1}^n\sum_{\substack{j=1\\j\ne i}}^n\sum_{\substack{J\subseteq[n]\\\{i,j\}\subseteq J}}\frac{\mathbb E\langle(\mathsf P_Jh_i)(S),f(S)\rangle}{|J|-1}\\
&=\sum_{\substack{i,j=1\\i\ne j}}^n\sum_{\substack{J\subseteq[n]\\\{i,j\}\subseteq J}}\frac{\mathbb E\langle h_i(S),(\mathsf P_Jf)(S)\rangle}{|J|-1}\\
&=\sum_{\substack{i,j=1\\i\ne j}}^n\mathbb E\langle\partial_jh_i,K_{ij}f\rangle.
\end{aligned}
\end{equation}
The sensitivity assumption gives $\|\partial_jh_i\|_{\mathcal H}=\|h_i(S)-h_i(S^{(j)})\|_{\mathcal H}\le\gamma_{ij}$ for every pair of distinct indices $i,j\in[n]$. Applying Cauchy--Schwarz in $\mathcal H$, followed by this pointwise bound and Lemma~\ref{lem:product-K}, we obtain, for $f\in L_2(\nu;\mathcal H)$ and $1<q\le3/2$,
\[
\begin{aligned}
\left|\mathbb E\left\langle\sum_{i=1}^nh_i(S),f(S)\right\rangle\right|&\le\sum_{\substack{i,j=1\\i\ne j}}^n\mathbb E\left[\|\partial_jh_i\|_{\mathcal H}\|K_{ij}f\|_{\mathcal H}\right]\\
&\le\sum_{\substack{i,j=1\\i\ne j}}^n\gamma_{ij}\|K_{ij}f\|_{L_1(\nu\otimes\nu;\mathcal H)}\\
&\le2C_q\|\Omega\|_F\|f\|_{L_q(\nu;\mathcal H)}.
\end{aligned}
\]
By~\eqref{eq:bounded-section}, the sum $\varphi=\sum_{i=1}^nh_i$ is bounded. For $p\ge3$ and $q=p/(p-1)$, H\"older's inequality gives $|\mathbb E\langle\varphi(S),f(S)\rangle|\le\|\varphi\|_{L_p(\nu;\mathcal H)}\|f\|_{L_q(\nu;\mathcal H)}$. If $\|\varphi\|_{L_p(\nu;\mathcal H)}>0$, equality is attained by the bounded function $f(s)=\|\varphi(s)\|_{\mathcal H}^{p-2}\varphi(s)/\|\varphi\|_{L_p(\nu;\mathcal H)}^{p-1}$, which has $L_q$ norm one because $(p-1)q=p$. If $\varphi=0$ almost everywhere, the desired conclusion is immediate. Thus
\[
\|\varphi\|_{L_p(\nu;\mathcal H)}=\sup_{\substack{f\in L_\infty(\nu;\mathcal H)\\\|f\|_{L_q(\nu;\mathcal H)}\le1}}\mathbb E\langle\varphi(S),f(S)\rangle.
\]
All such functions belong to $L_2(\nu;\mathcal H)$. Taking this supremum in the preceding bound and using $q/(q-1)=p$, $q/(2-q)=p/(p-2)$, and $1/q=(p-1)/p$, we obtain
\[
\begin{aligned}
\|\varphi\|_{L_p(\nu;\mathcal H)}&\le2C_q\|\Omega\|_F=2\left(\frac{15q}{2(q-1)}+\frac{9q}{2-q}\right)^{1/q}\|\Omega\|_F\\
&=2\left(\frac{15p}{2}+\frac{9p}{p-2}\right)^{(p-1)/p}\|\Omega\|_F.
\end{aligned}
\]
To obtain the stated linear dependence on $p$, factor the quantity inside the parentheses as $\frac{15p}{2}\left(1+\frac{6}{5(p-2)}\right)$. Since $1+t\le e^t$ for every $t\ge0$, we have, for every $p\ge3$,
\[
\left(1+\frac{6}{5(p-2)}\right)^{p-1}\le\exp\left(\frac{6(p-1)}{5(p-2)}\right)\le e^{12/5}<\frac{15p}{2}.
\]
Taking the $p$th root of this inequality gives
\[
\begin{aligned}
\left(\frac{15p}{2}+\frac{9p}{p-2}\right)^{(p-1)/p}&=\left(\frac{15p}{2}\right)^{(p-1)/p}\left(1+\frac{6}{5(p-2)}\right)^{(p-1)/p}\\
&\le\left(\frac{15p}{2}\right)^{(p-1)/p}\left(\frac{15p}{2}\right)^{1/p}=\frac{15p}{2}.
\end{aligned}
\]
It follows that $\|\varphi\|_{L_p(\nu;\mathcal H)}\le15p\|\Omega\|_F$ for every $p\ge3$. For $2\le p<3$, monotonicity of $L_p$ norms on probability spaces and the unsimplified bound at $p=3$ give
\[
\begin{aligned}
\|\varphi\|_{L_p(\nu;\mathcal H)}&\le\|\varphi\|_{L_3(\nu;\mathcal H)}\le2\left(\frac{15\cdot3}{2}+\frac{9\cdot3}{3-2}\right)^{2/3}\|\Omega\|_F\\
&=2\left(\frac{99}{2}\right)^{2/3}\|\Omega\|_F<30\|\Omega\|_F\le15p\|\Omega\|_F.
\end{aligned}
\]
The bounded representatives agree almost surely with the original functions, and $\|\varphi(S)\|_{L_p(\mathcal H)}=\|\varphi\|_{L_p(\nu;\mathcal H)}$, so the conclusion holds for the functions in Theorem~\ref{thm:main}.
\end{proof}

\subsection{Consequences of the concentration theorem}\label{sec:concentration-consequences}
\begin{proposition}\label{prop:sharper-L2}
Under the assumptions of Theorem~\ref{thm:main},
\[
\left\|\sum_{i=1}^n h_i(S)\right\|_{L_2(\mathcal H)}\le\|\Omega\|_F.
\]
\end{proposition}
\begin{proof}
We use the Hoeffding decomposition to exploit both centering conditions. First, integrating the bounded differences inequality over the coordinates other than $i$ and using $\mathbb E[h_i(S)\mid Z_i]=0$ gives $\|h_i\|_{L_\infty(\nu;\mathcal H)}\le\sum_{\substack{j=1\\j\ne i}}^n\gamma_{ij}$. In particular, each $h_i$ belongs to $L_2(\nu;\mathcal H)$.

The two centering identities imply $\mathsf E_i h_i=0$ and $\prod_{j\ne i}\mathsf E_j h_i=0$. Thus $\mathsf P_Jh_i=0$ unless $i\in J$ and $|J|\ge2$. Every remaining Hoeffding component involves a coordinate $j\ne i$, and whenever $i\in J$, there are $|J|-1$ such coordinates. Using orthogonality, Cauchy--Schwarz, and $|J|\le2(|J|-1)$ for $|J|\ge2$, we obtain
\[
\begin{aligned}
\left\|\sum_{i=1}^n h_i\right\|_{L_2(\nu;\mathcal H)}^2
&=\sum_{\substack{J\subseteq[n]\\|J|\ge2}}\left\|\sum_{i\in J}\mathsf P_Jh_i\right\|_{L_2(\nu;\mathcal H)}^2\\
&\le\sum_{\substack{J\subseteq[n]\\|J|\ge2}}|J|\sum_{i\in J}\|\mathsf P_Jh_i\|_{L_2(\nu;\mathcal H)}^2\\
&\le2\sum_{\substack{i,j=1\\i\ne j}}^n
\sum_{\substack{J\subseteq[n]\\\{i,j\}\subseteq J}}
\|\mathsf P_Jh_i\|_{L_2(\nu;\mathcal H)}^2.
\end{aligned}
\]
For $i\ne j$, the identity $\mathsf D_jh_i=\sum_{\substack{J\subseteq[n]\\\{i,j\}\subseteq J}}\mathsf P_Jh_i$ and orthogonality identify the last sum with the squared norms of the coordinate differences. Let $S^{(j)}$ be obtained by replacing $Z_j$ with an independent copy. Conditional on $S_{-j}$, the random vectors $h_i(S)$ and $h_i(S^{(j)})$ are independent and identically distributed. Using these identities,~\eqref{eq:product-replacement}, and the sensitivity bounds gives
\[
\begin{aligned}
2\sum_{\substack{i,j=1\\i\ne j}}^n
\sum_{\substack{J\subseteq[n]\\\{i,j\}\subseteq J}}
\|\mathsf P_Jh_i\|_{L_2(\nu;\mathcal H)}^2 
&=2\sum_{\substack{i,j=1\\i\ne j}}^n
\|\mathsf D_jh_i\|_{L_2(\nu;\mathcal H)}^2\\
&=\sum_{\substack{i,j=1\\i\ne j}}^n\mathbb E\|h_i(S)-h_i(S^{(j)})\|_{\mathcal H}^2\le\|\Omega\|_F^2.
\end{aligned}
\]
Taking square roots completes the proof.
\end{proof}

We now extend Theorem~\ref{thm:main} to sums satisfying the weaker centering condition $\mathbb E[g_i(S)\mid S_{-i}]=0$.

\begin{corollary}\label{cor:general-moment}
Let $g_i\colon\mathcal X\to\mathcal H$, $i\in[n]$, be measurable, with $g_i(S)$ Bochner-integrable. Assume that $\|g_i(s)-g_i(\tilde s)\|_{\mathcal H}\le\gamma_{ij}$ whenever $s,\tilde s$ differ only in coordinate $j\ne i$, and let $\Omega=(\gamma_{ij})$ be the sensitivity matrix. Suppose also that, for some $M\ge0$ and every $i\in[n]$,
\begin{equation}\label{eq:main-centering}
\mathbb E[g_i(S)\mid S_{-i}]=0,\qquad
\|\mathbb E[g_i(S)\mid Z_i]\|_{\mathcal H}\le M
\end{equation}
almost surely. Then, for every $p\ge2$,
\begin{equation}\label{eq:general-moment}
\begin{aligned}
\left\|\sum_{i=1}^n g_i(S)\right\|_{L_p(\mathcal H)}
&\le15p\|\Omega\|_F
+\left\|\sum_{i=1}^n\mathbb E[g_i(S)\mid Z_i]\right\|_{L_p(\mathcal H)}\\
&\le15p\|\Omega\|_F+2M\sqrt{pn}.
\end{aligned}
\end{equation}
\end{corollary}

Under the same assumptions, orthogonality gives a sharper bound at $p=2$. Writing $\sigma^2=\sum_{i=1}^n\mathbb E\|\mathbb E[g_i(S)\mid Z_i]\|_{\mathcal H}^2$, we have
\[\left\|\sum_{i=1}^n g_i(S)\right\|_{L_2(\mathcal H)}\le\sqrt{\|\Omega\|_F^2+\sigma^2}.\]
\begin{proof}
For each $i\in[n]$, choose a measurable function $k_i$ such that $k_i(Z_i)=\mathbb E[g_i(S)\mid Z_i]$ almost surely and $\|k_i(z_i)\|_{\mathcal H}\le M$ for every $z_i$. Such a function is obtained by integrating $g_i$ over the coordinates other than $i$, setting the result to zero wherever the integral is undefined or exceeds $M$ in norm. 

Set $h_i(s)=g_i(s)-k_i(z_i)$. For every $j\ne i$, changing coordinate $j$ leaves $k_i(z_i)$ unchanged. Hence $h_i$ satisfies the same sensitivity bound $\gamma_{ij}$ as $g_i$. The definition of $k_i$, the centering condition~\eqref{eq:main-centering}, and independence of $Z_i$ from $S_{-i}$ give
\[
\mathbb E[h_i(S)\mid Z_i]=0, \qquad
\mathbb E[h_i(S)\mid S_{-i}]=0.
\]

Theorem~\ref{thm:main} and Minkowski's inequality give the first bound in~\eqref{eq:general-moment}. The second follows from Lemma~\ref{lem:independent}, since $\|k_i(Z_i)\|_{\mathcal H}\le M$ almost surely.

For the $L_2$ bound, double centering gives $\mathbb E\langle h_i(S),k_j(Z_j)\rangle_{\mathcal H}=0$ for every $i,j\in[n]$, by conditioning on $Z_i$ when $i=j$ and on $S_{-i}$ when $i\ne j$. Independence and centering of the variables $k_i(Z_i)$ give $\|\sum_{i=1}^n k_i(Z_i)\|_{L_2(\mathcal H)}^2=\sigma^2$. Thus, by orthogonality of the dependent and independent components and Proposition~\ref{prop:sharper-L2},
\[\left\|\sum_{i=1}^n g_i(S)\right\|_{L_2(\mathcal H)}^2=\left\|\sum_{i=1}^n h_i(S)\right\|_{L_2(\mathcal H)}^2+\sigma^2\le\|\Omega\|_F^2+\sigma^2.\]
Taking square roots gives the stated $L_2$ bound.
\end{proof}

For the corresponding high-probability estimate, apply Markov's inequality to the dependent component and Lemma~\ref{lem:bernstein} to the independent sum. Their failure probabilities can be bounded by $\delta/e^2$ and $2\delta/e^2$, respectively. Hence, for every $\delta\in(0,1)$, with probability at least $1-\delta$,
\[
\left\|\sum_{i=1}^n g_i(S)\right\|_{\mathcal H}
\le15e\|\Omega\|_F\log(e^2/\delta)
+2\sigma\sqrt{\log(e^2/\delta)}+2M\log(e^2/\delta).
\]
Thus the dependent component is controlled by the sensitivity matrix, while the independent term can be controlled either by its envelope or by its second moment.

We next consider a common sensitivity bound $\gamma\ge0$. Suppose that~\eqref{eq:main-centering} holds and that, for all distinct $i,j\in[n]$ and all samples $s,\tilde s\in\mathcal X$ differing only in coordinate $j$,
\begin{equation}\label{eq:uniform-sensitivity}
\|g_i(s)-g_i(\tilde s)\|_{\mathcal H}\le\gamma.
\end{equation}
The sensitivity matrix then satisfies $\|\Omega\|_F\le\sqrt{n(n-1)}\,\gamma$.

\begin{corollary}\label{cor:uniform-moment}
Under~\eqref{eq:main-centering} and~\eqref{eq:uniform-sensitivity}, for every $p\ge2$,
\[
\left\|\sum_{i=1}^n g_i(S)\right\|_{L_p(\mathcal H)}
\le15p\sqrt{n(n-1)}\,\gamma+2M\sqrt{pn}.
\]
\end{corollary}
\begin{proof}
Apply Corollary~\ref{cor:general-moment} with $\|\Omega\|_F\le\sqrt{n(n-1)}\,\gamma$.
\end{proof}

The simpler upper bound $15pn\gamma+2M\sqrt{pn}$ follows from $\sqrt{n(n-1)}\le n$. Applying Markov's inequality at moment order $\log(e^2/\delta)$ therefore gives, for every $\delta\in(0,1)$, with probability at least $1-\delta$,
\[
\left\|\sum_{i=1}^n g_i(S)\right\|_{\mathcal H}
\le e\left(15n\gamma\log(e^2/\delta)
+2M\sqrt{n\log(e^2/\delta)}\right).
\]
The second-moment bound also gives $\|\sum_{i=1}^n g_i(S)\|_{L_2(\mathcal H)}^2\le n(n-1)\gamma^2+\sigma^2$. Together, these estimates distinguish the dependent component from the sampling error associated with the independent first-order projection in a Hoeffding decomposition~\citep{hoeffding1948}. This is the specialization used for the learning bounds in Section~\ref{sec:statistics}.

For scalar-valued functions, taking $\mathcal H=\mathbb R$ gives $\left\|\sum_{i=1}^n g_i(S)\right\|_{L_p}\le15pn\gamma+2M\sqrt{pn}$ for every $p\ge2$. Under the same centering and sensitivity assumptions, \citet[Theorem~4]{bkz2020} obtained
\[
\left\|\sum_{i=1}^n g_i(S)\right\|_{L_p}\le12\sqrt2\,pn\gamma\lceil\log_2n\rceil+4M\sqrt{pn}.
\]
The Hilbert-space bound removes $\log n$ from the dependent term without changing the order of the independent term. Corollary~\ref{cor:independent-lower} below shows that both terms have the optimal worst-case order throughout $2\le p\le n$.

\section{Sharpness of the concentration bounds}\label{sec:sharpness}
This section gives Rademacher constructions that determine the worst case orders of the moment and tail bounds.

\subsection{Sharpness from Rademacher constructions}\label{subsec:lower-bound}

We show that the factor $p\|\Omega\|_F$ in Theorem~\ref{thm:main} is optimal in the worst case for $2\le p\le n$, even for scalar-valued functions. Throughout this section, Rademacher variables are uniform on $\{-1,1\}$. Our construction uses a quadratic Rademacher chaos, as in the case $M=0$ of \citet[Proposition~9]{bkz2020}. It gives a lower bound of order $\min\{p,n\}\|\Omega\|_F$ for every $p\ge2$. We also establish a corresponding sharpness result for the independent component in Section~\ref{subsec:independent-sharpness}.

\begin{proposition}\label{prop:hilbert-lower}
Let $n\ge2$, $\gamma\ge0$, and let $\mathcal H$ be a nonzero real separable Hilbert space.
For independent Rademacher variables $\varepsilon_1,\ldots,\varepsilon_n$, there exist functions $h_i\colon\{-1,1\}^n\to\mathcal H$, $i\in[n]$, satisfying both conditional centering identities in~\eqref{eq:double-centering} and $\gamma_{ij}=\gamma$ for every $i\ne j$.
For every $p\ge2$, these functions satisfy
\[
\left\|\sum_{i=1}^n h_i(\varepsilon)\right\|_{L_p(\mathcal H)} \ge\frac{\gamma n}{9e^2}\min\{p,n\}.
\]
\end{proposition}

For this construction, $\|\Omega\|_F=\gamma\sqrt{n(n-1)}\le n\gamma$. Hence the proposition also yields a lower bound of $\min\{p,n\}\|\Omega\|_F/(9e^2)$. In particular, the linear dependence on $p$ in Theorem~\ref{thm:main} cannot be improved uniformly over sensitivity matrices throughout $2\le p\le n$.

To prove Proposition~\ref{prop:hilbert-lower}, we first establish a lower bound for the $L_p$ norm of a sum of independent Rademacher variables. 
\begin{lemma}\label{lem:rademacher-lower}
For every integer $m\ge1$, let $\varepsilon_1,\ldots,\varepsilon_m$ be independent Rademacher variables and set $U_m=\sum_{j=1}^m\varepsilon_j$. Then, for every real $p\ge2$,
\[
\|U_m\|_{L_p}\ge\frac{1}{e\sqrt2}\sqrt{m\min\{p,m\}}.
\]
\end{lemma}
\begin{proof}
Fix $m\ge1$ and $p\ge2$, and set $k=\min\{m,\lfloor p/2\rfloor\}$. Then $1\le k\le m$ and $2k\le p$. Since $p\ge2$ implies $\lfloor p/2\rfloor\ge p/4$, we also have $k\ge\min\{p,m\}/4$.

Expand $U_m^{2k}$ into products of $2k$ Rademacher variables. Independence and symmetry imply that the expectation of a product is zero whenever some index occurs an odd number of times. If every index occurs an even number of times, the product equals one. Thus every nonzero term in the moment expansion contributes positively.

Retain the terms involving exactly $k$ distinct indices, each appearing twice. There are $\binom{m}{k}$ choices of indices and $(2k)!/2^k$ arrangements for each choice. Each selected term has expectation one, so $\mathbb E U_m^{2k}\ge\binom{m}{k}(2k)!/2^k$.
For each integer $k$ with $1\le k\le m$, the product formula and the inequalities $\frac{m-r}{k-r}\ge \frac{m}{k}$, valid for every integer $r$ with $0\le r<k$, give
\[
\binom{m}{k}=\prod_{r=0}^{k-1}\frac{m-r}{k-r}\ge\left(\frac{m}{k}\right)^k.
\]
Also, $\sum_{r=1}^{2k}\log r\ge\int_1^{2k}\log x\,\mathrm{d}x$ yields $(2k)!\ge(2k/e)^{2k}$. Consequently,
\[
\mathbb E U_m^{2k}\ge\left(\frac{m}{k}\right)^k\frac{(2k/e)^{2k}}{2^k}=\left(\frac{2mk}{e^2}\right)^k.
\]
By monotonicity of moment norms on a probability space and $k\ge\min\{p,m\}/4$, we obtain
\[
\|U_m\|_{L_p}\ge\|U_m\|_{L_{2k}}\ge\frac{\sqrt{2mk}}{e}\ge\frac{1}{e\sqrt2}\sqrt{m\min\{p,m\}}.
\]
This gives the stated lower bound.
\end{proof}

\begin{proof}[\textnormal{\bfseries Proof of Proposition~\ref{prop:hilbert-lower}}]
Since $\mathcal H$ is nonzero, choose $u\in\mathcal H$ with $\|u\|_{\mathcal H}=1$. For each $i\in[n]$, define
\begin{equation}\label{eq:hilbert-lower-construction} h_i(\varepsilon)=\frac{\gamma}{2}\varepsilon_i\left(\sum_{\substack{j=1\\j\ne i}}^n\varepsilon_j\right)u. \end{equation}

Fix $i\in[n]$. Since the Rademacher coordinates are independent and have mean zero, conditioning on $\varepsilon_i$ gives
\[
\mathbb E[h_i(\varepsilon)\mid\varepsilon_i]=\frac{\gamma}{2}\varepsilon_i\left(\sum_{\substack{j=1\\j\ne i}}^n\mathbb E[\varepsilon_j\mid\varepsilon_i]\right)u=0.
\]
Conditioning instead on $\varepsilon_{-i}$ gives
\[
\mathbb E[h_i(\varepsilon)\mid\varepsilon_{-i}]=\frac{\gamma}{2}\mathbb E[\varepsilon_i\mid\varepsilon_{-i}]\left(\sum_{\substack{j=1\\j\ne i}}^n\varepsilon_j\right)u=0.
\]
For each $j\in[n]$ distinct from $i$, let $\varepsilon^{(j)}$ denote the vector obtained by flipping the $j$th sign. Only one term in the sum defining $h_i$ changes, and therefore $h_i(\varepsilon)-h_i(\varepsilon^{(j)})=\gamma\varepsilon_i\varepsilon_j u$. Its norm is $\gamma$ for every $\varepsilon$, so $\gamma_{ij}=\gamma$ for every pair of distinct indices $i,j\in[n]$, and $\|\Omega\|_F=\gamma\sqrt{n(n-1)}$.

Set $\varphi(\varepsilon)=\sum_{i=1}^n h_i(\varepsilon)$. Each product $\varepsilon_i\varepsilon_j$ with distinct indices occurs twice in the double sum. Since $\varepsilon_i^2=1$,
\[
\varphi(\varepsilon)=\gamma\left(\sum_{1\le i<j\le n}\varepsilon_i\varepsilon_j\right)u=\frac{\gamma}{2}\left[\left(\sum_{i=1}^n\varepsilon_i\right)^2-n\right]u.
\]
Partition $[n]$ into disjoint sets $A$ and $B$ with $a=|A|=\lfloor n/2\rfloor$ and $b=|B|=\lceil n/2\rceil$. Define $U_A=\sum_{i\in A}\varepsilon_i$ and $U_B=\sum_{i\in B}\varepsilon_i$. The two sums are independent because they depend on disjoint collections of independent signs.

Let $T_A$ flip the signs indexed by $A$ and leave those indexed by $B$ unchanged. Then $U_A$ changes to $-U_A$ while $U_B$ is unchanged. Consequently,
\[
\varphi(\varepsilon)=\frac{\gamma}{2}[(U_A+U_B)^2-n]u,\qquad \varphi(T_A\varepsilon)=\frac{\gamma}{2}[(U_B-U_A)^2-n]u.
\]
Taking the difference gives
\begin{equation}\label{eq:lower-sign-projection} \frac{\varphi(\varepsilon)-\varphi(T_A\varepsilon)}{2}=\gamma U_AU_B u. \end{equation}

The transformation $T_A$ preserves the joint Rademacher distribution. Hence $\varphi(\varepsilon)$ and $\varphi(T_A\varepsilon)$ have equal $L_p(\mathcal H)$ norms. Minkowski's inequality yields
\[
\begin{aligned}
\left\|\frac{\varphi(\varepsilon)-\varphi(T_A\varepsilon)}{2}\right\|_{L_p(\mathcal H)}
&\le\frac12(\|\varphi(\varepsilon)\|_{L_p(\mathcal H)}+\|\varphi(T_A\varepsilon)\|_{L_p(\mathcal H)})\\
&=\|\varphi(\varepsilon)\|_{L_p(\mathcal H)}.
\end{aligned}
\]
On the other hand, by \eqref{eq:lower-sign-projection}, the independence of $U_A$ and $U_B$, and $\|u\|_{\mathcal H}=1$,
\[
\left\|\frac{\varphi(\varepsilon)-\varphi(T_A\varepsilon)}{2}\right\|_{L_p(\mathcal H)}=\gamma(\mathbb E|U_A|^p\mathbb E|U_B|^p)^{1/p}=\gamma\|U_A\|_{L_p}\|U_B\|_{L_p}.
\]
Combining the two estimates and applying Lemma~\ref{lem:rademacher-lower} to each block,
\[
\|\varphi(\varepsilon)\|_{L_p(\mathcal H)}\ge\gamma\|U_A\|_{L_p}\|U_B\|_{L_p}\ge\frac{\gamma}{2e^2}\sqrt{ab\min\{p,a\}\min\{p,b\}}.
\]
For every $1\le r\le n$, we have $\min\{p,r\}\ge(r/n)\min\{p,n\}$. To see this, if $p\ge n$ both sides equal $r$; if $p<n$, the inequality follows by considering $r\le p$ and $r>p$ separately. Applying it with $r=a$ and $r=b$ gives
\[
\|\varphi(\varepsilon)\|_{L_p(\mathcal H)}\ge\frac{\gamma ab}{2e^2n}\min\{p,n\}.
\]
Finally, if $n$ is even, $ab=n^2/4\ge2n^2/9$. If $n$ is odd, then $n\ge3$ and $ab=(n^2-1)/4\ge2n^2/9$. Therefore $ab\ge2n^2/9$ for every $n\ge2$. Since $\|\Omega\|_F=\gamma\sqrt{n(n-1)}\le\gamma n$, we obtain
\[
\|\varphi(\varepsilon)\|_{L_p(\mathcal H)}\ge\frac{\gamma n}{9e^2}\min\{p,n\}\ge\frac{\min\{p,n\}}{9e^2}\|\Omega\|_F.
\]
Thus the construction attains the claimed lower bound.
\end{proof}

\begin{remark}\label{rem:lower-scope}
Every admissible family in Theorem~\ref{thm:main} satisfies
\begin{equation}\label{eq:lower-envelope} \left\|\sum_{i=1}^n h_i(S)\right\|_{L_\infty(\mathcal H)} \le\sum_{\substack{i,j=1\\i\ne j}}^n\gamma_{ij} \le\sqrt{n(n-1)}\|\Omega\|_F. \end{equation}
Indeed, for almost every fixed $s_i$, conditional centering gives
\[
h_i(s_i,s_{-i}) =\int_{\mathcal X_{-i}}[h_i(s_i,s_{-i})-h_i(s_i,y_{-i})] \,\mathrm{d}\nu_{-i}(y_{-i}),
\]
and replacing coordinates successively bounds its norm by $\sum_{\substack{j=1\\j\ne i}}^n\gamma_{ij}$. Summation and Cauchy--Schwarz give
\eqref{eq:lower-envelope}. A two sided bound with explicit quantifiers is stated in Corollary~\ref{cor:sharp-order}.

The sharpness assertion concerns the worst case over sensitivity matrices and does not give a $p\|\Omega\|_F$ lower bound for every prescribed matrix. For example, if only $\gamma_{12}$ is nonzero, \eqref{eq:lower-envelope} already bounds every admissible sum by $\gamma_{12}=\|\Omega\|_F$, independently of $p$ and $n$.
\end{remark}

\begin{proposition}\label{prop:matrix-lower}
For every nonnegative matrix $\Omega$ with zero diagonal and every nonzero real separable Hilbert space $\mathcal H$, there exists a Rademacher family satisfying the centering conditions of Theorem~\ref{thm:main}, with sensitivity matrix $\Omega$, such that
\[
\left\|\sum_{i=1}^n h_i(\varepsilon)\right\|_{L_2(\mathcal H)}
\ge\frac12\|\Omega\|_F.
\]
\end{proposition}
\begin{proof}
For a fixed unit vector $u$, set $h_i(\varepsilon)=(1/2)\varepsilon_i\sum_{\substack{j=1\\j\ne i}}^n\gamma_{ij}\varepsilon_j u$. The centering identities and sensitivity equalities follow as above. Since distinct quadratic Rademacher monomials are orthogonal in $L_2$,
\[
\left\|\sum_{i=1}^n h_i(\varepsilon)\right\|_{L_2(\mathcal H)}^2 =\frac14\sum_{1\le i<j\le n}(\gamma_{ij}+\gamma_{ji})^2 \ge\frac14\sum_{1\le i<j\le n}(\gamma_{ij}^2+\gamma_{ji}^2) =\frac14\|\Omega\|_F^2.
\]
Thus every prescribed matrix requires the Frobenius scale in the second moment, whereas linear growth with $p$ occurs only in the worst case.
\end{proof}

Fix a nonzero real separable Hilbert space $\mathcal H$. For integers $n\ge2$, real $p\ge2$, and $\gamma\ge0$, let $\mathfrak R_{n,p}(\gamma)$ be the supremum of $\left\|\sum_{i=1}^n h_i(S)\right\|_{L_p(\mathcal H)}$ over independent coordinate laws and measurable families satisfying the centering assumptions of Theorem~\ref{thm:main}, with $\gamma_{ij}\le\gamma$ for all distinct $i,j\in[n]$.

\begin{corollary}\label{cor:sharp-order}
For every integer $n\ge2$, every $\gamma\ge0$, and every real $p\ge2$,
\[
\frac{\gamma n\min\{p,n\}}{9e^2}
\le\mathfrak R_{n,p}(\gamma)
\le15\gamma n\min\{p,n\}.
\]
\end{corollary}
\begin{proof}
For $2\le p\le n$, Theorem~\ref{thm:main} bounds $\mathfrak R_{n,p}(\gamma)$ by $15pn\gamma$. For $p\ge n$, the deterministic estimate~\eqref{eq:lower-envelope} gives $\mathfrak R_{n,p}(\gamma)\le n(n-1)\gamma\le n^2\gamma$. Proposition~\ref{prop:hilbert-lower} supplies the lower bound for every $p\ge2$.
\end{proof}

Thus the worst-case order is $\Theta\!\left(\gamma n\min\{p,n\}\right)$, uniformly for $n\ge2$, $p\ge2$, and $\gamma>0$. For $\gamma=0$, the supremum vanishes.

For a prescribed nonnegative sensitivity matrix $\Omega$ with zero diagonal, let $\mathfrak R_2(\Omega)$ denote the corresponding supremum at $p=2$. Propositions~\ref{prop:sharper-L2} and~\ref{prop:matrix-lower} give
\[
\frac12\|\Omega\|_F\le\mathfrak R_2(\Omega)\le\|\Omega\|_F.
\]
Both comparisons concern suprema over admissible families; they are not lower bounds for every such family.

For the family in~\eqref{eq:hilbert-lower-construction}, write $\varphi(\varepsilon)=\sum_{i=1}^n h_i(\varepsilon)$. Its sensitivity matrix satisfies $\|\Omega\|_F=\gamma\sqrt{n(n-1)}$.

\begin{corollary}\label{cor:tail-lower}
For every real $t\in[1,n]$,
\begin{equation}\label{eq:tail-lower}
\mathbb P\left\{
\|\varphi(\varepsilon)\|_{\mathcal H}\ge\frac{t}{18}\|\Omega\|_F
\right\}
\ge\frac12\exp\left(-4t\log24\right).
\end{equation}
\end{corollary}
\begin{proof}
The construction in Proposition~\ref{prop:hilbert-lower} expresses the
quadratic Rademacher sum in terms of two independent sums over a balanced
partition. We first obtain a lower-tail estimate for their product and then
transfer it to $\varphi$ using the sign-flip identity~\eqref{eq:lower-sign-projection}.

We begin with a one-sided estimate for an ordinary Rademacher sum. Recall
that $U_m$ denotes the sum of $m\ge1$ independent Rademacher signs. Its
second and fourth moments satisfy
\[
\mathbb E U_m^2=m,\qquad
\mathbb E U_m^4=3m^2-2m\le3m^2.
\]
The Paley--Zygmund inequality~\citep{boucheron2013}, applied to $U_m^2$,
gives
\[
\mathbb P\left\{U_m^2\ge\frac m2\right\}
\ge\frac{(\mathbb E U_m^2)^2}{4\mathbb E U_m^4}
\ge\frac1{12}.
\]
By symmetry, the positive and negative deviations have equal probability.
Consequently,
\[
\mathbb P\left\{U_m\ge\sqrt{m/2}\right\}\ge\frac1{24}.
\]

To obtain an exponential tail at a larger threshold, partition the $m$
signs into $k$ groups of sizes differing by at most one, for an arbitrary integer $k$ with $1\le k\le m$. Each group contains at least $\lfloor m/k\rfloor\ge m/(2k)$ signs.
The preceding estimate applies independently to each group sum. If every
group sum is at least the square root of half its group size, then their
total is at least $k\sqrt{m/(4k)}=\frac12\sqrt{mk}$. The intersection has
probability at least $24^{-k}$, so
\begin{equation}\label{eq:sign-block-tail}
\mathbb P\left\{U_m\ge\frac12\sqrt{mk}\right\}\ge24^{-k}.
\end{equation}

We now apply this estimate to the two halves of the construction. Recall
the partition $A\cup B=[n]$ from Proposition~\ref{prop:hilbert-lower}, with
$a=|A|=\lfloor n/2\rfloor$ and $b=|B|=\lceil n/2\rceil$. The sums
$U_A=\sum_{i\in A}\varepsilon_i$ and $U_B=\sum_{i\in B}\varepsilon_i$ are
independent. For a fixed $t\in[1,n]$, choose the numbers of groups as
\[
k_A=\min\{a,\lceil t\rceil\},\qquad
k_B=\min\{b,\lceil t\rceil\}.
\]
Moreover, $k_A\ge at/n$: if $k_A=a$ this follows from $t\le n$,
and otherwise $k_A=\lceil t\rceil\ge t\ge at/n$.
The same argument applies to $k_B$. Since $t\ge1$, we therefore have
\[
k_A\ge\frac{at}{n},\qquad
k_B\ge\frac{bt}{n},\qquad
k_A+k_B\le2\lceil t\rceil\le4t.
\]
Applying~\eqref{eq:sign-block-tail} to the two independent sums gives
\[
\mathbb P\left\{
U_A\ge\frac12\sqrt{ak_A},\quad U_B\ge\frac12\sqrt{bk_B}
\right\}
\ge24^{-(k_A+k_B)}\ge24^{-4t}.
\]
On this event both sums are positive, and the bounds on $k_A$ and $k_B$
yield
\[
\begin{aligned}
\gamma U_AU_B
&\ge\frac\gamma4\sqrt{abk_Ak_B}
\ge\frac{\gamma ab}{4n}t \ge\frac{\gamma n}{18}t
\ge\frac{t}{18}\|\Omega\|_F.
\end{aligned}
\]
Here we used $ab\ge2n^2/9$, as verified in the proof of
Proposition~\ref{prop:hilbert-lower}, and
$\|\Omega\|_F=\gamma\sqrt{n(n-1)}\le\gamma n$.

It remains to transfer this deviation to the original sum. Let $T_A$
flip the signs in $A$, as in Proposition~\ref{prop:hilbert-lower}. By
\eqref{eq:lower-sign-projection} and the triangle inequality,
\[
\gamma|U_AU_B|
=\left\|\frac{\varphi(\varepsilon)-\varphi(T_A\varepsilon)}2\right\|_{\mathcal H} \le\max\left\{ \|\varphi(\varepsilon)\|_{\mathcal H}, \|\varphi(T_A\varepsilon)\|_{\mathcal H}
\right\}.
\]
The sign flip preserves the Rademacher distribution. Therefore, for every
$v\ge0$, the union bound gives
\[
\mathbb P\left\{\|\varphi(\varepsilon)\|_{\mathcal H}\ge v\right\}
\ge\frac12\mathbb P\left\{\gamma|U_AU_B|\ge v\right\}.
\]
Taking $v=t\|\Omega\|_F/18$ and using the preceding product estimate
proves~\eqref{eq:tail-lower}.
\end{proof}

\subsection{Sharpness of the independent component}\label{subsec:independent-sharpness}

\begin{corollary}
\label{cor:independent-lower} Let $n\ge2$, $\gamma,M\ge0$, and $\mathcal H\ne\{0\}$. There is a single Rademacher family $g_i$ satisfying \eqref{eq:main-centering} and the uniform sensitivity condition~\eqref{eq:uniform-sensitivity} such that, for every real $p\ge2$,
\[
\left\|\sum_{i=1}^n g_i(\varepsilon)\right\|_{L_p(\mathcal H)} \ge \frac{1}{18e^2} \left[\gamma n\min\{p,n\} +M\sqrt{n\min\{p,n\}}\right].
\]
\end{corollary}
\begin{proof}
Use the construction in \eqref{eq:hilbert-lower-construction}, with its unit vector $u$, and set $g_i(\varepsilon)=h_i(\varepsilon)+M\varepsilon_i u$. Its conditional mean given $\varepsilon_i$ is $M\varepsilon_i u$, and its conditional mean given $\varepsilon_{-i}$ is zero. The sensitivities remain equal to $\gamma$. Recall that $U_n=\sum_{i=1}^n\varepsilon_i$, and write $\psi(\varepsilon)=\sum_{i=1}^n g_i(\varepsilon)=\varphi(\varepsilon)+MU_nu$. Since $\varphi(-\varepsilon)=\varphi(\varepsilon)$ and $U_n(-\varepsilon)=-U_n(\varepsilon)$,
\[
\frac{\psi(\varepsilon)+\psi(-\varepsilon)}{2}=\varphi(\varepsilon), \qquad \frac{\psi(\varepsilon)-\psi(-\varepsilon)}{2}=MU_nu.
\]
The random vectors $\varepsilon$ and $-\varepsilon$ have the same distribution, so Minkowski's inequality bounds the $L_p(\mathcal H)$ norm of each expression on the left by $\|\psi(\varepsilon)\|_{L_p(\mathcal H)}$. Therefore
\[
\|\psi(\varepsilon)\|_{L_p(\mathcal H)} \ge\max\{\|\varphi(\varepsilon)\|_{L_p(\mathcal H)},M\|U_n\|_{L_p}\} \ge\frac12(\|\varphi(\varepsilon)\|_{L_p(\mathcal H)}+M\|U_n\|_{L_p}).
\]
Combining this estimate with Proposition~\ref{prop:hilbert-lower} and Lemma~\ref{lem:rademacher-lower} gives the claimed lower bound, since $1/\left(2e\sqrt2\right)\ge1/(18e^2)$.
\end{proof}

For $2\le p\le n$, this construction matches the order $pn\gamma+M\sqrt{pn}$ in Corollary~\ref{cor:uniform-moment}.

\subsection{Sharpness for uniformly stable learning algorithms}\label{subsec:learning-sharpness}
The preceding lower bounds concern sums of dependent functions. We now show that both the stability and sampling terms are necessary in generalization bounds for learning algorithms with bounded losses. Earlier lower bounds for stable learning algorithms establish their orders at constant confidence~\citep{liu2020tight}, while the Rademacher construction of \citet[Proposition~9]{bkz2020} exhibits the dependence on the moment parameter for a sum of dependent functions. We construct a single learning problem whose generalization gap has a lower tail of order $\min\{M,\gamma p+M\sqrt{p/n}\}$ at probability at least $e^{-p}$ for every $p\ge 1$.
In the main construction, we count training observations with value $1$ in a fixed binary coordinate. The amount by which this count exceeds a fixed threshold determines the stability term to the generalization gap.

\begin{proposition}\label{prop:learning-lower}
For every integer $n\ge1$, every $M>0$, and every $\gamma\ge0$, there exist a finite observation space $\mathcal Z$ with a law $\mathrm P$, a hypothesis set $\mathcal W$, a loss $\ell:\mathcal W\times\mathcal Z\to[0,M]$, and a deterministic uniformly $\gamma$-stable learning algorithm $A:\mathcal Z^n\to\mathcal W$ such that the following holds for $S\sim\mathrm P^n$ and every real $p\ge1$:
\begin{equation}\label{eq:learning-lower-tail}
\mathbb P\left\{R(A(S))-R_{\text{emp}}(A(S))\ge\frac1{256}\min\left\{M,\gamma p+M\sqrt{\frac pn}\right\}\right\}\ge e^{-p}.
\end{equation}
\end{proposition}

The construction below has $|\mathcal Z|\le2^n$; the observation and hypothesis spaces may depend on $n$, and no computational restriction is imposed on the algorithm.

We use the block estimate~\eqref{eq:sign-block-tail} for moderate and large deviations of a Rademacher sum. The next lemma provides a complementary probability bound for $p$ close to $1$.

\begin{lemma}\label{lem:learning-rademacher-tail}
Let $\varepsilon_1,\ldots,\varepsilon_n$ be independent Rademacher variables and $U_n=\sum_{i=1}^n\varepsilon_i$. If $n\ge256$, then for every real $1\le p\le n$,
\[
\mathbb P\left\{U_n\ge\frac{\sqrt{np}}{64}\right\}\ge
\begin{cases}
2/5,&1\le p<16,\\
e^{-p/2},&16\le p\le n.
\end{cases}
\]
\end{lemma}
\begin{proof}
We first bound the largest point mass of a Rademacher sum. As above, let $U_m$ denote a sum of $m$ independent Rademacher variables, with $U_0=0$. For an even number of signs, the largest mass occurs at zero, and we claim that, for every integer $m\ge0$,
\[
\mathbb P\{U_{2m}=0\}=\frac1{4^m}\binom{2m}{m}\le\frac1{\sqrt{2m+1}}.
\]
The bound is an equality when $m=0$. If it holds at $m$, then the ratio of consecutive central binomial masses and the identity $(2m+1)(2m+3)=(2m+2)^2-1$ give
\[
\begin{aligned}
\mathbb P\{U_{2m+2}=0\}
&=\frac{2m+1}{2m+2}\mathbb P\{U_{2m}=0\}\\
&\le\frac{\sqrt{2m+1}}{2m+2}
\le\frac1{\sqrt{2m+3}}.
\end{aligned}
\]
This proves the claim by induction.
For an odd number of signs, the largest masses occur at $1$ and $-1$, and
\[
\begin{aligned}
\mathbb P\{U_{2m+1}=1\}
&=\frac1{2^{2m+1}}\binom{2m+1}{m+1}\\
&=\frac{2m+1}{2m+2}\mathbb P\{U_{2m}=0\}
\le\frac1{\sqrt{2m+1}}.
\end{aligned}
\]
Thus, for both parities, every point mass of $U_n$ is at most $1/\sqrt n$.

We use this estimate when $1\le p<16$. By symmetry, for every $t>0$,
\[
\mathbb P\{U_n\ge t\}
=\frac12-\frac12\mathbb P\{U_n=0\}-\mathbb P\{0<U_n<t\}.
\]
The interval $[0,t)$ contains at most $t/2+1$ possible values of $U_n$, because consecutive values are spaced by two. The point-mass bound therefore yields
\[
\mathbb P\{U_n\ge t\}\ge\frac12-\frac{t/2+1}{\sqrt n}.
\]
Taking $t=\sqrt{np}/64$ and using $p<16$ and $n\ge256$, we obtain
\[
\mathbb P\left\{U_n\ge\frac{\sqrt{np}}{64}\right\}
\ge\frac12-\frac{\sqrt p}{128}-\frac1{\sqrt n}
\ge\frac{13}{32}>\frac25.
\]

For $16\le p\le n$, set $b=\lceil p/16\rceil$, the number of blocks in~\eqref{eq:sign-block-tail}. Then $p/16\le b\le p/8$ and $b\le n$, so the block estimate applies. Moreover, $\frac12\sqrt{nb}\ge\sqrt{np}/8$. Consequently,
\[
\begin{aligned}
\mathbb P\left\{U_n\ge\frac{\sqrt{np}}{64}\right\}
&\ge\mathbb P\left\{U_n\ge\frac12\sqrt{nb}\right\}\\
&\ge24^{-b}\ge\exp\left(-\frac{p\log24}{8}\right)
\ge e^{-p/2},
\end{aligned}
\]
since $\log24<4$.
\end{proof}

\begin{proof}[\textnormal{\bfseries Proof of Proposition~\ref{prop:learning-lower}}]
For each observation space $\mathcal Z$ constructed below, take $\mathcal W=\{f:\mathcal Z\to[0,M]\}$ and the evaluation loss $\ell(f,z)=f(z)$. We define $f_s\in\mathcal W$ for every sample $s\in\mathcal Z^n$ and set $A(s)=f_s$. Then
\[
R(A(S))-R_{\text{emp}}(A(S))
=\mathbb E_Z f_S(Z)-\frac1n\sum_{i=1}^n f_S(Z_i).
\]
Thus it suffices to keep $f_s$ in $[0,M]$, bound its change under replacement of one observation, and establish the claimed tail estimate.

If $\gamma>M$, we may carry out the construction with $M$ in place of $\gamma$: its stability is still at most $\gamma$, and the minimum in~\eqref{eq:learning-lower-tail} is unchanged for every $p\ge1$. We therefore assume $0\le\gamma\le M$.

\paragraph{The case $n<256$.}
For $n<256$, let $\mathcal Z=\{0,1\}$, choose $\mathrm P(\{1\})=1-e^{-1/n}$, and take $f_s(z)=Mz$ independently of the sample. With probability $e^{-1}$ all training observations equal zero, and on this event
\[
R(A(S))-R_{\text{emp}}(A(S))=M(1-e^{-1/n})\ge\frac{M}{n+1}\ge\frac{M}{256}.
\]
The inequality $1-e^{-1/n}\ge1/(n+1)$ follows from $e^{1/n}\ge1+1/n$. Since the algorithm has stability zero, the threshold in~\eqref{eq:learning-lower-tail} is at most $M/256$ and $e^{-1}\ge e^{-p}$ for every $p\ge1$.

\paragraph{Construction and stability for $n\ge256$.}
The construction uses pairwise independent binary coordinates obtained from linear forms over the field with two elements~\citep[Section~2.4]{luby2005}.
Let $k=\lceil3n/4\rceil$ be the threshold for the number of positive observations in a coordinate,
and let $Y\sim\operatorname{Bin}(n,1/2)$ represent that count.
Define $\beta=\mathbb P\{Y\ge k+1\}$, the probability that one coordinate exceeds the threshold.
We choose $N=\lfloor1/(16\beta)\rfloor$ coordinates, so that the expected number of threshold exceedances is of constant order.
Since $n\ge256$, we have $k+1\le n$ and hence $\beta>0$.
Hoeffding's inequality, together with $k+1-n/2\ge n/4$, gives
\[
\begin{aligned}
\beta
&=\mathbb P\left\{Y-\frac n2\ge k+1-\frac n2\right\}\\
&\le\exp\left(-\frac{2(k+1-n/2)^2}{n}\right)
\le e^{-n/8}\le\frac1{32}.
\end{aligned}
\]
Consequently, $N\beta\le1/16$ and $N\beta\ge1/16-\beta\ge1/32$. Moreover, $N\ge e^{n/8}/16-1\ge n$ for $n\ge256$. These bounds will control the probabilities of the exceptional events and ensure that the training sample contains at most half the points of the observation space.

Let $d=\lceil\log_2(N+1)\rceil$ and let $\mathbb F_2$ denote the field with two elements. Since $d\le N\le2^d-1$, choose distinct nonzero vectors $a_1,\ldots,a_N\in\mathbb F_2^d$ containing the standard basis. We use the uniform distribution on
\[
\mathcal Z=\{-1,1\}\times\left\{\bigl((-1)^{\langle a_j,u\rangle}\bigr)_{j=1}^N:u\in\mathbb F_2^d\right\},
\]
where $\langle a_j,u\rangle$ is the dot product over $\mathbb F_2$. The map from $u$ to the vector of signs is injective because the vectors $a_j$ contain the standard basis, so $|\mathcal Z|=2^{d+1}$. Since $\beta\ge\mathbb P\{Y=n\}=2^{-n}$, we have $N\le2^{n-4}$ and hence $|\mathcal Z|\le4(N+1)\le2^n$. Distinct nonzero vectors over $\mathbb F_2$ are linearly independent in pairs. Thus each binary coordinate is unbiased, and any two are independent under the uniform law.

For each real number $x$, write $x_+=\max\{x,0\}$. Write each observation as $z=(\varepsilon,x)$, where $x\in\{-1,1\}^N$ and $\varepsilon\in\{-1,1\}$ are independent. For a sample $s=(z_1,\ldots,z_n)$ with $z_i=(\varepsilon_i,x_i)$, define the number of positive observations in coordinate $j\in[N]$ by $\mathsf N_j(s)=\sum_{i=1}^n\mathbf1_{\{x_{ij}=1\}}$. Set
\[
h_s(x)=\max_{j\in[N]}
\left\{\min\left\{\frac M8,\frac\gamma2\bigl(\mathsf N_j(s)-k\bigr)_+\right\}
\mathbf1_{\{x_j=1\}}\right\}.
\]
Thus $h_s(x)=0$ if $\mathsf N_j(s)\le k$ for every $j\in[N]$, and $0\le h_s(x)\le M/8$ for every $s\in\mathcal Z^n$ and $x\in\{-1,1\}^N$. Let $\mathcal O_s=\{z_1,\ldots,z_n\}$ be the set of distinct observations in $s$. The function returned by the algorithm is
\begin{equation}\label{eq:learning-lower-predictor}
f_s(\varepsilon,x)=\frac M2-h_s(x)
-\frac\gamma8\mathbf1_{\{(\varepsilon,x)\in\mathcal O_s\}}
-\frac M4\varepsilon.
\end{equation}
The first subtracted term uses coordinates whose empirical counts exceed $k$. The indicator $\mathbf1_{\{(\varepsilon,x)\in\mathcal O_s\}}$ decreases the loss at observed points, while the independent sign produces a centered sampling term.

The bounds $0\le h_s(x)\le M/8$, $0\le(\gamma/8)\mathbf1_{\{z\in\mathcal O_s\}}\le M/8$, and $-M/4\le-(M/4)\varepsilon\le M/4$ show that $0\le f_s(z)\le3M/4$ for every $s$ and $z$. If $s$ and $\tilde s$ are neighboring samples, then $|\mathsf N_j(s)-\mathsf N_j(\tilde s)|\le1$ for every $j\in[N]$. Taking the positive part, truncating at $M/8$, and taking a finite maximum preserve the bound $\gamma/2$ on the change, so $|h_s(x)-h_{\tilde s}(x)|\le\gamma/2$ for every $x\in\{-1,1\}^N$. Furthermore, the indicators of $\mathcal O_s$ and $\mathcal O_{\tilde s}$ differ pointwise by at most one. Hence
\[
\sup_{z\in\mathcal Z}|f_s(z)-f_{\tilde s}(z)|
\le\frac\gamma2+\frac\gamma8=\frac{5\gamma}{8}\le\gamma.
\]

\paragraph{Decomposition of the generalization gap.}
For a random training sample $S=((\varepsilon_i,X_i))_{i=1}^n$, recall $U_n=\sum_{i=1}^n\varepsilon_i$. Every training observation belongs to $\mathcal O_S$, whereas an independent observation belongs to this set with probability $|\mathcal O_S|/2^{d+1}\le1/2$, since $|\mathcal O_S|\le n\le N<2^d$. Taking population and empirical means in~\eqref{eq:learning-lower-predictor} gives
\begin{equation}\label{eq:learning-lower-gap}
\begin{aligned}
R(A(S))-R_{\text{emp}}(A(S))
&=\frac1n\sum_{i=1}^n h_S(X_i)-\mathbb E_X h_S(X)+\frac\gamma8\left(1-\frac{|\mathcal O_S|}{2^{d+1}}\right)+\frac{M}{4n}U_n\\
&\ge\frac1n\sum_{i=1}^n h_S(X_i)-\mathbb E_X h_S(X) +\frac\gamma{16}+\frac{M}{4n}U_n.
\end{aligned}
\end{equation}
The difference involving $h_S$ depends only on $X_1,\ldots,X_n$, whereas $U_n$ depends only on the independent signs.

\paragraph{Probability bounds for large coordinate counts.}
Let $\mathcal E_0$ be the event that $\mathsf N_j(S)\le k$ for every $j\in[N]$. For each integer $r$ with $1\le r\le\lceil n/16\rceil$, let $\mathcal E_r$ be the event that exactly one index $j^*\in[N]$ satisfies $\mathsf N_{j^*}(S)\ge k+r$, while $\mathsf N_j(S)\le k$ for every $j\ne j^*$. The counts $\mathsf N_j(S)$ are pairwise independent and each has distribution $\operatorname{Bin}(n,1/2)$, since the observations are independent and each pair of binary coordinates is independent. A union bound therefore gives $\mathbb P(\mathcal E_0)\ge1-N\beta\ge15/16$.

The excess $r$ above the threshold determines the size of the term involving $h_S$, so we bound the probability of $\mathcal E_r$ from below. For every integer $m$ with $0\le m<n$, the consecutive binomial masses satisfy
\[
\frac{\mathbb P\{Y=m+1\}}{\mathbb P\{Y=m\}}=\frac{n-m}{m+1}.
\]
For $m\ge k+1$, this ratio is at most $1/3$ because $k\ge3n/4$. Summing the geometric upper bound on the tail yields
\[
\beta\le\mathbb P\{Y=k+1\}\sum_{j=0}^{\infty}\frac{1}{3^j}
=\frac32\mathbb P\{Y=k+1\}.
\]
In the shorter range $k+1\le m<k+r$, the ratio is also bounded below. Indeed, $k+r\le13n/16+2$ and $n\ge256$ give
\[
\frac{n-m}{m+1}\ge\frac{3n/16-2}{13n/16+3}\ge\frac18.
\]
There are $r-1$ ratios between the masses at $k+1$ and $k+r$. Multiplying these bounds gives
\[
\begin{aligned}
\mathbb P\{Y\ge k+r\}
&\ge\mathbb P\{Y=k+r\}\\
&\ge\mathbb P\{Y=k+1\}\frac{1}{8^{r-1}}
\ge\frac{2\beta}{3\cdot8^{r-1}}.
\end{aligned}
\]
For $r=1$, the product has no factors and the same inequality holds.
For each fixed $r$, the events corresponding to different exceptional indices are disjoint. Subtracting the events on which another count exceeds $k$ and using pairwise independence gives
\[
\begin{aligned}
\mathbb P(\mathcal E_r)
&\ge\sum_{j=1}^N\left(\mathbb P\{\mathsf N_j(S)\ge k+r\}-\sum_{\substack{l=1\\l\ne j}}^N\mathbb P\{\mathsf N_j(S)\ge k+r,\ \mathsf N_l(S)>k\}\right)\\
&=N\,\mathbb P\{Y\ge k+r\}\bigl(1-(N-1)\beta\bigr).
\end{aligned}
\]
Since $N\beta\ge1/32$ and $1-(N-1)\beta\ge1-N\beta\ge15/16$, we obtain
\begin{equation}\label{eq:learning-lower-events}
\mathbb P(\mathcal E_r)\ge\frac{15}{16}N\,\mathbb P\{Y\ge k+r\}\ge\frac1{64\cdot8^{r-1}}.
\end{equation}

On $\mathcal E_0$, all counts are at most $k$, so $h_S=0$. On $\mathcal E_r$, only coordinate $j^*$ can contribute to $h_S$. Its population frequency is $1/2$, while its empirical frequency is at least $3/4$. Thus
\begin{equation}\label{eq:learning-lower-feature}
\begin{aligned}
\frac1n\sum_{i=1}^n h_S(X_i)-\mathbb E_X h_S(X)
&=\min\left\{\frac M8,\frac\gamma2\bigl(\mathsf N_{j^*}(S)-k\bigr)\right\}
\left(\frac{\mathsf N_{j^*}(S)}n-\frac12\right)\\
&\ge\frac14\min\left\{\frac M8,\frac{\gamma r}{2}\right\}.
\end{aligned}
\end{equation}

\paragraph{Lower-tail bound for all $p\ge1$.}
For $1\le p<16$, use the event $\mathcal E_0\cap\{U_n\ge\sqrt{np}/64\}$. Independence and Lemma~\ref{lem:learning-rademacher-tail} give probability at least $(15/16)(2/5)=3/8>e^{-p}$. On this event, $h_S=0$, and~\eqref{eq:learning-lower-gap} gives
\[
R(A(S))-R_{\text{emp}}(A(S))
\ge\frac\gamma{16}+\frac M{256}\sqrt{\frac pn}
\ge\frac1{256}\left(\gamma p+M\sqrt{\frac pn}\right).
\]

For $16\le p\le n$, take $r=\lceil p/16\rceil$. By~\eqref{eq:learning-lower-events} and $p\ge16$,
\[
\mathbb P(\mathcal E_r)\ge\frac1{64\cdot8^{r-1}}
\ge\frac1{64}\exp\left(-\frac{p\log8}{16}\right)
\ge e^{-p/2}.
\]
The independent event $\{U_n\ge\sqrt{np}/64\}$ also has probability at least $e^{-p/2}$ by Lemma~\ref{lem:learning-rademacher-tail}. Hence their intersection has probability at least $e^{-p}$. Because $r\ge p/16$, equations~\eqref{eq:learning-lower-gap} and~\eqref{eq:learning-lower-feature} imply on this event that
\[
\begin{aligned}
R(A(S))-R_{\text{emp}}(A(S))
&\ge\frac1{128}\min\{M,\gamma p\}+\frac M{256}\sqrt{\frac pn}\\
&\ge\frac1{256}\min\left\{M,\gamma p+M\sqrt{\frac pn}\right\}.
\end{aligned}
\]
For $p>n$, the event just obtained with $p=n$ has probability at least $e^{-n}\ge e^{-p}$ and gives a gap of at least $M/256$, which suffices because the minimum in~\eqref{eq:learning-lower-tail} equals $M$.

The distribution, loss, and algorithm were chosen independently of $p$; only the events used in the argument vary with $p$.
\end{proof}

The construction also gives a lower bound in any nonzero real Hilbert space $\mathcal H$. Fix $u\in\mathcal H$ with $\|u\|_{\mathcal H}=1$ and set $F(s,z)=\ell(A(s),z)u$. Its generalization gap equals $\bigl(R(A(S))-R_{\text{emp}}(A(S))\bigr)u$, so~\eqref{eq:learning-lower-tail} implies the same lower-tail bound for its norm.

\section{Applications to statistical learning}\label{sec:applications}
Starting from~\eqref{eq:vector-reduction}, we apply Theorem~\ref{thm:main} to the doubly centered component and Bernstein's inequality to the independent component.

Throughout this section, let $S=(Z_1,\ldots,Z_n)\sim\mathrm{P}^n$ be the training sample on $\mathcal Z^n$. Let $Z\sim\mathrm{P}$ denote an independent test observation. A deterministic learning algorithm returns $A(S)\in\mathcal W$, while a randomized algorithm returns $A(S,\xi)\in\mathcal W$, with an independent random seed $\xi\sim\mathrm{Q}$. All mappings and Hilbert space random variables are assumed measurable. Unless stated otherwise, probability bounds are with respect to the training sample $S$, with confidence parameter $\delta\in(0,1)$, and expectations over $\xi$ are conditional on $S$. When comparing our bounds with those from prior work, we take $n\ge2$ and $\delta\le1/e$; then $\log(1/\delta)\le\log(e^2/\delta)\le3\log(1/\delta)$.

\subsection{Concentration under uniform stability}\label{sec:statistics}
We apply Theorem~\ref{thm:main} to the Hilbert-valued generalization gap introduced in Section~\ref{sec:introduction}. Let $\mathcal H$ be a real separable Hilbert space, and let $F:\mathcal Z^n\times\mathcal Z\to\mathcal H$ be jointly measurable. Recall that
\[
\Delta(S)=\mathbb E_ZF(S,Z) -\frac{1}{n}\sum_{i=1}^nF(S,Z_i).
\]
The expectation over $Z$ is taken with $S$ fixed. We assume that the function is uniformly bounded and allow the stability bound to depend on the coordinate being replaced. For constants $M,\gamma_1,\ldots,\gamma_n\ge0$, assume
\begin{equation}\label{eq:statistic-assumptions}
\|F(s,z)\|_{\mathcal H}\le M,\qquad \|F(s,z)-F(\tilde s,z)\|_{\mathcal H}\le\gamma_j
\end{equation}
for every $z\in\mathcal Z$ and every pair of samples $s,\tilde s\in\mathcal X$ differing only in coordinate $j$. Define $\gamma_{\mathrm{rms}}=\left(\frac{1}{n}\sum_{j=1}^n\gamma_j^2\right)^{1/2}$. In particular, $\gamma_{\mathrm{rms}}=\gamma$ when $\gamma_j=\gamma$ for every $j$.

\citet[Lemma~7 and Corollary~8]{bkz2020} reduce scalar generalization to a sum of conditionally centered functions with bounded coordinate sensitivities. They estimate the dependent part by a telescoping decomposition over nested coordinate partitions; summing over the levels creates a factor $\log n$. \citet[Theorem~1]{fanlei2024} extend this argument to Hilbert spaces and retain a separate sensitivity $\gamma_j$ for each coordinate. Their applications to losses and gradients give, with probability at least $1-\delta$,
\[
\|\Delta(S)\|_{\mathcal H}\lesssim\gamma_{\mathrm{rms}}\log n\log(1/\delta)+M\sqrt{\frac{\log(1/\delta)}{n}}.
\]

The preceding bound allows different stability parameters for individual observations, but its stability term still contains a factor $\log n$, and its sampling term depends only on the uniform bound $M$. We improve both terms by applying Theorem~\ref{thm:main} directly to the dependent component, using its Frobenius norm estimate to remove the factor $\log n$. For the independent component, Bernstein's inequality uses the variance of the mean function. A bounded differences argument then controls this variance by $\Var_Z(F(S,Z))$.

\begin{corollary}\label{cor:statistic}
Suppose \eqref{eq:statistic-assumptions} holds. Then, for every $\delta\in(0,1)$, with probability at least $1-\delta$,
\[
\|\Delta(S)\|_{\mathcal H}
\le 84\gamma_{\mathrm{rms}}\log(e^2/\delta)
+2\sqrt{\frac{\log(e^2/\delta)}{n}\Var_Z(F(S,Z))}
+\frac{4M\log(e^2/\delta)}{n}.
\]
\end{corollary}

Under the assumptions of Corollary~\ref{cor:statistic}, the same decomposition and $L_2$ orthogonality give, for every $p\ge2$,
\begin{equation}\label{eq:statistic-moment}
\begin{aligned}
\|\Delta(S)\|_{L_2(\mathcal H)} &\le 2\gamma_{\mathrm{rms}}+\sqrt{\frac{4(n-1)\gamma_{\mathrm{rms}}^2}{n}+\frac{\mathbb E_S\Var_Z(F(S,Z))}{n}},\\
\|\Delta(S)\|_{L_p(\mathcal H)} &\le 31p\gamma_{\mathrm{rms}}+4M\sqrt{\frac{p}{n}}.
\end{aligned}
\end{equation}
The $L_2$ bound also holds with $\mathbb E_S\Var_Z(F(S,Z))$ replaced by $M^2$.

Since the population variance is at most $M^2$, the sampling terms are at most $6M\sqrt{\log(e^2/\delta)/n}$ whenever $\log(e^2/\delta)\le n$. Retaining the variance improves this estimate when it is much smaller than $M^2$, up to the term of order $M\log(e^2/\delta)/n$. We prove Corollary~\ref{cor:statistic} in Appendix~\ref{app:statistics}.

\subsection{Randomized algorithms}\label{subsec:randomization}
For randomized learning algorithms, the generalization gap also depends on the random seed. Averaging over the seed before taking the norm can lead to cancellation and does not control the second moment of the gap. We therefore apply Corollary~\ref{cor:statistic} in the Hilbert space of square-integrable functions of the seed to bound the conditional root mean square of the generalization gap. 

Let $\mathcal K$ be a real separable Hilbert space, and let $\Upsilon(S,z,\xi)$ take values in $\mathcal K$, with the independent seed $\xi\sim\mathrm{Q}$ introduced above. We regard $\Upsilon(s,z,\cdot)$ as a function of the random seed and define $F(s,z)=\Upsilon(s,z,\cdot)$, taking values in $L_2(\mathrm{Q};\mathcal K)$. Assume that $F$ is strongly measurable and takes values in a separable subspace of $L_2(\mathrm{Q};\mathcal K)$, with $\|F(s,z)\|_{L_2(\mathrm{Q};\mathcal K)}\le M$ for all $s,z$. For every pair of neighboring samples $s,\tilde s$, suppose that
\begin{equation}\label{eq:mean-square-stability}
\sup_{z\in\mathcal Z}(\mathbb E_\xi\|\Upsilon(s,z,\xi)-\Upsilon(\tilde s,z,\xi)\|_{\mathcal K}^2)^{1/2}\le\gamma.
\end{equation}
Here, the same random seed is used for both samples. By the definition of the $L_2$ norm, this condition is equivalent to the uniform stability assumption for $F$ in $L_2(\mathrm{Q};\mathcal K)$.

By Bochner Fubini, the gap $\Delta(S)$ in $L_2(\mathrm{Q};\mathcal K)$ has norm
\[
\|\Delta(S)\|_{L_2(\mathrm{Q};\mathcal K)}=\left(\mathbb E_\xi\left\|\mathbb E_Z\Upsilon(S,Z,\xi)-\frac1n\sum_{i=1}^n\Upsilon(S,Z_i,\xi)\right\|_{\mathcal K}^2\right)^{1/2}.
\]
Fubini's theorem also gives $\Var_Z(F(S,Z))=\mathbb E_\xi\Var_Z(\Upsilon(S,Z,\xi))$. Applying Corollary~\ref{cor:statistic} in the closure of the separable subspace specified above gives, with probability at least $1-\delta$,
\begin{equation}\label{eq:random-statistic}
\|\Delta(S)\|_{L_2(\mathrm{Q};\mathcal K)}
\le84\gamma\log(e^2/\delta)
+2\sqrt{\frac{\log(e^2/\delta)}{n}\mathbb E_\xi\Var_Z(\Upsilon(S,Z,\xi))}
+\frac{4M\log(e^2/\delta)}{n}.
\end{equation}

For bounded scalar losses, \citet[Theorem~1]{yuanli2023}
established, under $L_2$-uniform stability, the following
bound with probability at least $1-\delta$ over $S$:
\[
\mathbb E_\xi|\Delta(S,\xi)| \lesssim\gamma\log n\log(1/\delta) +M\sqrt{\frac{\log(1/\delta)}{n}}.
\]
In comparison,~\eqref{eq:random-statistic} removes the factor $\log n$ from the stability term, extends the result to Hilbert-valued statistics, and uses the population variance averaged over the seed in the sampling term.

The same application of Corollary~\ref{cor:statistic} also yields, for every $p\ge2$,
\[
\|\Delta\|_{L_p(\mathrm{P}^n;L_2(\mathrm{Q};\mathcal K))}\le31p\gamma+4M\sqrt{\frac{p}{n}}.
\]
Both bounds control the conditional root mean square generalization gap over the random seed, rather than the gap for each individual seed. They also control the conditional expected norm of that gap, by Cauchy--Schwarz. When the stability parameters vary across coordinates, the same bounds use $\gamma_{\mathrm{rms}}$ in place of $\gamma$.

\subsection{Loss generalization and excess risk}\label{sec:loss-applications}
\subsubsection{Generalization of bounded losses}\label{sec:bkz}
Let $\ell:\mathcal W\times\mathcal Z\to[0,M]$ be measurable, and let $A(S,\xi)\in\mathcal W$ be the output of a learning algorithm. Taking $\Upsilon(S,z,\xi)=\ell(A(S,\xi),z)$, the generalization gap becomes $\Delta(S,\xi)=R(A(S,\xi))-R_{\text{emp}}(A(S,\xi))$, where $\Delta(S,\cdot)$ denotes the $L_2(\mathrm{Q})$-valued generalization gap defined in Subsection~\ref{subsec:randomization}. Assume that, for every pair of neighboring samples $s,\tilde s$,
\begin{equation}\label{eq:loss-mean-square-stability}
\sup_{z\in\mathcal Z}(\mathbb E_\xi|\ell(A(s,\xi),z)-\ell(A(\tilde s,\xi),z)|^2)^{1/2}\le\gamma.
\end{equation}
For deterministic algorithms, this reduces to the usual uniform stability condition for the loss, and we omit $\xi$.

The scalar concentration argument of \citet[Corollary~8]{bkz2020} gives a sampling term of order $M/\sqrt n$, up to the confidence factor. \citet[Section~3.3, Eq.~(12)]{klochkov2021} refine that argument by applying Bernstein's inequality to the independent component and localizing its second moment through a weakly self-bounding concentration inequality. For deterministic algorithms, they obtain
\[
\Delta(S)\lesssim\gamma\log n\log(1/\delta)+\sqrt{\frac{MR(A(S))\log(1/\delta)}{n}}+\frac{M\log(1/\delta)}{n}
\]
with probability at least $1-\delta$.

Applying~\eqref{eq:random-statistic} with $\Upsilon(S,z,\xi)=\ell(A(S,\xi),z)$ gives the following bound.

\begin{corollary}\label{cor:uniform}
Suppose $0\le\ell(w,z)\le M$ for every $w\in\mathcal W$ and $z\in\mathcal Z$, and assume the measurability conditions of Subsection~\ref{subsec:randomization} and the mean square loss-stability condition~\eqref{eq:loss-mean-square-stability}. Then, for every $\delta\in(0,1)$, with probability at least $1-\delta$,
\[
(\mathbb E_\xi|\Delta(S,\xi)|^2)^{1/2}\le84\gamma\log(e^2/\delta)+2\sqrt{\frac{\log(e^2/\delta)}{n}\mathbb E_\xi\Var_Z(\ell(A(S,\xi),Z))}+\frac{4M\log(e^2/\delta)}{n}.
\]
\end{corollary}

Since $0\le\ell\le M$, for every $w\in\mathcal W$ we have $\Var_Z(\ell(w,Z))\le R(w)[M-R(w)]\le MR(w)$ and $\Var_Z(\ell(w,Z))\le M^2/4$. Substituting the first bound in Corollary~\ref{cor:uniform} gives a sampling term in terms of the population risk. The second gives a bound of order $\gamma\log(e^2/\delta)+M\sqrt{\log(e^2/\delta)/n}$ when $\log(e^2/\delta)\le n$; otherwise, $|\Delta(S,\xi)|\le M$ gives the same order.

For deterministic algorithms, the conditional root mean square reduces to $|\Delta(S)|$. If $\gamma=O(1/n)$ and $R(A(S))=O(1/n)$ on the event under consideration, the bound is $O(\log(e^2/\delta)/n)$ for fixed $M$. For randomized algorithms, the same conclusion holds with $\mathbb E_\xi R(A(S,\xi))$ in place of $R(A(S))$.

We next determine the worst-case order for bounded losses and deterministic uniformly stable algorithms by combining the moment bound of Corollary~\ref{cor:statistic} with Proposition~\ref{prop:learning-lower}. The supremum below ranges over distributions $\mathrm P$, measurable losses $\ell:\mathcal W\times\mathcal Z\to[0,M]$, and deterministic uniformly $\gamma$-stable learning algorithms $A:\mathcal Z^n\to\mathcal W$. The observation and hypothesis spaces may vary, and each moment is taken under the corresponding sample law $\mathrm P^n$.

\begin{corollary}\label{cor:learning-sharp-order}
For all integers $n\ge1$ and real parameters $M>0$, $\gamma\ge0$, and $p\ge1$,
\begin{equation}\label{eq:learning-sharp-moment}
\sup_{\mathrm P,\ell,A}\|R(A(S))-R_{\text{emp}}(A(S))\|_{L_p}
\asymp\min\left\{M,\gamma p+M\sqrt{\frac pn}\right\},
\end{equation}
with universal implied constants.
\end{corollary}

Markov's inequality and Proposition~\ref{prop:learning-lower} also give matching upper and positive lower tails at probability scale $e^{-p}$, with one learning problem attaining the lower bounds for all $p\ge1$; see Appendix~\ref{app:learning-sharp-order}.

\subsubsection{Excess risk under a generalized Bernstein condition}\label{sec:generalized-bernstein}
A small excess risk need not imply a small loss variance, for example in binary classification with label noise. The generalized Bernstein condition instead controls the variance of an appropriate excess loss by the excess risk.

Let $A:\mathcal X\to\mathcal W$ be deterministic and uniformly $\gamma$-stable. Suppose the set $\mathcal W^\star=\arg\min_{w\in\mathcal W}R(w)$ is nonempty, and write $R^\star=\min_{w\in\mathcal W}R(w)$. Define the empirical optimization error by
\[
\Delta_{\mathrm{opt}}(S)
=R_{\text{emp}}(A(S))-\inf_{w\in\mathcal W}R_{\text{emp}}(w).
\]
Assume that $\Delta_{\mathrm{opt}}(S)$ is measurable. Let $M,B\ge0$ and assume $0\le\ell(w,z)\le M$ for every $w\in\mathcal W$ and $z\in\mathcal Z$. We use a centered form of the generalized Bernstein condition of \citet[Assumption~1.1]{klochkov2021}: for each $w\in\mathcal W$, there exists $u\in\mathcal W^\star$ such that
\begin{equation}\label{eq:generalized-bernstein}
\Var_Z(\ell(w,Z)-\ell(u,Z))\le B[R(w)-R^\star].
\end{equation}
The minimizer $u$ may depend on $w$. The uncentered second-moment condition in that reference implies~\eqref{eq:generalized-bernstein}. We also assume that $s\mapsto\ell(A(s),\cdot)$ is strongly measurable as an $L_2(\mathrm P)$-valued map. This condition holds when $\mathcal Z$ is standard Borel and $(s,z)\mapsto\ell(A(s),z)$ is jointly measurable and bounded, because $L_2(\mathrm P)$ is then separable.

\citet[Theorem~3.3]{bartlett2005} use a relation between variance and expected excess loss to localize Rademacher complexity through a complexity fixed point. \citet[Theorem~1.1]{klochkov2021} instead use uniform stability and the corresponding uncentered second-moment condition. For every fixed $\eta>0$, they prove, with probability at least $1-\delta$,
\[
\begin{aligned}
R(A(S))-R^\star
&\lesssim\Delta_{\mathrm{opt}}(S)+\eta\mathbb E_S\Delta_{\mathrm{opt}}(S)+(1+1/\eta)\left(\gamma\log n+\frac{M+B}{n}\right)\log(1/\delta).
\end{aligned}
\]
The resulting fast rate depends on algorithmic stability rather than on a complexity of the entire loss class, but retains $\log n$ in the stability term.

Condition~\eqref{eq:generalized-bernstein} allows the population minimizer to depend on the algorithm output, so the corresponding excess loss need not inherit uniform stability.

Instead, set $f_S=\ell(A(S),\cdot)$ and let $\bar f=\mathbb E_Sf_S\in L_2(\mathrm P)$. Let $\mathcal C$ be the closed convex hull in $L_2(\mathrm P)$ of $\{\ell(u,\cdot):u\in\mathcal W^\star\}$, and let $\tilde f$ be the metric projection of $\bar f$ onto $\mathcal C$. The function $\tilde f$ is deterministic, with $\mathbb E_Z\tilde f(Z)=R^\star$ and $\frac1n\sum_{i=1}^n\tilde f(Z_i)\ge\inf_{w\in\mathcal W}R_{\text{emp}}(w)$ almost surely. The projection argument in Appendix~\ref{app:generalized-bernstein-risk} gives
\[
\Var_Z(\bar f(Z)-\tilde f(Z))
\le B\left(\mathbb E_S\Delta_{\mathrm{opt}}(S)+\gamma\right).
\]
We apply Theorem~\ref{thm:main} to the centered gap of $f_S-\bar f$ and Bernstein's inequality to the independent average associated with $\bar f-\tilde f$.

\begin{corollary}\label{cor:generalized-bernstein-risk}
Under the uniform stability, boundedness, generalized Bernstein,
and measurability assumptions stated above, for every
$\delta\in(0,1)$, with probability at least $1-\delta$, 
\begin{equation}\label{eq:generalized-bernstein-exact}
\begin{aligned}
R(A(S))-R^\star
&\le\Delta_{\mathrm{opt}}(S)+84\gamma\log(e^2/\delta)\\
&\quad+2\sqrt{\frac{B\log(e^2/\delta)}{n}
\left(\mathbb E_S\Delta_{\mathrm{opt}}(S)+\gamma\right)}
+\frac{4M\log(e^2/\delta)}{n}.
\end{aligned}
\end{equation}
\end{corollary}

Under the same assumptions, the proof also gives a bound involving only the realized optimization error. On the same event of probability at least $1-\delta$, simultaneously for every $0<\eta<1$,
\begin{equation}\label{eq:generalized-bernstein-realized}
R(A(S))-R^\star\le\frac{\Delta_{\mathrm{opt}}(S)+\left(84\gamma+(4M+B/\eta)/n\right)\log(e^2/\delta)}{1-\eta}.
\end{equation}

To compare with \citet[Theorem~1.1]{klochkov2021}, use subadditivity of the square root and Young's inequality~\citep[Theorem~61]{hardylittlewoodpolya1952} in~\eqref{eq:generalized-bernstein-exact}. The square-root term is at most $\eta\mathbb E_S\Delta_{\mathrm{opt}}(S)+B\log(e^2/\delta)/(\eta n)+\gamma\log(e^2/\delta)+B/n$. Hence, simultaneously for every $\eta>0$,
\[
\begin{aligned}
R(A(S))-R^\star
&\le\Delta_{\mathrm{opt}}(S)+\eta\mathbb E_S\Delta_{\mathrm{opt}}(S)+86(1+1/\eta)\left(\gamma+\frac{M+B}{n}\right)\log(e^2/\delta).
\end{aligned}
\]
Compared with \citet[Theorem~1.1]{klochkov2021}, our displayed bound preserves both $\Delta_{\mathrm{opt}}(S)$ and $\eta\mathbb E_S\Delta_{\mathrm{opt}}(S)$, but replaces $\gamma\log n$ by $\gamma$. It retains the order $(M+B)/n$ of the remaining statistical term. The comparison applies under the earlier uncentered Bernstein condition; our result also allows the centered condition~\eqref{eq:generalized-bernstein}, with an additional $L_2(\mathrm P)$ measurability assumption in the projection argument. If the algorithm minimizes the empirical risk exactly for every sample, both optimization errors vanish, and choosing $\eta=1$ gives an excess risk bound of order $(\gamma+(M+B)/n)\log(e^2/\delta)$. For an approximate minimizer,~\eqref{eq:generalized-bernstein-realized} instead depends only on the realized optimization error, with its coefficient increased to $1/(1-\eta)$. Neither form uniformly improves the other. Subsection~\ref{subsec:approximate} gives a separate argument under empirical quadratic growth, without a population Bernstein condition. The projection argument and the proofs of both bounds are given in Appendix~\ref{app:generalized-bernstein-risk}.

For randomized algorithms, define $\Delta_{\mathrm{opt}}(S,\xi)=R_{\text{emp}}(A(S,\xi))-\inf_{w\in\mathcal W}R_{\text{emp}}(w)$. The same conclusions hold with $R(A(S))-R^\star$ and $\Delta_{\mathrm{opt}}(S)$ replaced by the conditional expectations of $R(A(S,\xi))-R^\star$ and $\Delta_{\mathrm{opt}}(S,\xi)$ over $\xi$, provided the mean loss $f_S=\mathbb E_\xi\ell(A(S,\xi),\cdot)$ is strongly measurable in $L_2(\mathrm P)$ and uniformly $\gamma$-stable. This condition concerns the mean loss and does not require mean square stability. The proof uses convexity of squared distance to the same closed convex comparator set.

\subsection{Gradient generalization and excess risk}\label{sec:gradient}
We use the gradient generalization bound to control population gradients and, under gradient domination, excess risk.

\subsubsection{Gradient generalization}\label{subsec:gradient-generalization}
Let $\mathcal W\subseteq\mathbb R^d$ be nonempty, closed, and convex, and let $A(S,\xi)\in\mathcal W$. Suppose each loss is differentiable on a neighborhood of $\mathcal W$, both risks are differentiable, and differentiation commutes with population expectation. Thus $\nabla R(w)=\mathbb E_Z\nabla \ell(w,Z)$ and $\nabla R_{\text{emp}}(w)=(1/n)\sum_{i=1}^n\nabla \ell(w,Z_i)$. Assume that $R^\star=\min_{w\in\mathcal W}R(w)$ is attained and that $\|\nabla\ell(A(s,\xi),z)\|_2\le L_\ell$ for every $s,\xi,z$. The gradient envelope is required only at algorithm outputs. For a deterministic algorithm, uniform gradient stability means that, for neighboring $s,\tilde s$,
\[
\sup_{z\in\mathcal Z}\|\nabla\ell(A(s),z)-\nabla\ell(A(\tilde s),z)\|_2\le\gamma.
\]
For the randomized algorithm, assume the measurability conditions of Subsection~\ref{subsec:randomization} with $\Upsilon(s,z,\xi)=\nabla\ell(A(s,\xi),z)$ and that, for every pair of neighboring samples $s,\tilde s$,
\[
\sup_{z\in\mathcal Z}(\mathbb E_\xi\|\nabla \ell(A(s,\xi),z)-\nabla \ell(A(\tilde s,\xi),z)\|_2^2)^{1/2}\le\gamma.
\]
For deterministic algorithms, the condition reduces to uniform gradient stability. When the stability bound depends on the coordinate being replaced, we use $\gamma_j$ and $\gamma_{\mathrm{rms}}$ as defined in Section~\ref{sec:statistics}. 

For deterministic learning algorithms, \citet[Theorem~3]{fanlei2024} establish a gradient generalization bound independent of the parameter dimension, with a sampling term of order $L_\ell\sqrt{\log(e/\delta)/n}$ and a stability term of order $\gamma_{\mathrm{rms}}\log n\log(e/\delta)$. For randomized algorithms, \citet[Definition~2 and Theorem~2]{zhu2025} assume mean square gradient stability and bound the expected norm of the gradient generalization gap over the random seed. With probability at least $1-\delta$, their estimate is
\[
\begin{aligned}
\mathbb E_\xi\|\nabla R(A(S,\xi))-\nabla R_{\text{emp}}(A(S,\xi))\|_2
&\lesssim\gamma\log n\log(e/\delta)\\
&\quad+\sqrt{\frac{\log(e/\delta)}{n}\mathbb E_{\xi,Z}\|\nabla \ell(A(S,\xi),Z)\|_2^2}
+\frac{L_\ell\log(e/\delta)}{n}.
\end{aligned}
\]
Their proof of Lemma~13 in Appendix~C.3 also treats the conditional second moment, with a stability term of order $\gamma^2\log^2 n\log^2(1/\delta)$ at the squared scale. Applying~\eqref{eq:random-statistic} with $\Upsilon(S,z,\xi)=\nabla\ell(A(S,\xi),z)$ gives the following conditional root mean square bound, with the population variance in place of the raw second moment and without $\log n$ in the stability term. With probability at least $1-\delta$,
\begin{equation}\label{eq:gradient-refined} \begin{aligned} &(\mathbb E_\xi\|\nabla R(A(S,\xi))-\nabla R_{\text{emp}}(A(S,\xi))\|_2^2)^{1/2}\\ &\quad\le84\gamma\log(e^2/\delta)+2\sqrt{\frac{\log(e^2/\delta)}{n}\mathbb E_\xi\Var_Z(\nabla\ell(A(S,\xi),Z))}+\frac{4L_\ell\log(e^2/\delta)}{n}. \end{aligned} \end{equation}
The bound also holds with $\gamma$ replaced by $\gamma_{\mathrm{rms}}$ when the stability parameters vary across observations. In particular, if $\gamma_{\mathrm{rms}}=O(1/n)$, the stability term is of order $O(1/n)$ at fixed confidence, improving upon the $O(\log n/n)$ term in the preceding bound. We next give conditions that control this variance through the population gradient or the excess risk. 

\paragraph{Bounds on the population gradient.} 
For constants $\kappa\ge0$ and $\sigma_g\ge0$, suppose that at every output $w=A(s,\xi)$,
\begin{equation}\label{eq:gradient-norm-growth}
\Var_Z(\nabla\ell(w,Z))\le\kappa\|\nabla R(w)\|_2^2+\sigma_g^2.
\end{equation}
The case $\kappa=0$ gives a uniform variance bound~\citep{ghadimi2013}. By the variance identity, the case $\sigma_g=0$ is equivalent to the strong growth condition with constant $1+\kappa$ at the outputs~\citep{vaswani2019}.

The triangle inequality and~\eqref{eq:gradient-refined} give the following bound with probability at least $1-\delta$ whenever $n\ge16\kappa\log(e^2/\delta)$:
\begin{equation}\label{eq:strong-growth}
\begin{aligned}
\left(\mathbb E_\xi\|\nabla R(A(S,\xi))\|_2^2\right)^{1/2}
&\le2\left(\mathbb E_\xi\|\nabla R_{\text{emp}}(A(S,\xi))\|_2^2\right)^{1/2}+168\gamma\log(e^2/\delta)\\
&\quad+4\sigma_g\sqrt{\frac{\log(e^2/\delta)}n}+\frac{8L_\ell\log(e^2/\delta)}n.
\end{aligned}
\end{equation}
Using the gradient envelope in~\eqref{eq:gradient-refined} instead gives a bound on the conditional mean norm without a sample size restriction. With probability at least $1-\delta$, simultaneously for every $\theta>0$,
\begin{equation}\label{eq:strong-growth-mean}
\begin{aligned}
\mathbb E_\xi\|\nabla R(A(S,\xi))\|_2
&\le(1+\theta)\mathbb E_\xi\|\nabla R_{\text{emp}}(A(S,\xi))\|_2+84(1+\theta)\gamma\log(e^2/\delta)\\
&\quad+2(1+\theta)\sigma_g\sqrt{\frac{\log(e^2/\delta)}n}+\left(4(1+\theta)+\frac{\kappa(1+\theta)^2}{\theta}\right)\frac{L_\ell\log(e^2/\delta)}n.
\end{aligned}
\end{equation}
Both bounds hold on the event of~\eqref{eq:gradient-refined} when their respective sample size conditions are satisfied. For fixed $\theta>0$, structural constants, and confidence, if $\sigma_g=0$ and $\gamma=O(1/n)$, their statistical terms are of order $1/n$, whereas a fixed positive $\sigma_g$ leaves a term of order $1/\sqrt n$. In the strong growth case,~\eqref{eq:strong-growth-mean} removes $\log n$ from the stability term in \citet[Proposition~1]{zhu2025} without imposing a sample size restriction. We give the proofs in Appendix~\ref{app:strong-growth} and the comparison in Table~\ref{tab:gradient-comparison}.

\subsubsection{Excess risk under gradient domination}\label{subsec:gradient-risk}
To obtain an excess risk bound, we combine a condition on the stochastic gradient variance with a condition on the geometry of the population risk. The first controls the sampling term in~\eqref{eq:gradient-refined}; the second converts stationarity into excess risk. For constants $v_0,v_1\ge0$, assume that, at every output $w=A(s,\xi)$,
\begin{equation}\label{eq:gradient-growth}
\Var_Z(\nabla\ell(w,Z))\le v_0+v_1[R(w)-R^\star].
\end{equation}
This condition is weaker than the corresponding bound on the raw second moment, which includes the squared norm of the population gradient~\citep{vaswani2019,gower2019}.

For the geometric condition, we use the stationarity residual $\mathcal R_R(w)$ from Section~\ref{subsec:learning-notation}. Unlike the gradient norm, this residual vanishes at a constrained minimizer. For some $\mu>0$, assume that the following inequality holds at every output $w=A(s,\xi)$:
\begin{equation}\label{eq:constrained-pl}
R(w)-R^\star\le\frac{\mathcal{R}_R(w)^2}{2\mu}.
\end{equation}
Strong convexity of $R$ on $\mathcal W$ is sufficient for~\eqref{eq:constrained-pl}, as shown in Lemma~\ref{lem:strong-convexity-pl}. If $\mathcal W=\mathbb R^d$, then~\eqref{eq:constrained-pl} is the usual Polyak--\L{}ojasiewicz (PL) inequality, which can hold without convexity~\citep{karimi2016}.

Since $\mathcal R_R(w)\le\|\nabla R(w)\|_2$, conditions~\eqref{eq:gradient-growth} and~\eqref{eq:constrained-pl} imply~\eqref{eq:gradient-norm-growth} with $\kappa=v_1/(2\mu)$ and $\sigma_g^2=v_0$. The growth conditions are therefore related, but~\eqref{eq:strong-growth} remains applicable without PL.

\paragraph{Sufficient conditions for variance growth.}
Condition~\eqref{eq:gradient-growth} can be verified through expected smoothness or the self-bounding property of nonnegative smooth losses. Fix a population minimizer $w^\star$ and write $v_\star=\Var_Z(\nabla\ell(w^\star,Z))$ for its gradient variance. The expected smoothness condition~\citep{gower2019} requires that, for some $L_{\mathrm{ES}}>0$ and every $w\in\mathcal W$,
\begin{equation}\label{eq:expected-smoothness}
\mathbb E_Z\|\nabla\ell(w,Z)-\nabla\ell(w^\star,Z)\|_2^2
\le2L_{\mathrm{ES}}[R(w)-R^\star].
\end{equation}
Centering the decomposition of $\nabla\ell(w,Z)$ into its increment from $w^\star$ and its value at $w^\star$ gives~\eqref{eq:gradient-growth} with $v_0=2v_\star$ and $v_1=4L_{\mathrm{ES}}$. In particular, if each loss is convex and $L$-smooth on $\mathbb R^d$ for some $L>0$, then~\eqref{eq:expected-smoothness} holds with $L_{\mathrm{ES}}=L$. This remains true when $w^\star$ is on the boundary of $\mathcal W$: the proof uses the variational inequality at $w^\star$, not the identity $\nabla R(w^\star)=0$. Lemma~\ref{lem:optimization-facts} and the argument following it verify both implications. This route does not require nonnegative losses, and its constants are unchanged by adding a constant to the loss.

A second sufficient condition is the self-bounding property~\citep{srebro2010}. For some $L>0$, suppose that the following inequalities hold for every $w\in\mathcal W$ and every $z\in\mathcal Z$:
\begin{equation}\label{eq:self-bound}
\ell(w,z)\ge0,\qquad \|\nabla\ell(w,z)\|_2^2\le2L\ell(w,z).
\end{equation}
Taking expectation and using $\Var_Z(\nabla\ell(w,Z))\le\mathbb E_Z\|\nabla\ell(w,Z)\|_2^2$ give~\eqref{eq:gradient-growth} with $v_0=2LR^\star$ and $v_1=2L$. Condition~\eqref{eq:self-bound} holds if each loss is nonnegative and $L$-smooth on $\mathbb R^d$, without a convexity assumption. Smoothness and nonnegativity restricted to $\mathcal W$ do not suffice, because the descent step used to prove the self-bound may leave $\mathcal W$. These are sufficient conditions for~\eqref{eq:gradient-growth}, rather than additional assumptions in the result below. The gradient envelope and algorithmic stability remain separate requirements.

\paragraph{The excess risk bound.}
Combining~\eqref{eq:constrained-pl} with the gradient generalization bound~\eqref{eq:gradient-refined} and using~\eqref{eq:gradient-growth} yields the following result.
\begin{corollary}\label{cor:risk}
Assume the differentiability, gradient-envelope, and mean square gradient stability conditions of Subsection~\ref{subsec:gradient-generalization}, together with the variance-growth condition~\eqref{eq:gradient-growth} and the population PL condition~\eqref{eq:constrained-pl}. Then, for every $\delta\in(0,1)$ satisfying $n\ge8v_1\log(e^2/\delta)/\mu$, the following bound holds with probability at least $1-\delta$:
\begin{equation}\label{eq:risk-explicit}
\begin{aligned}
\mathbb E_\xi[R(A(S,\xi))-R^\star]
&\le\frac{4}{\mu}\mathbb E_\xi\mathcal R_{R_{\text{emp}}}(A(S,\xi))^2+\frac{8v_0\log(e^2/\delta)}{\mu n}\\
&\quad+\frac{128L_\ell^2\log^2(e^2/\delta)}{\mu n^2}+\frac{56448\gamma^2\log^2(e^2/\delta)}{\mu}.
\end{aligned}
\end{equation}
\end{corollary}

We prove Corollary~\ref{cor:risk} in Appendix~\ref{app:gradient-risk}. Under expected smoothness, it requires $n\ge32L_{\mathrm{ES}}\log(e^2/\delta)/\mu$ and has a sampling term proportional to $v_\star/n$. Under~\eqref{eq:self-bound}, it requires $n\ge16L\log(e^2/\delta)/\mu$ and has a term proportional to $LR^\star/n$.

For comparison, \citet[Theorem~3]{zhu2025} state, under population PL, $L$-smooth Lipschitz losses, and mean square gradient stability, a high-probability bound of order
\[
\begin{aligned}
\mathbb E_\xi[R(A(S,\xi))-R^\star]
&\lesssim\frac{\mathbb E_\xi\|\nabla R_{\text{emp}}(A(S,\xi))\|_2^2}{\mu}
+\frac{LR^\star\log(1/\delta)}{\mu n}\\
&\quad+\frac{L_\ell^2\log^2(1/\delta)}{\mu n^2}
+\frac{\gamma^2\log^2n\log^2(1/\delta)}{\mu},
\end{aligned}
\]
with the sufficient sample size condition $n\ge32(L/\mu)^2\log(6/\delta)$. Their proof uses a PL distance bound and the self-bounding property, which do not follow from ordinary PL and smoothness on a constrained domain. When these two properties hold and the stationarity residual equals the ambient gradient norm,~\eqref{eq:risk-explicit} has the same orders for empirical stationarity, $R^\star/n$, and $L_\ell^2/n^2$ after self-bounding, but replaces $\gamma^2\log^2n$ by $\gamma^2$ and has a sufficient sample size linear rather than quadratic in $L/\mu$. For constrained problems, we use the normal-cone residual and explicitly assume~\eqref{eq:constrained-pl} and~\eqref{eq:gradient-growth}.

Fix the structural parameters and confidence level. Suppose that $\gamma=O(1/n)$ and that the conditional second moment of the empirical stationarity residual is $O(1/n^2)$ on the event under consideration. Then~\eqref{eq:risk-explicit} gives order $1/n^2$ if $v_0=O(1/n)$, whereas a fixed positive $v_0$ leaves a term of order $1/n$. The squared-loss example in Appendix~\ref{app:fixed-noise} shows that the latter order cannot be improved uniformly, even with zero empirical stationarity residual and gradient stability of order $1/n$. Thus the question of \citet[Section~1]{zhu2025} about stability-based excess-risk bounds faster than $1/n$ must be understood in a restricted regime rather than uniformly over the present assumptions. Under expected smoothness, $v_\star=O(1/n)$ suffices for the required variance bound; under self-bounding, $R^\star=O(1/n)$ suffices. Such conditions describe a sequence of problems unless the corresponding quantity is zero.

At a constrained minimizer $w^\star$, the population gradient need not vanish. If~\eqref{eq:gradient-growth} is imposed there, $v_0=0$ requires only that the stochastic gradient equal its population mean almost surely at $w^\star$, which is compatible with $\nabla R(w^\star)\ne0$. Under expected smoothness, the choice $v_0=2v_\star$ therefore allows a nonzero deterministic gradient at the optimum. For randomized algorithms, Corollary~\ref{cor:risk} controls the excess risk averaged over the seed conditional on the sample; deterministic algorithms are covered by omitting these expectations.

\subsection{Empirical risk minimization and gradient methods}\label{sec:algorithms}
For approximate empirical minimization, we use empirical quadratic growth and Lipschitz continuity to control mixed differences of the empirical optimum value without requiring stability of a minimizer selection or population quadratic growth. For smooth empirical minimization, we compare population risks across empirical minimizers.

\subsubsection{Approximate empirical risk minimization}\label{subsec:approximate}
\citet[Assumptions~2 and~4, Theorem~9]{escande2024} study concentration of distances between empirical and population minimizer sets under local growth conditions for both risks. The result below instead bounds excess risk under empirical quadratic growth, without a population growth assumption. Mixed coordinate differences also quantify weak interactions in \citet[Definition~1]{maurerpontil2018}. Here we control those differences for the empirical optimal value and apply Theorem~\ref{thm:main} to its doubly centered components.

Let $\mathcal W$ be a nonempty subset of a real Hilbert space $\mathcal H$, and suppose the real-valued loss is jointly measurable and each $\ell(\cdot,z)$ is $L_\ell$-Lipschitz on $\mathcal W$. Write $\operatorname{dist}_{\mathcal H}(w,\mathcal M)=\inf_{u\in\mathcal M}\|w-u\|_{\mathcal H}$ for distance to a nonempty set $\mathcal M\subseteq\mathcal H$. Assume that $\mathbb E_Z|\ell(w_0,Z)|<\infty$ for some $w_0\in\mathcal W$, and write $R^\star=\inf_{w\in\mathcal W}R(w)$. For each sample $s\in\mathcal Z^n$, write $\mathcal M_s=\arg\min_{w\in\mathcal W}R_{\text{emp}}(w;s)$ and $R_{\text{emp}}^\star(s)=\inf_{w\in\mathcal W}R_{\text{emp}}(w;s)$. Assume that $s\mapsto R_{\text{emp}}^\star(s)$ is measurable. This holds when $\mathcal W$ is separable, by continuity of the losses, or when a measurable exact minimizer is available. Assume that, for every sample $s\in\mathcal Z^n$, the set $\mathcal M_s$ is nonempty and the empirical risk satisfies the quadratic growth condition~\citep[Definition~5]{charles2018} with a common constant $\lambda>0$. Thus, for every $s\in\mathcal Z^n$ and $w\in\mathcal W$,
\begin{equation}\label{eq:qg-pop-emp}
R_{\text{emp}}(w;s)-R_{\text{emp}}^\star(s)\ge\frac{\lambda}{2}\operatorname{dist}_{\mathcal H}(w,\mathcal M_s)^2.
\end{equation}
Lipschitz continuity and~\eqref{eq:qg-pop-emp} imply $0\le R_{\text{emp}}(w;s)-R_{\text{emp}}^\star(s)\le2L_\ell^2/\lambda$. Thus the empirical optimum is integrable and $R^\star$ is finite. Neither nonnegative losses nor attainment of the population infimum is required. For a measurable output $A(S)\in\mathcal W$, retain $\Delta_{\mathrm{opt}}(S)=R_{\text{emp}}(A(S))-R_{\text{emp}}^\star(S)$.

For strongly convex $L_\ell$-Lipschitz losses, \citet[Proposition~2.1]{klochkov2021} obtain a high-probability excess-risk bound of order
\[
\epsilon_{\mathrm{opt}}+
\left(\frac{L_\ell^2}{\lambda n}
+L_\ell\sqrt{\frac{\epsilon_{\mathrm{opt}}}{\lambda}}\right)
\log n\log(1/\delta).
\]
Under~\eqref{eq:qg-pop-emp}, the distance from each minimizer of one empirical risk to the minimizer set of a neighboring empirical risk is small. This does not imply stability of an arbitrary minimizer selection. If $s,\tilde s$ differ in one observation, then every $u\in\mathcal M_s$ satisfies
\[
\operatorname{dist}_{\mathcal H}(u,\mathcal M_{\tilde s})\le\frac{4L_\ell}{\lambda n},\qquad R_{\text{emp}}(u;\tilde s)-R_{\text{emp}}^\star(\tilde s)\le\frac{8L_\ell^2}{\lambda n^2}.
\]
These bounds control mixed coordinate differences of $R_{\text{emp}}^\star(S)$. Applying Theorem~\ref{thm:main} to its doubly centered components gives a population risk bound common to all empirical minimizers. The quadratic growth condition~\eqref{eq:qg-pop-emp} then extends it to approximate minimizers.

\begin{corollary}\label{cor:approximate}
Under the Lipschitz and empirical quadratic growth assumptions above, for every $\delta\in(0,1)$, there exists a measurable event of probability at least $1-\delta$ on which every measurable output $A(S)\in\mathcal W$ satisfies
\[
R(A(S))-R^\star\le\frac{1310L_\ell^2\log(e^2/\delta)}{\lambda n}+L_\ell\sqrt{\frac{2\Delta_{\mathrm{opt}}(S)}{\lambda}}.
\]
\end{corollary}

Strong convexity with parameter $\lambda$ on a closed convex domain implies~\eqref{eq:qg-pop-emp} with the same parameter. A uniform optimization error bound $\epsilon_{\mathrm{opt}}$ can be substituted for $\Delta_{\mathrm{opt}}(S)$ in Corollary~\ref{cor:approximate}. Compared with \citet[Proposition~2.1]{klochkov2021}, the statistical term has no factor $\log n$, and the square-root optimization term has no logarithmic factor. The displayed constants are not asserted to be smaller than those in that reference. Setting $\Delta_{\mathrm{opt}}(S)=0$ gives a bound simultaneously for all empirical minimizers, with no restriction on their selection. More generally, the corollary uses the realized optimization error and applies on the same event to randomized outputs for every seed. Its proof and the neighboring-set estimates are in Appendix~\ref{app:approximate}.

\paragraph{Projected subgradient descent.} Under the assumptions of Corollary~\ref{cor:approximate}, suppose that $\mathcal W$ is closed and convex and that the losses are $\lambda$-strongly convex. Starting at a fixed $w_1\in\mathcal W$, let a deterministic measurable oracle return $g_t\in\partial R_{\text{emp}}(w_t)$ with $\|g_t\|_{\mathcal H}\le L_\ell$. For each integer $t\ge1$, set $w_{t+1}=\Pi_{\mathcal W}(w_t-2g_t/[\lambda(t+1)])$. After $T\ge1$ iterations, return the weighted average $\bar w_{S,T}=2\sum_{t=1}^Ttw_t/[T(T+1)]$.

\citet[Section~2.3]{klochkov2021} combine the $O(1/T)$ empirical suboptimality of this weighted projected method with stability of approximate minimizers. The resulting stability error is proportional to $1/\sqrt T$, so taking $T$ of order $n^2$ suffices to reach the statistical rate of empirical minimization in their bound. The deterministic optimization guarantee is
\[
R_{\text{emp}}(\bar w_{S,T})-\min_{w\in\mathcal W}R_{\text{emp}}(w)
\le\frac{2L_\ell^2}{\lambda(T+1)}.
\]
Corollary~\ref{cor:approximate} consequently gives an excess risk bound of order $(L_\ell^2/\lambda)\left(\log(e^2/\delta)/n+1/\sqrt T\right)$. The choice $T$ of order $n^2$ yields order $L_\ell^2\log(e^2/\delta)/(\lambda n)$: the concentration improvement removes $\log n$, while the required optimization accuracy is unchanged. The projected subgradient estimate is proved in Appendix~\ref{app:subgradient}.

\subsubsection{Empirical risk minimization with smooth losses}\label{subsec:smooth-erm}
Empirical PL gives a distance bound to the empirical minimizer set in the unconstrained setting, but does not ensure stability of a minimizer selection. Moreover, the squared-loss example in Appendix~\ref{app:fixed-noise} rules out a uniform $1/n^2$ excess-risk rate even under smoothness and strong convexity. We give bounds depending on the gradient variance or the risk at a population minimizer, in the spirit of the problem-dependent rates of \citet{zhang2017}. Our argument compares population risks across the empirical minimizer set and uses concentration only at a fixed population minimizer, without requiring Theorem~\ref{thm:main}. For either parameter set considered below, retain $\mathcal M_s=\arg\min_{w\in\mathcal W}R_{\text{emp}}(w;s)$ for the empirical minimizer set.

\paragraph{Unconstrained empirical minimization.}
For this part, let $\mathcal W=\mathcal H$, where $\mathcal H$ is a real separable Hilbert space. Assume that $\ell:\mathcal H\times\mathcal Z\to[0,\infty)$ is measurable in the data argument and continuously Fr\'echet differentiable in its first argument, with strongly measurable gradients. Each loss $\ell(\cdot,z)$ is $L$-smooth on $\mathcal H$, for a common $L>0$. Fix a population minimizer $w^\star$ with finite risk $R^\star=R(w^\star)$, and suppose that $\|\nabla\ell(w^\star,z)\|_{\mathcal H}\le G$ for every $z\in\mathcal Z$, with $G\ge0$.

For some $0<\mu\le L$, assume that there is a measurable set $\mathcal E\subseteq\mathcal Z^n$ with $\mathrm P^n(\mathcal E)=1$ such that the following empirical PL inequality holds for every $s\in\mathcal E$ and every $w\in\mathcal H$:
\begin{equation}\label{eq:smooth-erm-pl}
R_{\text{emp}}(w;s)-\inf_{\tilde w\in\mathcal H}R_{\text{emp}}(\tilde w;s)
\le\frac{\|\nabla R_{\text{emp}}(w;s)\|_{\mathcal H}^2}{2\mu}.
\end{equation}
By independence and continuity, this condition also implies the individual-loss PL inequality almost surely, as shown in Appendix~\ref{app:erm-unconstrained-proof}.

Define $m(z)=\inf_{w\in\mathcal H}\ell(w,z)$ and $\tilde{\ell}(w,z)=\ell(w,z)-m(z)$. Separability and continuity make $m$ measurable. The normalized optimum risk is $r_\star=\mathbb E_Z\tilde{\ell}(w^\star,Z)$, so $0\le r_\star\le R^\star$. Write $v_\star=\Var_Z(\nabla\ell(w^\star,Z))$ for the gradient variance at this minimizer. This normalization leaves both the empirical minimizers and the population excess risk unchanged.

\begin{proposition}\label{prop:smooth-erm}
Under the assumptions of the unconstrained setting above, $\mathcal M_S$ is nonempty almost surely. If $n\ge4L/\mu$, then for every $\delta\in(0,1)$, there exists a measurable event of probability at least $1-\delta$ on which every $\widehat w\in\mathcal M_S$ satisfies
\begin{equation}\label{eq:smooth-erm-risk}
R(\widehat w)-R^\star
\le\frac{14Lv_\star\log(6/\delta)}{\mu^2n}
+\frac{10LG^2\log^2(6/\delta)}{\mu^2n^2}.
\end{equation}
\end{proposition}

The self-bounding inequality gives $v_\star\le2Lr_\star\le2LR^\star$, so the variance can also be replaced by either upper bound. If $v_\star=O(1/n)$ and $L$, $G$, and $1/\mu$ remain bounded, then the bound is $O(\log^2(6/\delta)/n^2)$. In particular, $r_\star=O(1/n)$ suffices by self-bounding. The proposition also applies to randomized selections of empirical minimizers.

\paragraph{Constrained empirical minimization.}
Now let $\mathcal W$ be a nonempty closed convex subset of $\mathcal H$. The nonnegative losses are restrictions of continuously Fr\'echet differentiable functions on open neighborhoods of $\mathcal W$, and their strongly measurable gradients are $L$-Lipschitz on $\mathcal W$. Assume that differentiation passes through the population expectation and that $R$ has a minimizer $w^\star\in\mathcal W$. We use the normal cone $\mathcal N_{\mathcal W}$ and stationarity residual $\mathcal R_\Psi$ from Section~\ref{subsec:learning-notation}, interpreted in $\mathcal H$: distances are computed with $\|\cdot\|_{\mathcal H}$ and denoted by $\operatorname{dist}_{\mathcal H}$.
For some $0<\mu\le L$, suppose that there exists a measurable set $\mathcal E\subseteq\mathcal Z^n$ with $\mathrm P^n(\mathcal E)=1$ such that, for every sample $s\in\mathcal E$ and every $w\in\mathcal W$,
\begin{equation}\label{eq:smooth-erm-constrained-pl}
R_{\text{emp}}(w;s)-\inf_{\tilde w\in\mathcal W}R_{\text{emp}}(\tilde w;s)
\le\frac{\mathcal R_{R_{\text{emp}}(\cdot;s)}(w)^2}{2\mu}.
\end{equation}
Assume also the Hilbert-space version of the self-bounding condition~\eqref{eq:self-bound}, namely $\|\nabla\ell(w,z)\|_{\mathcal H}^2\le2L\ell(w,z)$ for every $w\in\mathcal W$ and $z\in\mathcal Z$, and suppose that $0\le\ell(w^\star,Z)\le B$ almost surely for some $B\ge0$. As noted after~\eqref{eq:self-bound}, the self-bounding condition follows from nonnegative $L$-smooth extensions to all of $\mathcal H$, but not necessarily from smoothness and nonnegativity restricted to $\mathcal W$. Unlike in the unconstrained case, a minimizer on the boundary need not have a vanishing gradient. The normal-cone residual in~\eqref{eq:smooth-erm-constrained-pl} accounts for this difference, allowing the same risk-comparison approach to be used for constrained empirical minimizers.

\begin{proposition}\label{prop:smooth-erm-constrained}
Under the assumptions of the constrained setting above, $\mathcal M_S$ is nonempty almost surely. If $n\ge4L/\mu$, then for every $\delta\in(0,1)$, there exists a measurable event of probability at least $1-\delta$ on which every $\widehat w\in\mathcal M_S$ satisfies
\begin{equation}\label{eq:smooth-erm-constrained-risk}
R(\widehat w)-R^\star \le 264 \frac{L^2}{\mu^2}\left( \frac{R^\star\log(3/\delta)}{n} +\frac{B\log^2(3/\delta)}{n^2} \right).
\end{equation}
\end{proposition}

If $L$, $B$, and $1/\mu$ remain bounded and $R^\star=O(1/n)$, our bound yields an excess risk of $O(\log^2(3/\delta)/n^2)$. The first example in Appendix~\ref{app:smooth-erm-counterexamples} attains the $1/n^2$ rate in expectation and with probability bounded away from zero, although the gradient stability parameter of its ERM rule does not tend to zero. We prove both ERM results in Appendix~\ref{app:erm}.

\paragraph{Comparison and relation to stability.}
\citet[Theorems~3 and~5]{zhang2017} establish, for smooth losses and a strongly convex population risk, an ERM excess-risk bound of order
\[
\frac{1}{\lambda n^2}+\frac{\kappa R^\star}{n},
\]
where $\kappa=L/\lambda$ is the condition number, with $L$
the smoothness constant and $\lambda$ the population strong convexity constant. The required sample size is
$n=\tilde{\Omega}(\kappa d)$ for convex losses and
$n=\tilde{\Omega}(\kappa^2d)$ without assuming convexity of the losses, while retaining convexity of the empirical risk.
For regularized generalized linear models, \citet[Theorem~7]{zhang2017} also obtain a dimension-independent bound under $n=\tilde{\Omega}(\kappa^2)$, with the stochastic loss at a population minimizer replacing $R^\star$.

Our unconstrained bound~\eqref{eq:smooth-erm-risk} and constrained bound~\eqref{eq:smooth-erm-constrained-risk} instead assume empirical PL rather than population strong convexity, allowing nonconvex empirical objectives and multiple empirical minimizers. The unconstrained result applies to general smooth losses on a separable Hilbert space with $n\ge4L/\mu$, independently of dimension. As in \citet[Assumption~5]{zhang2017}, it requires a gradient bound only at $w^\star$. The constrained result uses normal-cone empirical PL and self-bounding.

\citet[Theorem~4]{zhu2025} state the following high-probability bound for ERM under Lipschitz continuity, smoothness, nonnegativity, and PL for both risks:
\[
R(\widehat w)-R^\star\lesssim\frac{LR^\star\log(1/\delta)}{\mu n}+\frac{L^2L_\ell^2\log^2n\log^2(1/\delta)}{\mu^3n^2},
\]
provided $n\ge32(L/\mu)^2\log(6/\delta)$. Their proof in Appendix~D applies Theorem~3 using the gradient-stability bound of Lemma~1 and zero empirical gradient at an empirical minimizer. We disprove Lemma~1 and Theorem~4 under their stated assumptions in Appendix~\ref{app:smooth-erm-counterexamples}.

First, the proof of Lemma~1 incorrectly infers uniqueness of the empirical minimizer from PL and smoothness. For neighboring samples $s,\tilde s$ and empirical minimizers $\widehat w_s\in\mathcal M_s$ and $\widehat w_{\tilde s}\in\mathcal M_{\tilde s}$, it uses
\[
R_{\text{emp}}(\widehat w_{\tilde s};s)-R_{\text{emp}}^\star(s)\ge\frac{\mu}{2}\|\widehat w_{\tilde s}-\widehat w_s\|_2^2.
\]
Even where PL implies quadratic growth, the lower bound involves distance to $\mathcal M_s$, not to an arbitrarily selected $\widehat w_s$~\citep{karimi2016}. The first example in Appendix~\ref{app:smooth-erm-counterexamples} gives a three-dimensional problem satisfying the assumptions of Lemma~1, with gradient difference $2-1/n$ for neighboring samples instead of the claimed upper bound $100/n$. Thus Lemma~1 is false, even when every empirical minimizer is stationary.

Second, Definition~1 of \citet{zhu2025} permits a constrained domain and uses ordinary PL with the ambient gradient. The second example in Appendix~\ref{app:smooth-erm-counterexamples} satisfies all the stated assumptions of Theorem~4 with constants independent of $n$ and $R^\star=0$, but has excess risk at least $1/2$ for every sample. At fixed confidence, the claimed upper bound tends to zero. Hence Theorem~4 is false. The proof uses zero ambient gradient at an empirical minimizer and, through Theorem~3, self-bounding at a population minimizer. Neither property follows from its hypotheses on a constrained domain. These counterexamples do not disprove an unconstrained version with valid stationarity and self-bounding; such a version would still need a separate justification of gradient stability for the chosen empirical minimizers.

Since Theorem 4 does not hold under its stated assumptions, we do not use it as an official result for comparison in this paper.  
Propositions~\ref{prop:smooth-erm} and~\ref{prop:smooth-erm-constrained} use empirical PL to find a minimizer near $w^\star$ and compare population risks across the empirical minimizer set, without assuming gradient stability or population PL. On a constrained domain, we explicitly use normal-cone empirical PL and self-bounding. The conclusions hold on a common event for all empirical minimizers.

\paragraph{Gradient descent under empirical PL.} Under the unconstrained assumptions of Proposition~\ref{prop:smooth-erm}, fix an initial point $w_0\in\mathcal H$ independent of the data. For each sample $s\in\mathcal X$, set $w_0(s)=w_0$ and $w_{t+1}(s)=w_t(s)-\nabla R_{\text{emp}}(w_t(s);s)/L$ for every integer $t\ge0$. If $n\ge4L/\mu$, then, on a measurable event of probability at least $1-\delta$, we have, simultaneously for every integer $T\ge1$,
\begin{equation}\label{eq:gd-empirical-pl}
\begin{aligned}
R(w_T(S))-R^\star
&\le\frac{28Lv_\star\log(6/\delta)}{\mu^2n}+\frac{20LG^2\log^2(6/\delta)}{\mu^2n^2}\\
&\quad+\frac{2L}{\mu}\left(1-\frac\mu L\right)^T\left(R_{\text{emp}}(w_0;S)-R_{\text{emp}}^\star(S)\right).
\end{aligned}
\end{equation}
We prove the gradient descent bound in Appendix~\ref{app:gd-empirical-pl}.

\phantomsection
\addcontentsline{toc}{section}{References}
\bibliographystyle{plainnat}
\bibliography{references}

\appendix
\section{Proofs for Section~\ref{sec:applications}}\label{app:applications}
The concentration lemmas used below are stated in Section~\ref{subsec:independent-concentration} and proved in Appendix~\ref{app:auxiliary}.

\subsection{Concentration under uniform stability and coordinate sensitivities}\label{app:statistics}
\subsubsection{Proof of Corollary~\ref{cor:statistic}}
\begin{proof}[\textnormal{\bfseries Proof}]
We first reduce the generalization gap to a sum of conditionally centered functions. For each $i\in[n]$, let $Z_i'$ be an independent copy of $Z_i$ and let $S^{(i)}$ be the sample obtained by replacing $Z_i$ with $Z_i'$. Define
\[
g_i(S)=\mathbb E_{Z_i'}\left[\mathbb E_ZF(S^{(i)},Z)-F(S^{(i)},Z_i)\right].
\]
The expectation over $Z_i'$ is taken with $S$ fixed. Conditional on $S_{-i}$, the modified sample $S^{(i)}$ is independent of the original observation $Z_i$. Averaging over $Z_i$ therefore gives $\mathbb E[g_i(S)\mid S_{-i}]=0$. By~\eqref{eq:statistic-assumptions}, replacing $S$ with $S^{(i)}$ changes each of the two terms defining the generalization gap by at most $\gamma_i$. By Cauchy--Schwarz, $(1/n)\sum_{i=1}^n\gamma_i\le\gamma_{\mathrm{rms}}$. Consequently,
\begin{equation}\label{eq:vector-reduction}
\left\|\Delta(S)-\frac1n\sum_{i=1}^n g_i(S)\right\|_{\mathcal H}\le2\gamma_{\mathrm{rms}}.
\end{equation}
Moreover, replacing a coordinate $j\ne i$ changes $g_i$ by at most $2\gamma_j$, and $\|g_i(S)\|_{\mathcal H}\le2M$.

To separate the dependent and independent parts of this sum, define the deterministic mean function $\bar F(z)=\mathbb E_{S'}F(S',z)$, where $S'\sim\mathrm P^n$ is independent of $S$. Integrating the definition of $g_i$ over $S_{-i}$ and $Z_i'$ yields
\[
k_i(Z_i):=\mathbb E[g_i(S)\mid Z_i]
=\mathbb E_Z\bar F(Z)-\bar F(Z_i).
\]
In particular, the variables $k_i(Z_i)$ are independent, centered, and bounded in norm by $2M$. Set $h_i(S)=g_i(S)-k_i(Z_i)$. The function $h_i$ is centered conditional on $Z_i$ by construction. It is also centered conditional on $S_{-i}$, since $k_i(Z_i)$ is centered and independent of $S_{-i}$. Thus $h_i$ satisfies both conditions in~\eqref{eq:double-centering}. Subtracting $k_i(Z_i)$ does not change the sensitivity bounds for coordinates $j\ne i$.

The sensitivity matrix of $(h_i)_{i\in[n]}$ has Frobenius norm at most $2\sqrt{n(n-1)}\,\gamma_{\mathrm{rms}}$. Theorem~\ref{thm:main} gives $\|\sum_{i=1}^n h_i(S)\|_{L_p(\mathcal H)}\le30pn\gamma_{\mathrm{rms}}$, while Lemma~\ref{lem:independent} gives $\|\sum_{i=1}^n k_i(Z_i)\|_{L_p(\mathcal H)}\le4M\sqrt{pn}$. Hence, by~\eqref{eq:vector-reduction}, for every $p\ge2$,
\[
\begin{aligned}
\|\Delta(S)\|_{L_p(\mathcal H)}
&\le2\gamma_{\mathrm{rms}}+30p\gamma_{\mathrm{rms}}+4M\sqrt{p/n}\\
&\le31p\gamma_{\mathrm{rms}}+4M\sqrt{p/n}.
\end{aligned}
\]
For $p=2$, write $\sigma^2=\Var_Z(\bar F(Z))$, so that $\mathbb E\|k_i(Z_i)\|_{\mathcal H}^2=\sigma^2$. The $L_2$ bound following Corollary~\ref{cor:general-moment} and~\eqref{eq:vector-reduction} give
\[
\|\Delta(S)\|_{L_2(\mathcal H)}\le2\gamma_{\mathrm{rms}}+\sqrt{4(n-1)\gamma_{\mathrm{rms}}^2/n+\sigma^2/n}.
\]
Applying Jensen's inequality to the centered functions gives
\[
\sigma^2=\Var_Z(\bar F(Z))\le\mathbb E_S\Var_Z(F(S,Z))\le M^2.
\]
This gives the $L_2$ bound in~\eqref{eq:statistic-moment}.

It remains to obtain a confidence bound in terms of the variance of $F(S,\cdot)$. We control the two components separately. Applying Markov's inequality to the preceding dependent-component moment bound at order $\log(e^2/\delta)$ gives, with probability at least $1-\delta/e^2$,
\[
\left\|\frac1n\sum_{i=1}^n h_i(S)\right\|_{\mathcal H}
\le30e\gamma_{\mathrm{rms}}\log(e^2/\delta).
\]
The independent centered variables $k_i(Z_i)$ have second moment $\sigma^2$ and norm at most $2M$. Lemma~\ref{lem:bernstein} therefore implies, with probability at least $1-2\delta/e^2$,
\[
\left\|\frac1n\sum_{i=1}^n k_i(Z_i)\right\|_{\mathcal H}
\le2\sigma\sqrt{\frac{\log(e^2/\delta)}{n}}
+\frac{4M\log(e^2/\delta)}{n}.
\]

To relate $\sigma^2$ to the variance of $F(S,Z)$, let $P_0$ denote the orthogonal projection in $L_2(\mathrm P;\mathcal H)$ onto the subspace of functions with mean zero, given by $P_0 f=f-\mathbb E_Z f(Z)$. Fubini's theorem gives $(P_0\bar F)(z)=\mathbb E_S[(P_0 F(S,\cdot))(z)]$ for $\mathrm P$-almost every $z$. Minkowski's integral inequality then implies
\[
\sigma=\|P_0\bar F\|_{L_2(\mathrm P;\mathcal H)}\le\mathbb E_S\|P_0 F(S,\cdot)\|_{L_2(\mathrm P;\mathcal H)}.
\]
The reverse triangle inequality and the contraction property of $P_0$ imply that replacing observation $j$ changes $\|P_0F(S,\cdot)\|_{L_2(\mathrm P;\mathcal H)}=\sqrt{\Var_Z(F(S,Z))}$ by at most $\gamma_j$. The lower-tail bounded differences inequality~\citep{mcdiarmid1989} therefore gives, with probability at least $1-\delta/e^2$,
\begin{equation}\label{eq:second-moment-localization} \sigma\le\sqrt{\Var_Z(F(S,Z))}+\gamma_{\mathrm{rms}}\sqrt{\frac{n\log(e^2/\delta)}{2}}. \end{equation}

The three probability estimates hold simultaneously outside an event of probability at most $4\delta/e^2<\delta$. Combining them with~\eqref{eq:vector-reduction} and~\eqref{eq:second-moment-localization}, we obtain
\[
\begin{aligned}
\|\Delta(S)\|_{\mathcal H}
&\le\left(2+(30e+\sqrt2)\log(e^2/\delta)\right)\gamma_{\mathrm{rms}}+2\sqrt{\frac{\log(e^2/\delta)}{n}\Var_Z(F(S,Z))}
+\frac{4M\log(e^2/\delta)}{n}.
\end{aligned}
\]
Since $\log(e^2/\delta)\ge2$ and $30e+\sqrt2+1<84$, this is the claimed confidence bound.
\end{proof}

\subsection{Loss generalization and excess risk}\label{app:loss-applications}

\subsubsection{Proof of Corollary~\ref{cor:learning-sharp-order}}\label{app:learning-sharp-order}
\begin{proof}[\textnormal{\bfseries Proof}]
We first prove the upper moment bound for each learning problem in the supremum. Apply Corollary~\ref{cor:statistic} with $\mathcal H=\mathbb R$ and $F(s,z)=\ell(A(s),z)$. For $p\ge2$, the moment estimate~\eqref{eq:statistic-moment} gives
\[
\|R(A(S))-R_{\text{emp}}(A(S))\|_{L_p}
\le31\gamma p+4M\sqrt{\frac{p}{n}}.
\]
For $1\le p<2$, monotonicity of moment norms and the second-moment estimate in the same corollary give
\[
\begin{aligned}
\|R(A(S))-R_{\text{emp}}(A(S))\|_{L_p}
&\le\|R(A(S))-R_{\text{emp}}(A(S))\|_{L_2}\\
&\le2\gamma+\sqrt{\frac{4(n-1)\gamma^2}{n}+\frac{M^2}{n}}
\le31\gamma p+4M\sqrt{\frac{p}{n}}.
\end{aligned}
\]
Both risks lie in $[0,M]$, so their difference has absolute value at most $M$. Combining this deterministic bound with the moment estimate yields, for every $p\ge1$,
\[
\begin{aligned}
\|R(A(S))-R_{\text{emp}}(A(S))\|_{L_p}
&\le\min\left\{M,31\gamma p+4M\sqrt{\frac pn}\right\}\\
&\le31\min\left\{M,\gamma p+M\sqrt{\frac pn}\right\}.
\end{aligned}
\]
Taking the supremum over the learning problems under consideration proves the upper estimate in~\eqref{eq:learning-sharp-moment}.

For the lower estimate, use the single learning problem from Proposition~\ref{prop:learning-lower}. Its positive tail is sufficient to bound the absolute moment. Indeed, multiplying the $p$th power of the threshold in~\eqref{eq:learning-lower-tail} by its probability gives
\[
\mathbb E_S|R(A(S))-R_{\text{emp}}(A(S))|^p
\ge e^{-p}\left(\frac1{256}\min\left\{M,\gamma p+M\sqrt{\frac pn}\right\}\right)^p.
\]
Taking $p$th roots proves the lower moment estimate with constant $1/(256e)$. Since the same distribution, loss, and algorithm satisfy~\eqref{eq:learning-lower-tail} for every $p$, the upper and lower estimates also match at the tail probability scale $e^{-p}$.

To see the implication for a prescribed failure probability more explicitly, fix $p\ge2$ and apply~\eqref{eq:learning-lower-tail} with $p/2$. The resulting event has probability at least $e^{-p/2}>e^{-p}$. Its threshold is at least half the lower-tail threshold at $p$, because
\[
\min\left\{M,\frac{\gamma p}{2}+M\sqrt{\frac{p}{2n}}\right\}
\ge\frac12\min\left\{M,\gamma p+M\sqrt{\frac pn}\right\}.
\]
Thus, for every $p\ge2$, no uniformly smaller-order threshold can give an $e^{-p}$ upper-tail guarantee over this class.
\end{proof}

\subsubsection{Proof of Corollary~\ref{cor:generalized-bernstein-risk}}\label{app:generalized-bernstein-risk}
\begin{proof}[\textnormal{\bfseries Proof}]
For each sample $s$, let $f_s(z)=\ell(A(s),z)$ and define the deterministic mean loss $\bar f(z)=\mathbb E_S f_S(z)$. By the assumed measurability and boundedness, $\bar f$ is also the Bochner mean of $f_S$ in $L_2(\mathrm P)$. We first control the expected excess risk, which will determine the variance of the comparison function below. Uniform stability gives
$\mathbb E_S[R(A(S))-R_{\text{emp}}(A(S))]\le\gamma$.
Moreover, for a fixed $u\in\mathcal W^\star$,
$\mathbb E_S\inf_{w\in\mathcal W}R_{\text{emp}}(w)
\le\mathbb E_S R_{\text{emp}}(u)=R^\star$.
By the definition of $\Delta_{\mathrm{opt}}(S)$, it follows that
\[
\mathbb E_S[R(A(S))-R^\star]
\le\mathbb E_S\Delta_{\mathrm{opt}}(S)+\gamma.
\]

The generalized Bernstein condition does not require the same population minimizer to control every loss $f_s$. To obtain a fixed comparator, let $\mathcal{C}$ be the closed convex hull in $L_2(\mathrm P)$ of the functions $\ell(u,\cdot)$ with $u\in\mathcal W^\star$. Every element of $\mathcal{C}$ has population mean $R^\star$ and takes values in $[0,M]$ almost surely. Let $\tilde f$ be the metric projection of $\bar f$ onto $\mathcal{C}$, and define $g=\bar f-\tilde f$. Both functions are deterministic, and $|g|\le M$ almost surely.

Let $P_0$ be the mean-zero projection from Appendix~\ref{app:statistics}. Every member of $\mathcal C$ has mean $R^\star$, so $P_0\mathcal C=\mathcal C-R^\star$ is closed and convex. Since all candidate comparators have the same mean, $P_0\tilde f$ is the projection of $P_0\bar f$ onto $P_0\mathcal C$. The squared distance to a nonempty closed convex set is convex in a Hilbert space. Jensen's inequality and~\eqref{eq:generalized-bernstein} therefore give
\begin{equation}\label{eq:generalized-projection}
\begin{aligned}
\Var_Z(g(Z))
&=\operatorname{dist}_{L_2(\mathrm P)}(P_0\bar f,P_0\mathcal C)^2\\
&\le\mathbb E_S\operatorname{dist}_{L_2(\mathrm P)}(P_0f_S,P_0\mathcal C)^2\\
&\le B\left(\mathbb E_S\Delta_{\mathrm{opt}}(S)+\gamma\right).
\end{aligned}
\end{equation}
We next compare $\tilde f$ with the empirical optimum. Every finite convex combination of population-optimal loss functions has empirical average at least $\inf_{w\in\mathcal W}R_{\text{emp}}(w)$ for every sample. Since $\tilde f$ is an $L_2(\mathrm P)$ limit of such combinations, a subsequence converges $\mathrm P$-almost surely. Evaluating this limit at the observations shows that
\[
\frac1n\sum_{i=1}^n \tilde f(Z_i)
\ge\inf_{w\in\mathcal W}R_{\text{emp}}(w)
\]
for $\mathrm P^n$-almost every sample.

Let $\Delta_{\mathrm{cen}}(S)$ denote the generalization gap of the centered loss $f_S-\bar f$:
\[
\Delta_{\mathrm{cen}}(S)
=\mathbb E_Z[f_S(Z)-\bar f(Z)]
-\frac1n\sum_{i=1}^n[f_S(Z_i)-\bar f(Z_i)].
\]
Since $\mathbb E_Z\tilde f(Z)=R^\star$ and $g=\bar f-\tilde f$, the preceding empirical comparison yields
\begin{equation}\label{eq:generalized-reduction}
R(A(S))-R^\star
\le\Delta_{\mathrm{opt}}(S)+\Delta_{\mathrm{cen}}(S)
+\frac1n\sum_{i=1}^n[\mathbb E_Zg(Z)-g(Z_i)].
\end{equation}
We first bound the dependent term. The centered function $F(s,z)=f_s(z)-\bar f(z)$ has the same stability parameter as $f_s$ and satisfies $\mathbb E_SF(S,z)=0$ for every $z$. Its first-order component in the replacement construction of the proof of Corollary~\ref{cor:statistic} therefore vanishes. The remaining terms are doubly centered and have sensitivities bounded by $2\gamma$. Their sensitivity matrix has Frobenius norm at most $2n\gamma$, and the normalized replacement error is at most $2\gamma$. By Theorem~\ref{thm:main}, the $L_p$ norm of the doubly centered sum divided by $n$ is at most $30p\gamma$ for every $p\ge2$.
Apply Markov's inequality at moment order $\log(e^2/\delta)$ to the doubly centered sum before adding the replacement error. With probability at least $1-\delta/e^2$,
\[
\Delta_{\mathrm{cen}}(S)\le2\gamma+30e\gamma\log(e^2/\delta).
\]
For the independent term in~\eqref{eq:generalized-reduction}, the variables $\mathbb E_Zg(Z)-g(Z_i)$ are centered, independent, and bounded in absolute value by $2M$. Lemma~\ref{lem:bernstein} gives, with probability at least $1-2\delta/e^2$,
\[
\frac1n\sum_{i=1}^n[\mathbb E_Zg(Z)-g(Z_i)]\le2\sqrt{\frac{\Var_Z(g(Z))\log(e^2/\delta)}n}+\frac{4M\log(e^2/\delta)}n.
\]
Using~\eqref{eq:generalized-projection} in this estimate proves~\eqref{eq:generalized-bernstein-exact}, since $2\gamma+30e\gamma\log(e^2/\delta)\le84\gamma\log(e^2/\delta)$.

To obtain the realized-error bound on the same event, we also localize the comparison variance. The distance to a fixed set is $1$-Lipschitz, and $P_0$ is a contraction. Hence replacing one observation changes $\operatorname{dist}_{L_2(\mathrm P)}(P_0f_S,P_0\mathcal C)$ by at most $\gamma$. Convexity of distance, the lower-tail bounded differences inequality, and~\eqref{eq:generalized-bernstein} give, with probability at least $1-\delta/e^2$,
\[
\begin{aligned}
\sqrt{\Var_Z(g(Z))}
&\le\mathbb E_S\operatorname{dist}_{L_2(\mathrm P)}(P_0f_S,P_0\mathcal C)\\
&\le\operatorname{dist}_{L_2(\mathrm P)}(P_0f_S,P_0\mathcal C)+\gamma\sqrt{n\log(e^2/\delta)/2}\\
&\le\sqrt{B[R(A(S))-R^\star]}+\gamma\sqrt{n\log(e^2/\delta)/2}.
\end{aligned}
\]
Combining these estimates in~\eqref{eq:generalized-reduction}, and using $30e+\sqrt2+1<84$, yields
\[
R(A(S))-R^\star\le\Delta_{\mathrm{opt}}(S)+84\gamma\log(e^2/\delta)+2\sqrt{\frac{B[R(A(S))-R^\star]\log(e^2/\delta)}n}+\frac{4M\log(e^2/\delta)}n.
\]
Young's inequality bounds the square-root term by $\eta[R(A(S))-R^\star]+B\log(e^2/\delta)/(\eta n)$ for every $0<\eta<1$. Rearranging proves~\eqref{eq:generalized-bernstein-realized}. We obtain both bounds on a common event with failure probability at most $4\delta/e^2<\delta$.

For the randomized extension, take $f_S=\mathbb E_\xi\ell(A(S,\xi),\cdot)$. The same stability argument controls the expected generalization gap of this mean loss. Convexity of squared distance gives $\operatorname{dist}_{L_2(\mathrm P)}(P_0f_S,P_0\mathcal C)^2\le B\mathbb E_\xi[R(A(S,\xi))-R^\star]$, without selecting a measurable optimal comparator for each seed. The empirical comparison with $\tilde f$ and every concentration step above are unchanged. They give the stated bounds for the conditional mean excess risk and optimization error.
\end{proof}

\subsection{Gradient generalization and excess risk}\label{app:gradient-applications}

\subsubsection{Proof of the population gradient bound}\label{app:strong-growth}
\begin{proof}[\textnormal{\bfseries Proof}]
Fix a sample $S$ in the event of probability at least $1-\delta$ on which~\eqref{eq:gradient-refined} holds. All expectations over the random seed in this proof are conditional on this sample. Condition~\eqref{eq:gradient-norm-growth} applies at $A(S,\xi)$ for each seed. Taking expectation over $\xi$ and using subadditivity of the square root gives
\[
\left(\mathbb E_\xi\Var_Z(\nabla\ell(A(S,\xi),Z))\right)^{1/2}
\le\sqrt\kappa\left(\mathbb E_\xi\|\nabla R(A(S,\xi))\|_2^2\right)^{1/2}+\sigma_g.
\]
Minkowski's inequality in $L_2(\mathrm{Q};\mathbb R^d)$ bounds the root mean square population gradient by the corresponding empirical gradient and their root mean square difference. Substituting the preceding inequality into~\eqref{eq:gradient-refined} and moving the population gradient term to the left therefore yields
\[
\begin{aligned}
&\left(1-2\sqrt{\frac{\kappa\log(e^2/\delta)}{n}}\right)
\left(\mathbb E_\xi\|\nabla R(A(S,\xi))\|_2^2\right)^{1/2}\\
&\quad\le\left(\mathbb E_\xi\|\nabla R_{\text{emp}}(A(S,\xi))\|_2^2\right)^{1/2}
+2\sigma_g\sqrt{\frac{\log(e^2/\delta)}{n}}
+\left(84\gamma+\frac{4L_\ell}{n}\right)\log(e^2/\delta).
\end{aligned}
\]
The condition $n\ge16\kappa\log(e^2/\delta)$ implies $2\sqrt{\kappa\log(e^2/\delta)/n}\le1/2$. The coefficient on the left is therefore at least $1/2$. Multiplication by two gives~\eqref{eq:strong-growth} on the same event.

For the mean-norm bound, Jensen's inequality and the output envelope imply $\|\nabla R(A(S,\xi))\|_2\le L_\ell$. Consequently,
\[
\mathbb E_\xi\Var_Z(\nabla\ell(A(S,\xi),Z))\le\kappa L_\ell\mathbb E_\xi\|\nabla R(A(S,\xi))\|_2+\sigma_g^2.
\]
The triangle inequality, Cauchy--Schwarz, and~\eqref{eq:gradient-refined} therefore give
\[
\begin{aligned}
\mathbb E_\xi\|\nabla R(A(S,\xi))\|_2
&\le\mathbb E_\xi\|\nabla R_{\text{emp}}(A(S,\xi))\|_2+84\gamma\log(e^2/\delta)+\frac{4L_\ell\log(e^2/\delta)}n\\
&\quad+2\sqrt{\frac{\kappa L_\ell\mathbb E_\xi\|\nabla R(A(S,\xi))\|_2\log(e^2/\delta)}n}+2\sigma_g\sqrt{\frac{\log(e^2/\delta)}n}.
\end{aligned}
\]
For every $\theta>0$, Young's inequality bounds the first square-root term by $\theta\mathbb E_\xi\|\nabla R(A(S,\xi))\|_2/(1+\theta)+\kappa(1+\theta)L_\ell\log(e^2/\delta)/(\theta n)$. Move its population-gradient term to the left and multiply by $1+\theta$ to obtain~\eqref{eq:strong-growth-mean}.
\end{proof}

\subsubsection{Proof of Corollary~\ref{cor:risk}}\label{app:gradient-risk}
\begin{proof}[\textnormal{\bfseries Proof}]
Fix a training sample $S$ on which~\eqref{eq:gradient-refined} holds. All expectations over $\xi$ below are conditional on this sample. For the choice $\Upsilon(S,z,\xi)=\nabla\ell(A(S,\xi),z)$, the generalization gap $\Delta(S)$ defined in Subsection~\ref{subsec:randomization} satisfies
\[
\Delta(S)(\xi)=\nabla R(A(S,\xi))-\nabla R_{\text{emp}}(A(S,\xi)).
\]

We first compare population and empirical stationarity. The normal cone at $w\in\mathcal W$ is the same for both objectives. Thus the triangle inequality gives
\[
\mathcal R_R(w)\le\mathcal R_{R_{\text{emp}}}(w)
+\|\nabla R(w)-\nabla R_{\text{emp}}(w)\|_2.
\]
Apply this inequality at $w=A(S,\xi)$ and use~\eqref{eq:constrained-pl}. Minkowski's inequality gives
\[
\sqrt{2\mu\mathbb E_\xi[R(A(S,\xi))-R^\star]}\le\left(\mathbb E_\xi\mathcal R_{R_{\text{emp}}}(A(S,\xi))^2\right)^{1/2}+\left(\mathbb E_\xi\|\Delta(S)(\xi)\|_2^2\right)^{1/2}.
\]
By~\eqref{eq:gradient-growth},
\[
\mathbb E_\xi\Var_Z(\nabla\ell(A(S,\xi),Z))\le v_0+v_1\mathbb E_\xi[R(A(S,\xi))-R^\star].
\]
Substituting this bound into~\eqref{eq:gradient-refined} and using the preceding stationarity estimate gives
\[
\begin{aligned}
\sqrt{2\mu\mathbb E_\xi[R(A(S,\xi))-R^\star]}&\le\left(\mathbb E_\xi\mathcal R_{R_{\text{emp}}}(A(S,\xi))^2\right)^{1/2}+84\gamma\log(e^2/\delta)+\frac{4L_\ell\log(e^2/\delta)}n\\
&\quad+2\sqrt{\frac{\log(e^2/\delta)}n\left(v_0+v_1\mathbb E_\xi[R(A(S,\xi))-R^\star]\right)}.
\end{aligned}
\]
For nonnegative $a,b,c,d$, Cauchy--Schwarz gives $(a+b+c+d)^2\le4a^2+8b^2+8c^2+2d^2$, since $1/4+1/8+1/8+1/2=1$. Apply this inequality to the four terms on the right, in their displayed order, to obtain
\[
\begin{aligned}
2\mu\mathbb E_\xi[R(A(S,\xi))-R^\star]&\le4\mathbb E_\xi\mathcal R_{R_{\text{emp}}}(A(S,\xi))^2+8\cdot84^2\gamma^2\log^2(e^2/\delta)+\frac{8\cdot4^2L_\ell^2\log^2(e^2/\delta)}{n^2}\\
&\quad+\frac{8\log(e^2/\delta)}n\left(v_0+v_1\mathbb E_\xi[R(A(S,\xi))-R^\star]\right).
\end{aligned}
\]
Moving the last excess-risk term to the left and evaluating the numerical coefficients give
\[
\begin{aligned}
&\left(2\mu-\frac{8v_1\log(e^2/\delta)}n\right)\mathbb E_\xi[R(A(S,\xi))-R^\star]\\
&\quad\le4\mathbb E_\xi\mathcal R_{R_{\text{emp}}}(A(S,\xi))^2+\frac{8v_0\log(e^2/\delta)}n+\frac{128L_\ell^2\log^2(e^2/\delta)}{n^2}+56448\gamma^2\log^2(e^2/\delta).
\end{aligned}
\]
The sample size condition $n\ge8v_1\log(e^2/\delta)/\mu$ implies $8v_1\log(e^2/\delta)/n\le\mu$, so the coefficient on the left is at least $\mu>0$. Dividing by $\mu$ yields~\eqref{eq:risk-explicit} on the same event.
\end{proof}

\subsection{Approximate empirical risk minimization}\label{app:algorithms}

\subsubsection{Proof of Corollary~\ref{cor:approximate}}\label{app:approximate}
\begin{proof}[\textnormal{\bfseries Proof}]
For each sample $s$, Lipschitz continuity gives $R_{\text{emp}}(w;s)-R_{\text{emp}}^\star(s)\le L_\ell\operatorname{dist}_{\mathcal H}(w,\mathcal M_s)$. Combining this inequality with~\eqref{eq:qg-pop-emp} gives $R_{\text{emp}}(w;s)-R_{\text{emp}}^\star(s)\le2L_\ell^2/\lambda$. Consequently,
\[
R_{\text{emp}}(w_0;s)-\frac{2L_\ell^2}{\lambda}\le R_{\text{emp}}^\star(s)\le R_{\text{emp}}(w_0;s).
\]
The empirical optimum is measurable by assumption. The anchor integrability assumption and the preceding bounds make it integrable, and taking expectations gives $R^\star\ge R(w_0)-2L_\ell^2/\lambda> -\infty$. For each fixed $w$, $\mathbb E_SR_{\text{emp}}^\star(S)\le R(w)$, so $\mathbb E_SR_{\text{emp}}^\star(S)\le R^\star$ even if the population infimum is not attained.

We first compare empirical minimizer sets. Let $s,\tilde s$ differ in one coordinate and fix $u\in\mathcal M_s$. The difference $R_{\text{emp}}(\cdot;\tilde s)-R_{\text{emp}}(\cdot;s)$ is $2L_\ell/n$-Lipschitz. Comparing $u$ with minimizers in $\mathcal M_{\tilde s}$ at distances approaching $\operatorname{dist}_{\mathcal H}(u,\mathcal M_{\tilde s})$ and applying~\eqref{eq:qg-pop-emp} yields
\[
\frac{\lambda}{2}\operatorname{dist}_{\mathcal H}(u,\mathcal M_{\tilde s})^2\le R_{\text{emp}}(u;\tilde s)-R_{\text{emp}}^\star(\tilde s)\le\frac{2L_\ell}{n}\operatorname{dist}_{\mathcal H}(u,\mathcal M_{\tilde s}).
\]
This proves the two neighboring-set estimates stated before the corollary, without assuming that a nearest minimizer exists.

Fix different coordinates $i,j$ and replacement observations. Write $s^{(i)}$, $s^{(j)}$, and $s^{(ij)}$ for the corresponding modified samples. For $u\in\mathcal M_s$,
\[
R_{\text{emp}}^\star(s^{(j)})-R_{\text{emp}}^\star(s)=\frac{\ell(u,z_j')-\ell(u,z_j)}n-\left(R_{\text{emp}}(u;s^{(j)})-R_{\text{emp}}^\star(s^{(j)})\right).
\]
Choose $v\in\mathcal M_{s^{(i)}}$ with $\|u-v\|_{\mathcal H}$ approaching its distance from that set and apply the same identity at $s^{(i)}$. The first terms differ by at most $2L_\ell\|u-v\|_{\mathcal H}/n$, whose limiting bound is $8L_\ell^2/(\lambda n^2)$. Both suboptimality terms belong to $[0,8L_\ell^2/(\lambda n^2)]$, so their difference has at most the same absolute bound. Hence
\begin{equation}\label{eq:erm-value-mixed}
\left|R_{\text{emp}}^\star(s^{(ij)})-R_{\text{emp}}^\star(s^{(i)})-R_{\text{emp}}^\star(s^{(j)})+R_{\text{emp}}^\star(s)\right|\le\frac{16L_\ell^2}{\lambda n^2}.
\end{equation}

Use the coordinate averages from Subsection~\ref{subsec:coordinate-averages}, with each marginal equal to $\mathrm P$, and define
\[
h_i=\mathsf D_iR_{\text{emp}}^\star-\left(\prod_{k=i+1}^n\mathsf E_k\right)\mathsf D_iR_{\text{emp}}^\star,\qquad i\in[n].
\]
The empty product is the identity. These are pointwise coordinate integrals, not arbitrary versions of conditional expectations. The anchor bounds above ensure that they are finite for every fixed sample. Both terms have zero average over coordinate $i$, and their averages over all other coordinates agree. Thus $h_i$ satisfies both centering conditions in~\eqref{eq:double-centering}. The second term is the Doob martingale difference $\mathbb E[R_{\text{emp}}^\star(S)\mid Z_1,\ldots,Z_i]-\mathbb E[R_{\text{emp}}^\star(S)\mid Z_1,\ldots,Z_{i-1}]$; see \citet[Exercise~8.17]{odonnell2014} for the corresponding coordinate-projection representation. Summing gives
\[
\sum_{i=1}^nh_i(S)=\sum_{i=1}^n\mathsf D_iR_{\text{emp}}^\star(S)-R_{\text{emp}}^\star(S)+\mathbb E_SR_{\text{emp}}^\star(S).
\]
By~\eqref{eq:erm-value-mixed}, replacing coordinate $j\ne i$ changes $\mathsf D_iR_{\text{emp}}^\star$ by at most $16L_\ell^2/(\lambda n^2)$. The averaged term has the same bound if $j<i$ and is unchanged if $j>i$. Thus every off-diagonal sensitivity of $h_i$ is at most $32L_\ell^2/(\lambda n^2)$. Theorem~\ref{thm:main} yields, for every $p\ge2$,
\[
\left\|\sum_{i=1}^nh_i(S)\right\|_{L_p}\le\frac{480pL_\ell^2}{\lambda n}.
\]

For every $u\in\mathcal M_s$, integrate the neighboring-sample suboptimality bound over the replacing observation and sum over its coordinate. Since $R_{\text{emp}}(u;s)=R_{\text{emp}}^\star(s)$, using the identity for $\sum_{i=1}^nh_i(s)$ and $\mathbb E_SR_{\text{emp}}^\star(S)\le R^\star$ gives
\[
R(u)\le R_{\text{emp}}^\star(s)-\sum_{i=1}^n\mathsf D_iR_{\text{emp}}^\star(s)+\frac{8L_\ell^2}{\lambda n}\le R^\star-\sum_{i=1}^nh_i(s)+\frac{8L_\ell^2}{\lambda n}.
\]
Applying Markov's inequality at moment order $\log(e^2/\delta)$ and using $\log(e^2/\delta)\ge2$ and $480e+4<1310$, we obtain a measurable event of probability at least $1-\delta$ on which every $u\in\mathcal M_S$ satisfies
\[
R(u)-R^\star\le\frac{(8+480e\log(e^2/\delta))L_\ell^2}{\lambda n}\le\frac{1310L_\ell^2\log(e^2/\delta)}{\lambda n}.
\]
Finally,~\eqref{eq:qg-pop-emp} gives $\operatorname{dist}_{\mathcal H}(A(S),\mathcal M_S)\le\sqrt{2\Delta_{\mathrm{opt}}(S)/\lambda}$. Compare $A(S)$ to empirical minimizers approaching this distance. Lipschitz continuity of $R$ and the common bound for all minimizers yield the stated inequality. No measurable choice of the auxiliary minimizers is needed.
\end{proof}

\subsubsection{Proof of the projected subgradient guarantee}
\label{app:subgradient}
\begin{proof}[\textnormal{\bfseries Proof}]
Fix a sample $S$, and let $\widehat w_S$ be the empirical risk minimizer. By nonexpansiveness of projection, the update rule, strong convexity, and $\|g_t\|_{\mathcal H}\le L_\ell$,
\[
\begin{aligned}
t\bigl(R_{\text{emp}}(w_t)-R_{\text{emp}}(\widehat w_S)\bigr)
&\le \frac{\lambda t(t-1)}{4}\|w_t-\widehat w_S\|_{\mathcal H}^2-\frac{\lambda t(t+1)}{4}\|w_{t+1}-\widehat w_S\|_{\mathcal H}^2+\frac{tL_\ell^2}{\lambda(t+1)}.
\end{aligned}
\]
Summing over $t=1,\ldots,T$ gives $\sum_{t=1}^T t(R_{\text{emp}}(w_t)-R_{\text{emp}}(\widehat w_S))\le TL_\ell^2/\lambda$, since the distance terms telescope to a nonpositive term. By convexity and the definition of $\bar w_{S,T}$, the empirical optimization error is at most $2L_\ell^2/[\lambda(T+1)]$.
\end{proof}

\subsection{Smooth empirical risk minimization}\label{app:erm}
\subsubsection{Auxiliary lemmas}
We first establish two lemmas for the proofs of both propositions. Lemma~\ref{lem:smooth-erm-distance} bounds the distance to the empirical minimizer set in terms of the stationarity residual.
Lemma~\ref{lem:smooth-erm-oscillation} uses the PL condition on samples with one observation replaced to bound differences in population risk between empirical minimizers.
Together, these lemmas allow us to control the population risk of every empirical minimizer using concentration only for the empirical loss and gradient at a fixed population minimizer.
For the gradient, we use the Hilbert-space Bernstein inequality in Lemma~\ref{lem:bernstein}. Throughout this part, distances and gradients are taken in
$\mathcal H$, following the conventions in
Section~\ref{subsec:smooth-erm}.

\begin{lemma}\label{lem:smooth-erm-distance}
Let $\mathcal W\subseteq\mathcal H$ be nonempty, closed, and convex, and let $f$ be continuously Fr\'echet differentiable on an open neighborhood of $\mathcal W$. Suppose $f^\star=\inf_{w\in\mathcal W}f(w)>-\infty$ and that, for some $\mu>0$, the inequality $f(w)-f^\star\le\mathcal R_f(w)^2/(2\mu)$ holds for every $w\in\mathcal W$. Write $\mathcal M=\arg\min_{w\in\mathcal W}f(w)$. Then the set $\mathcal M$ is nonempty, and for every $w\in\mathcal W$,
\begin{equation}\label{eq:smooth-erm-distance}
\operatorname{dist}_{\mathcal H}(w,\mathcal M)
\le\sqrt{\frac{2(f(w)-f^\star)}{\mu}}
\le\frac{\mathcal R_f(w)}{\mu}.
\end{equation}
\end{lemma}

\begin{proof}
Define $h(w)=\sqrt{2(f(w)-f^\star)/\mu}$ on the complete metric space $\mathcal W$. Fix $w_0\in\mathcal W$. If $h(w_0)=0$, then $w_0$ is a minimizer. Otherwise, fix $\varepsilon>0$ and apply Ekeland's variational principle~\citep[Theorem~1.1]{ekeland1974} to $h$, with accuracy $(1+\varepsilon/2)h(w_0)$ and localization radius $(1+\varepsilon)h(w_0)$. Set $c_\varepsilon=(1+\varepsilon/2)/(1+\varepsilon)<1$. The principle gives $v\in\mathcal W$ with $\|v-w_0\|_{\mathcal H}\le(1+\varepsilon)h(w_0)$ and $h(u)\ge h(v)-c_\varepsilon\|u-v\|_{\mathcal H}$ for every $u\in\mathcal W$.

Suppose $h(v)>0$. Then $h$ is differentiable near $v$. Taking directional derivatives along segments in $\mathcal W$ shows that $v$ minimizes the convex function $u\mapsto\langle\nabla h(v),u-v\rangle+c_\varepsilon\|u-v\|_{\mathcal H}$ on $\mathcal W$. Convex first-order optimality therefore gives $\mathcal R_h(v)\le c_\varepsilon$. On the other hand, $\nabla h(v)=\nabla f(v)/\sqrt{2\mu(f(v)-f^\star)}$. Since the normal cone is a cone, the assumed PL inequality gives
\[
\mathcal R_h(v)
=\frac{\mathcal R_f(v)}{\sqrt{2\mu(f(v)-f^\star)}}\ge1,
\]
which is a contradiction. Hence $h(v)=0$, so $\mathcal M$ is nonempty and $\operatorname{dist}_{\mathcal H}(w_0,\mathcal M)\le(1+\varepsilon)h(w_0)$. Letting $\varepsilon\downarrow0$ gives the first inequality in~\eqref{eq:smooth-erm-distance}. The second follows from the assumed PL inequality.
\end{proof}

\begin{lemma}\label{lem:smooth-erm-oscillation}
Let $\mathcal W\subseteq\mathcal H$ be nonempty, closed, and convex. Assume the differentiability and measurability conditions of the constrained smooth ERM setting in Section~\ref{subsec:smooth-erm}, that $R(w)<\infty$ for every $w\in\mathcal W$, and the empirical PL condition~\eqref{eq:smooth-erm-constrained-pl}. Suppose also that $\|\nabla\ell(w,z)\|_{\mathcal H}^2\le2L\ell(w,z)$ for every $w\in\mathcal W$ and $z\in\mathcal Z$. Write $R_{\text{emp}}^\star(s)=\inf_{w\in\mathcal W}R_{\text{emp}}(w;s)$. Then there exists a measurable set $\mathcal E_0\subseteq\mathcal Z^n$ with $\mathrm P^n(\mathcal E_0)=1$ such that, for every $s\in\mathcal E_0$, the set $\mathcal M_s$ is nonempty and the following inequality holds for all $w,\tilde w\in\mathcal M_s$:
\begin{equation}\label{eq:smooth-erm-one-sided}
R(w)-R(\tilde w)\le\frac{2L}{\mu n}\left(R(w)+R_{\text{emp}}^\star(s)\right).
\end{equation}
\end{lemma}

\begin{proof}
Let $\mathcal E$ be the measurable set in~\eqref{eq:smooth-erm-constrained-pl}. Since replacing one coordinate of $S$ by an independent observation preserves its law $\mathrm P^n$, Fubini's theorem gives a measurable set $\mathcal E_0\subseteq\mathcal E$ with $\mathrm P^n(\mathcal E_0)=1$ such that, for each $s\in\mathcal E_0$ and each $i\in[n]$, the replaced sample $s^{(i\leftarrow z)}$ belongs to $\mathcal E$ for $\mathrm P$-almost every $z$. These conditions hold for every parameter value on the same set of samples.

Fix $s=(z_1,\ldots,z_n)\in\mathcal E_0$. Lemma~\ref{lem:smooth-erm-distance} gives $\mathcal M_s\ne\varnothing$. Fix $w,\tilde w\in\mathcal M_s$, an index $i\in[n]$, and $z$ such that $s'=s^{(i\leftarrow z)}\in\mathcal E$. First-order optimality gives $-\nabla R_{\text{emp}}(w;s)\in\mathcal N_{\mathcal W}(w)$, so
\[
\mathcal R_{R_{\text{emp}}(\cdot;s')}(w)
\le\|\nabla R_{\text{emp}}(w;s')-\nabla R_{\text{emp}}(w;s)\|_{\mathcal H}
=\frac{1}{n}\|\nabla\ell(w,z)-\nabla\ell(w,z_i)\|_{\mathcal H}.
\]
Applying~\eqref{eq:smooth-erm-constrained-pl} on $s'$ and then the individual gradient bound gives
\[
0\le R_{\text{emp}}(w;s')-R_{\text{emp}}^\star(s')
\le\frac{2L}{\mu n^2}\left(\ell(w,z)+\ell(w,z_i)\right).
\]
Since $R_{\text{emp}}(\tilde w;s')\ge R_{\text{emp}}^\star(s')$ and $R_{\text{emp}}(w;s)=R_{\text{emp}}(\tilde w;s)$, subtracting the empirical objective values gives
\[
\ell(w,z)-\ell(\tilde w,z)-\ell(w,z_i)+\ell(\tilde w,z_i)\le\frac{2L}{\mu n}\left(\ell(w,z)+\ell(w,z_i)\right).
\]
Integrating over $z\sim\mathrm P$ and averaging over $i$, the signed empirical differences cancel because both empirical risks equal $R_{\text{emp}}^\star(s)$. This gives~\eqref{eq:smooth-erm-one-sided} for every pair $w,\tilde w\in\mathcal M_s$ simultaneously on the event $\mathcal E_0$.
\end{proof}

\subsubsection{Proof of Proposition~\ref{prop:smooth-erm} (unconstrained case)}\label{app:erm-unconstrained-proof}
\begin{proof}[\textnormal{\bfseries Proof}]
The bound on $\nabla\ell(w^\star,z)$ and $L$-smoothness imply that $R$ is finite and $L$-smooth on $\mathcal H$, with $\nabla R(w)=\mathbb E_Z\nabla\ell(w,Z)$. Indeed, Taylor's inequality bounds the loss at each $w$ by an integrable quantity, and $\|\nabla\ell(w,z)\|_{\mathcal H}\le G+L\|w-w^\star\|_{\mathcal H}$ justifies differentiation under the expectation on bounded neighborhoods. Population optimality gives $\nabla R(w^\star)=0$, so $v_\star=\mathbb E_Z\|\nabla\ell(w^\star,Z)\|_{\mathcal H}^2$.

Recall the normalized loss $\tilde{\ell}(w,z)=\ell(w,z)-m(z)$ and its optimum risk $r_\star$. Define $\tilde{R}(w)=\mathbb E_Z\tilde{\ell}(w,Z)$ and $\tilde{R}_{\text{emp}}(w;s)=\frac{1}{n}\sum_{i=1}^n\tilde{\ell}(w,z_i)$. Nonnegativity and smoothness of $\tilde{\ell}$ give the self-bounding inequality by taking a gradient step of length $1/L$. To obtain the individual PL inequality from the almost sure empirical assumption, fix $w,v\in\mathcal H$. Suppose that $\ell(w,Z)-\ell(v,Z)>\|\nabla\ell(w,Z)\|_{\mathcal H}^2/(2\mu)$ with positive probability. Independence gives positive probability that all $n$ observations satisfy this strict inequality. Averaging and Jensen's inequality would then give $R_{\text{emp}}(w;S)-R_{\text{emp}}(v;S)>\|\nabla R_{\text{emp}}(w;S)\|_{\mathcal H}^2/(2\mu)$, contradicting~\eqref{eq:smooth-erm-pl}. Thus the individual inequality holds almost surely for each pair $w,v$. Intersect the full-measure events for pairs in a countable dense subset of $\mathcal H$, and extend to every pair by continuity of the losses and gradients. Taking the infimum over $v$ proves the second inequality below, simultaneously for every $w\in\mathcal H$ and almost every $z$:
\begin{equation}\label{eq:smooth-erm-normalized}
\|\nabla\ell(w,z)\|_{\mathcal H}^2\le2L\tilde{\ell}(w,z),
\qquad
\tilde{\ell}(w,z)\le\frac{\|\nabla\ell(w,z)\|_{\mathcal H}^2}{2\mu}.
\end{equation}
In particular, $0\le\tilde{\ell}(w^\star,Z)\le G^2/(2\mu)$ almost surely and $r_\star\le v_\star/(2\mu)$. Subtracting $m(z)$ preserves empirical PL, and Lemma~\ref{lem:smooth-erm-oscillation} applies to the normalized losses.

Work on the measurable set $\mathcal E_0$ of probability one from Lemma~\ref{lem:smooth-erm-oscillation}. For $s\in\mathcal E_0$, define the gradient deviation $\mathcal G(s)=\|\nabla R_{\text{emp}}(w^\star;s)-\nabla R(w^\star)\|_{\mathcal H}$. Since $\nabla R(w^\star)=0$, this equals the empirical gradient norm at $w^\star$. Lemma~\ref{lem:smooth-erm-distance}, with $\mathcal W=\mathcal H$, shows that $\mathcal M_s$ is nonempty and that, for every $\varepsilon>0$, there is $v_{s,\varepsilon}\in\mathcal M_s$ with $\|v_{s,\varepsilon}-w^\star\|_{\mathcal H}\le \mathcal G(s)/\mu+\varepsilon$. The $L$-smoothness of $R$ then gives
\[
R(v_{s,\varepsilon})-R^\star
\le\frac{L}{2}\left(\frac{\mathcal G(s)}{\mu}+\varepsilon\right)^2.
\]
Fix $u\in\mathcal M_s$ and write $a=2L/(\mu n)\le1/2$. The normalized version of~\eqref{eq:smooth-erm-one-sided}, together with $\min_w\tilde{R}_{\text{emp}}(w;s)\le\tilde{R}_{\text{emp}}(w^\star;s)$, gives
\[
\begin{aligned}
(1-a)\bigl(R(u)-R^\star\bigr)
&\le R(v_{s,\varepsilon})-R^\star+a\left(r_\star+\tilde{R}_{\text{emp}}(w^\star;s)\right).
\end{aligned}
\]
Divide by $1-a$ and let $\varepsilon\downarrow0$ in the upper bound. For every $u\in\mathcal M_s$, this yields
\begin{equation}\label{eq:smooth-erm-deterministic}
R(u)-R^\star
\le\frac{L}{\mu^2}\mathcal G(s)^2
+\frac{4L}{\mu n}\left(r_\star+\tilde{R}_{\text{emp}}(w^\star;s)\right).
\end{equation}
Fix $\delta\in(0,1)$. For the gradient, the centered variables $X_i=\nabla\ell(w^\star,Z_i)$ satisfy $\|X_i\|_{\mathcal H}\le G$ and $\mathbb E\|X_i\|_{\mathcal H}^2=v_\star$. Lemma~\ref{lem:bernstein} gives, with probability at least $1-\delta/3$,
\[
\mathcal G(S)\le2\sqrt{\frac{v_\star \log(6/\delta)}{n}}+\frac{2G\log(6/\delta)}{n}.
\]
For the loss, let $Y_i=\tilde{\ell}(w^\star,Z_i)$ and $b=G^2/(2\mu)$. Then $0\le Y_i\le b$, $\mathbb E Y_i=r_\star$, and $\operatorname{Var}(Y_i)\le br_\star$. The scalar Bernstein inequality~\citep{boucheron2013} and $\sqrt{2xy}\le x+y/2$ for every $x,y\ge0$ give, with probability at least $1-\delta/6$,
\[
\tilde{R}_{\text{emp}}(w^\star;S)
\le r_\star+\sqrt{\frac{2br_\star \log(6/\delta)}{n}}+\frac{b\log(6/\delta)}{3n}
\le2r_\star+\frac{G^2\log(6/\delta)}{2\mu n}.
\]
These estimates also hold directly when $G=0$.

A union bound shows that both estimates hold on a measurable event of probability at least $1-\delta$. Squaring the gradient bound and substituting in~\eqref{eq:smooth-erm-deterministic} gives, simultaneously for every $u\in\mathcal M_S$,
\[
R(u)-R^\star
\le\frac{8Lv_\star\log(6/\delta)}{\mu^2n}
+\frac{12Lr_\star}{\mu n}
+\frac{8LG^2\log^2(6/\delta)}{\mu^2n^2}
+\frac{2LG^2\log(6/\delta)}{\mu^2n^2}.
\]
Since $r_\star\le v_\star/(2\mu)$ and $\log(6/\delta)\ge1$, this implies~\eqref{eq:smooth-erm-risk}. Neither the event nor the bound requires a measurable choice of the auxiliary minimizers $v_{s,\varepsilon}$.
\end{proof}

\subsubsection{Proof of Proposition~\ref{prop:smooth-erm-constrained} (constrained case)}\label{app:erm-constrained-proof}
\begin{proof}[\textnormal{\bfseries Proof}]
The self-bounding assumption at $w^\star$ gives $\|\nabla\ell(w^\star,Z)\|_{\mathcal H}\le\sqrt{2LB}$ almost surely. Smoothness then implies that $R(w)$ is finite for every $w\in\mathcal W$. Work on the measurable set $\mathcal E_0$ of probability one constructed in Lemma~\ref{lem:smooth-erm-oscillation}. For $s\in\mathcal E_0$, define the gradient deviation $\mathcal G(s)=\|\nabla R_{\text{emp}}(w^\star;s)-\nabla R(w^\star)\|_{\mathcal H}$. Population optimality gives $-\nabla R(w^\star)\in\mathcal N_{\mathcal W}(w^\star)$, and hence $\mathcal R_{R_{\text{emp}}(\cdot;s)}(w^\star)\le \mathcal G(s)$. By~\eqref{eq:smooth-erm-distance}, the empirical minimizer set is nonempty and, for every $\varepsilon>0$, it contains $v_{s,\varepsilon}$ such that $\|v_{s,\varepsilon}-w^\star\|_{\mathcal H}\le \mathcal G(s)/\mu+\varepsilon$.

The function $R-R_{\text{emp}}(\cdot;s)$ has a $2L$-Lipschitz gradient on $\mathcal W$. Empirical optimality and the smoothness inequality therefore give
\[
\begin{aligned}
R(v_{s,\varepsilon})-R^\star
&\le (R-R_{\text{emp}}(\cdot;s))(v_{s,\varepsilon})
-(R-R_{\text{emp}}(\cdot;s))(w^\star)\\
&\le \mathcal G(s)\|v_{s,\varepsilon}-w^\star\|_{\mathcal H}
+L\|v_{s,\varepsilon}-w^\star\|_{\mathcal H}^2.
\end{aligned}
\]
Using $\|v_{s,\varepsilon}-w^\star\|_{\mathcal H}\le\mathcal G(s)/\mu+\varepsilon$, the right-hand side is at most $\mathcal G(s)(\mathcal G(s)/\mu+\varepsilon)+L(\mathcal G(s)/\mu+\varepsilon)^2$, which tends to $(1/\mu+L/\mu^2)\mathcal G(s)^2\le2L\mathcal G(s)^2/\mu^2$ as $\varepsilon\downarrow0$.

Fix $u\in\mathcal M_s$ and set $a=2L/(\mu n)\le1/2$. Apply~\eqref{eq:smooth-erm-one-sided} with $w=u$ and $\tilde w=v_{s,\varepsilon}$, and use $R_{\text{emp}}^\star(s)\le R_{\text{emp}}(w^\star;s)$. As in the unconstrained proof,
\[
(1-a)(R(u)-R^\star)\le R(v_{s,\varepsilon})-R^\star+a\left(R^\star+R_{\text{emp}}(w^\star;s)\right).
\]
Dividing by $1-a$ and letting $\varepsilon\downarrow0$ in the upper bound gives
\begin{equation}\label{eq:smooth-erm-constrained-deterministic}
R(u)-R^\star
\le\frac{4L}{\mu^2}\mathcal G(s)^2
+\frac{4L}{\mu n}\left(R^\star+R_{\text{emp}}(w^\star;s)\right).
\end{equation}

For concentration, fix $\delta\in(0,1)$ and set $X_i=\nabla\ell(w^\star,Z_i)-\nabla R(w^\star)$. These variables are centered and satisfy $\|X_i\|_{\mathcal H}\le2\sqrt{2LB}$ and $\mathbb E\|X_i\|_{\mathcal H}^2\le2LR^\star$. Lemma~\ref{lem:bernstein} gives, with probability at least $1-2\delta/3$,
\[
\mathcal G(S)\le4\sqrt{\frac{LR^\star \log(3/\delta)}{n}}+\frac{4\sqrt{2LB}\,\log(3/\delta)}{n}.
\]
Applying the scalar Bernstein inequality to $\ell(w^\star,Z_i)\in[0,B]$, whose variance is at most $BR^\star$, and using $2ab\le a^2+b^2$ gives, with probability at least $1-\delta/3$,
\[
R_{\text{emp}}(w^\star;S)
\le R^\star+\sqrt{\frac{2BR^\star \log(3/\delta)}{n}}+\frac{B\log(3/\delta)}{3n}
\le2R^\star+\frac{2B\log(3/\delta)}{n}.
\]
A union bound gives a measurable event of probability at least $1-\delta$ on which both bounds hold. Squaring the gradient estimate, substituting both bounds into~\eqref{eq:smooth-erm-constrained-deterministic}, and using $\mu\le L$ and $\log(3/\delta)\ge1$ yields
\[
R(u)-R^\star
\le\frac{140L^2R^\star \log(3/\delta)}{\mu^2n}
+\frac{264L^2B\log^2(3/\delta)}{\mu^2n^2}
\le264\frac{L^2}{\mu^2}\left(\frac{R^\star \log(3/\delta)}{n}+\frac{B\log^2(3/\delta)}{n^2}\right).
\]
The event is common to all $u\in\mathcal M_S$, proving~\eqref{eq:smooth-erm-constrained-risk}.
\end{proof}

\subsubsection{Proof of the excess risk bound for gradient descent}\label{app:gd-empirical-pl}
\begin{proof}[\textnormal{\bfseries Proof}]
Work on the common event in Proposition~\ref{prop:smooth-erm}. The population excess risk is nonnegative and $L$-smooth on $\mathcal H$, so its gradient satisfies $\|\nabla R(v)\|_{\mathcal H}^2\le2L[R(v)-R^\star]$. For $w\in\mathcal H$ and $v\in\mathcal M_S$, smoothness gives
\[
R(w)-R^\star\le\left(\sqrt{R(v)-R^\star}+\sqrt{L/2}\|w-v\|_{\mathcal H}\right)^2.
\]
By Lemma~\ref{lem:smooth-erm-distance}, $\operatorname{dist}_{\mathcal H}(w,\mathcal M_S)\le\sqrt{2[R_{\text{emp}}(w;S)-R_{\text{emp}}^\star(S)]/\mu}$. Choose empirical minimizers approaching this distance, apply the common bound~\eqref{eq:smooth-erm-risk}, and use $(a+b)^2\le2a^2+2b^2$. For every $w\in\mathcal H$, this gives
\[
R(w)-R^\star\le\frac{28Lv_\star\log(6/\delta)}{\mu^2n}+\frac{20LG^2\log^2(6/\delta)}{\mu^2n^2}+\frac{2L}{\mu}\left(R_{\text{emp}}(w;S)-R_{\text{emp}}^\star(S)\right).
\]
The empirical smoothness inequality for a step of length $1/L$, followed by~\eqref{eq:smooth-erm-pl}, yields
\[
R_{\text{emp}}(w_{t+1};S)-R_{\text{emp}}^\star(S)\le\left(1-\frac\mu L\right)\left(R_{\text{emp}}(w_t;S)-R_{\text{emp}}^\star(S)\right).
\]
Iterating this deterministic inequality and substituting $w=w_T(S)$ gives the stated excess risk bound.
\end{proof}

\subsubsection{Counterexamples for minimizer selection and constrained PL}\label{app:smooth-erm-counterexamples}
The following example shows that the failure of gradient stability for an unrestricted ERM selection can coexist with a small excess risk. It satisfies the smoothness, Lipschitz, and empirical PL assumptions of \citet[Lemma~1]{zhu2025}, with constants independent of $n$.

\paragraph{Smoothness.} Take $\mathcal W=[-2,2]^2\times[-\pi,\pi]$ and $\mathcal Z=\{0,1\}$, with $Z$ Bernoulli of parameter $\rho\in(0,1)$. Write $w=(x,\theta)$, with $x\in\mathbb R^2$, define $\varphi(\theta)=(\cos\theta,\sin\theta)$, and set $\ell((x,\theta),z)=\|x-z\varphi(\theta)\|_2^2/2$. Let $I_2$ denote the identity matrix on $\mathbb R^2$. The gradient and Hessian are
\[
\nabla\ell((x,\theta),z)=
\begin{pmatrix}x-z\varphi(\theta)\\-z\langle x,\varphi'(\theta)\rangle\end{pmatrix},
\qquad
\nabla^2\ell((x,\theta),z)=
\begin{pmatrix}I_2&-z\varphi'(\theta)\\-z\varphi'(\theta)^\top&z\langle x,\varphi(\theta)\rangle\end{pmatrix}.
\]
On $\mathcal W$, the gradient has Euclidean norm less than $5$, and the Hessian has operator norm at most $2+2\sqrt2<5$. Convexity of $\mathcal W$ therefore makes every loss $5$-Lipschitz and $5$-smooth.

\paragraph{Empirical minimizers and PL.} For $\bar z_s=\frac{1}{n}\sum_{i=1}^nz_i$, completing the square gives
\[
R_{\text{emp}}((x,\theta);s)
=\frac12\|x-\bar z_s\varphi(\theta)\|_2^2
+\frac12\bar z_s(1-\bar z_s).
\]
Thus $\mathcal M_s=\{(\bar z_s\varphi(\theta),\theta):\theta\in[-\pi,\pi]\}$. All these empirical minimizers have zero ambient gradient, so the example does not rely on a failure of stationarity at the boundary. If $\bar z_s>0$, the principal submatrix of the Hessian of $R_{\text{emp}}(\cdot;s)$ at $(0,0)$ in the $x_2$ and $\theta$ coordinates has determinant $-\bar z_s^2<0$, so the empirical risk is nonconvex.

Every empirical risk also satisfies the normal-cone PL condition~\eqref{eq:smooth-erm-constrained-pl} with $\mu=1$. Indeed, for every $(u,v)\in\mathcal N_{\mathcal W}((x,\theta))$, applying the definition of the normal cone to $(\bar z_s\varphi(\theta),\theta)\in\mathcal W$ gives $\langle u,x-\bar z_s\varphi(\theta)\rangle\ge0$. It follows that
\[
\|\nabla R_{\text{emp}}((x,\theta);s)+(u,v)\|_2^2\ge\|x-\bar z_s\varphi(\theta)+u\|_2^2\ge\|x-\bar z_s\varphi(\theta)\|_2^2.
\]
Taking the infimum over the normal cone, we obtain
\[
\mathcal R_{R_{\text{emp}}(\cdot;s)}((x,\theta))^2\ge2\big(R_{\text{emp}}((x,\theta);s)-R_{\text{emp}}^\star(s)\big).
\]
Since the normal cone contains zero, ordinary PL also holds. Replacing $\bar z_s$ by $\rho$ gives the same PL conditions for the population risk.

\paragraph{Gradient instability.} Define a deterministic ERM rule by $\theta_s=0$ when $z_1=0$ and $\theta_s=\pi$ when $z_1=1$, and set $A(s)=(\bar z_s\varphi(\theta_s),\theta_s)$. For $n\ge2$, take $s=(0,\ldots,0)$ and $s'=(1,0,\ldots,0)$. At the test observation $z=1$, the gradient formula gives
\[
\|\nabla\ell(A(s),1)-\nabla\ell(A(s'),1)\|_2=2-\frac1n.
\]
The bound in \citet[Lemma~1]{zhu2025}, after taking a square root and using $L_\ell=L=5$ and $\mu=1$, would be $100/n$. Since $2-1/n>100/n$ for every $n\ge51$, the example disproves Lemma~1 of \citet{zhu2025}.

\paragraph{Excess risk.} The same example satisfies the assumptions of Proposition~\ref{prop:smooth-erm-constrained}. Since $\varphi'(\theta)$ is a unit vector orthogonal to $\varphi(\theta)$ and $z\in\{0,1\}$, we have
\[
\|\nabla\ell((x,\theta),z)\|_2^2=\|x-z\varphi(\theta)\|_2^2+z^2\langle x-z\varphi(\theta),\varphi'(\theta)\rangle^2\le4\ell((x,\theta),z).
\]
The population risk is $R((x,\theta))=\|x-\rho \varphi(\theta)\|_2^2/2+\rho(1-\rho)/2$, which is nonconvex by the same Hessian argument. A population minimizer is $w^\star=(\rho \varphi(0),0)$, with $R^\star=\rho(1-\rho)/2$ and $\ell(w^\star,z)=(\rho-z)^2/2\le1/2$ for every $z\in\mathcal Z$. We may therefore take $L=5$, $\mu=1$, and $B=1/2$, so Proposition~\ref{prop:smooth-erm-constrained} applies for $n\ge20$. Write $\bar Z=\frac{1}{n}\sum_{i=1}^n Z_i$. For every empirical minimizer $\widehat w\in\mathcal M_S$, the population risk formula gives
\[
R(\widehat w)-R^\star=\frac12(\bar Z-\rho)^2,\qquad
\mathbb E_S[R(\widehat w)-R^\star]=\frac{\rho(1-\rho)}{2n}.
\]

Consider the sequence of distributions $(\mathrm P_n)_{n\ge2}$ obtained by taking $\rho=1/n$. Then $R^\star=(n-1)/(2n^2)$ and the expected excess risk equals $(n-1)/(2n^3)$. On the event $\bar Z=0$, which has probability $(1-1/n)^n\to e^{-1}$, the excess risk is $1/(2n^2)$ for every empirical minimizer. Thus the order $1/n^2$ at fixed confidence can occur even when the gradient stability parameter of the chosen ERM rule is bounded away from zero. In contrast, the loss $\ell(A(s),z)=(\bar z_s-z)^2/2$ is independent of $\theta_s$, and changing one observation changes this loss by at most $1/n$.

\paragraph{Ordinary PL on a constrained domain.} A separate example contradicts \citet[Theorem~4]{zhu2025} under the constrained-domain hypotheses permitted by their Definition~1. It also shows why Proposition~\ref{prop:smooth-erm-constrained} needs assumptions beyond ordinary PL, smoothness, and bounded gradients. Fix $n\ge1$, put $d=2n$, take $\mathcal W_n=[0,1]\times[-1,1]^d$, and let $Z$ be uniform on $\{1,\ldots,d\}$. For $w=(u,v_1,\ldots,v_d)$, define
\[
\ell(w,j)=1-u+\frac{(1-v_j)^2}{4}.
\]
The loss belongs to $[0,2]$, has gradient norm at most $\sqrt2$, and is $1/2$-smooth. Every empirical risk has a suboptimality gap at most $2$, whereas the squared ambient gradient norm is at least $1$ because its $u$ derivative equals $-1$. Hence ordinary PL holds with $\mu=1/4$, uniformly in $n$. The same argument applies to the population risk, whose minimum is $R^\star=0$.

Choose $\widehat u_s=1$, set $\widehat v_{s,j}=1$ for observed indices, and set $\widehat v_{s,j}=-1$ otherwise. This measurable ERM rule has zero empirical risk. Its population risk is the fraction of unobserved indices, which is at least $(d-n)/d=1/2$ for every sample. Thus ordinary constrained PL, smoothness, and Lipschitz continuity do not suffice for a dimension-free vanishing excess-risk bound that holds for every ERM selection. For these losses, the constants in \citet[Theorem~4]{zhu2025} may be taken as $L_\ell=\sqrt2$, $L=1/2$, and $\mu=1/4$. Its sample size condition is $n\ge128\log(6/\delta)$, and its claimed upper bound is of order $\log^2n\log^2(1/\delta)/n^2$, since $R^\star=0$. At every fixed $\delta\in(0,1)$, this tends to zero and contradicts the lower bound $1/2$. The constants do not depend on $d=2n$, so this disproves the stated dimension-independent conclusion of Theorem~4. The population minimizer is unique, and the ERM rule has gradient stability at most $1$, so the projection and existence of a finite stability parameter required by their Theorem~3 hold. The failed steps are the $1/n$ gradient-stability conclusion of Lemma~1 and the use of unconstrained stationarity and self-bounding: at $u=1$ and $v_j=1$, the individual loss is zero but its $u$ derivative remains $-1$. This example does not satisfy the self-bounding assumption of Proposition~\ref{prop:smooth-erm-constrained}. It also does not satisfy its empirical normal-cone PL condition with a constant bounded away from zero uniformly in $n$: take a sample containing an index $j$ exactly once, and set $u=1$, $v_j=-1$, and all other $v_k=1$. The empirical gap is $1/n$ and the stationarity residual is $1/n$, so the normal-cone PL parameter can be at most $1/(2n)$. The example does not address an unconstrained formulation of Theorem~4.

\paragraph{Excess risk at fixed noise.}\label{app:fixed-noise} Let $\mathcal H=\mathbb R$, let $Z$ be a Rademacher random variable, and set $\ell(w,z)=(w-z)^2/2$. For $\mathcal W=\mathbb R$, each loss is nonnegative, $1$-smooth and $1$-strongly convex. The population minimizer is $w^\star=0$, and the unique empirical minimizer is $\widehat w_S=\bar Z=(1/n)\sum_{i=1}^n Z_i$. The assumptions of Proposition~\ref{prop:smooth-erm} hold with $L=\mu=G=1$, and $v_\star=1$, $R^\star=1/2$. Since $R(w)=(w^2+1)/2$, the excess risk is $\bar Z^2/2$. Independence gives
\[
\mathbb E_S\bar Z^2=\frac1n,\qquad \mathbb E_S\bar Z^4=\frac{n+3n(n-1)}{n^4}=\frac{3-2/n}{n^2}.
\]
Applying the Paley--Zygmund inequality~\citep{boucheron2013} to $\bar Z^2$ gives
\[
\mathbb P_S\left\{R(\widehat w_S)-R^\star\ge\frac1{4n}\right\}\ge\frac{(\mathbb E_S\bar Z^2)^2}{4\mathbb E_S\bar Z^4}=\frac{1}{4(3-2/n)}\ge\frac1{12}.
\]
Restricting the domain to $\mathcal W=[-1,1]$ leaves both minimizers unchanged. Strong convexity implies the empirical and population normal-cone PL inequalities with $\mu=1$ by Lemma~\ref{lem:strong-convexity-pl}. The losses satisfy $\lvert\nabla\ell(w,z)\rvert^2=2\ell(w,z)$ and $\ell(w^\star,Z)=1/2$, so Proposition~\ref{prop:smooth-erm-constrained} applies with $L=1$ and $B=1/2$. For the ERM rule, the empirical stationarity residual is zero, the gradient envelope is $L_\ell=2$, and the uniform gradient stability parameter is $\gamma=2/n$, since $\nabla\ell(w,z)=w-z$ and a single replacement changes $\bar Z$ by at most $2/n$. The gradient variance is $1$ at every output, so~\eqref{eq:gradient-growth} holds with $v_0=1$ and $v_1=0$. Corollary~\ref{cor:risk} therefore also applies. This example rules out a uniform $O(1/n^2)$ bound in all three settings at fixed confidence $\delta<1/12$ without a further restriction on the class of problems.

\section{Auxiliary concentration and optimization proofs}\label{app:auxiliary}

\begin{proof}[\textnormal{\bfseries Proof of Lemma~\ref{lem:independent}}]
If $M=0$, each $X_i$ vanishes almost surely, and the conclusion
follows. Assume $M>0$ and let
$Y=\left\|\sum_{i=1}^nX_i\right\|_{\mathcal H}$.
Since the $X_i$ are independent and centered, the cross terms in
the expansion of $\mathbb E Y^2$ vanish. The Cauchy--Schwarz
inequality therefore gives
\[
\mathbb E Y \le (\mathbb E Y^2)^{1/2} =\left(\sum_{i=1}^n\mathbb E\|X_i\|_{\mathcal H}^2\right)^{1/2} \le M\sqrt n.
\]

To estimate the deviation of $Y$ from its expectation, observe
that replacing one $X_i$ by another value in its essential range
changes $Y$ by at most $2M$, by the reverse triangle inequality.
McDiarmid's bounded differences inequality~\citep{mcdiarmid1989}
then yields, for every $u>0$,
\[
\mathbb P\{Y-\mathbb E Y>u\} \le \exp\left(-\frac{u^2}{2nM^2}\right).
\]

For $p\ge2$, consider the positive deviation
$V=(Y-\mathbb E Y)_+$. Using the layer-cake representation
$V^2=\int_0^\infty\mathbf1_{\{V^2>s\}}\,\mathrm{d}s$,
Minkowski's integral inequality with exponent $p/2\ge1$,
and the preceding tail bound, we obtain
\[
\left(\mathbb E V^p\right)^{2/p} \le\int_0^\infty \mathbb P\{V^2>s\}^{2/p}\,\mathrm{d}s \le\int_0^\infty \exp\left(-\frac{s}{npM^2}\right)\,\mathrm{d}s =npM^2.
\]

Since $Y\le\mathbb E Y+V$, Minkowski's inequality and the
bounds established above imply
\[
\left\|\sum_{i=1}^nX_i\right\|_{L_p(\mathcal H)} =(\mathbb E Y^p)^{1/p} \le\mathbb E Y+(\mathbb E V^p)^{1/p} \le M\sqrt n+M\sqrt{np} \le2M\sqrt{np}.
\]
The asserted moment bound follows.
\end{proof}

\subsection{Bernstein inequality in a Hilbert space}

\begin{proof}[\textnormal{\bfseries Proof of Lemma~\ref{lem:bernstein}}]
If $M=0$ or $\sigma=0$, then every $X_i$ vanishes almost surely, and the conclusion follows. Assume $M>0$ and $\sigma>0$. Fix $\lambda>0$ and define $\phi:\mathcal H\to\mathbb R$ by $\phi(x)=\cosh(\lambda\|x\|_{\mathcal H})$.

We first bound the Hessian of $\phi$. For $x\ne0$, the directional derivative of $\|x\|_{\mathcal H}$ along $y$ is $\langle x,y\rangle/\|x\|_{\mathcal H}$. Applying the chain rule and differentiating once more gives
\[
\nabla^2\phi(x)[y,y]
=\lambda^2\cosh(\lambda\|x\|_{\mathcal H})\frac{\langle x,y\rangle^2}{\|x\|_{\mathcal H}^2}
+\frac{\lambda\sinh(\lambda\|x\|_{\mathcal H})}{\|x\|_{\mathcal H}}
\left(\|y\|_{\mathcal H}^2-\frac{\langle x,y\rangle^2}{\|x\|_{\mathcal H}^2}\right).
\]
The expression in parentheses is nonnegative by Cauchy--Schwarz. Since $\sinh(z)\le z\cosh(z)$ for every $z\ge0$, the coefficient of this expression is at most $\lambda^2\cosh(\lambda\|x\|_{\mathcal H})$. Hence $\nabla^2\phi(x)[y,y]\le\lambda^2\|y\|_{\mathcal H}^2\phi(x)$. At $x=0$, the Hessian is $\lambda^2\mathrm{Id}$, so the bound holds for all $x,y\in\mathcal H$.

Now fix $x,y\in\mathcal H$ with $\|y\|_{\mathcal H}\le M$. For every $s\in[0,1]$, the triangle inequality and $\cosh(a+b)\le e^b\cosh(a)$ for every $a,b\ge0$ imply $\phi(x+sy)\le e^{\lambda sM}\phi(x)$. Apply the second-order Taylor formula with integral remainder to $s\mapsto\phi(x+sy)$ on $[0,1]$. Using the Hessian bound and $\int_0^1(1-s)e^{\lambda sM}\,\mathrm{d}s=(e^{\lambda M}-1-\lambda M)/(\lambda M)^2$ for $\lambda M>0$, we obtain
\[
\begin{aligned}
\phi(x+y)
&=\phi(x)+\langle\nabla\phi(x),y\rangle
+\int_0^1(1-s)\nabla^2\phi(x+sy)[y,y] \,\mathrm{d}s\\
&\le\phi(x)+\langle\nabla\phi(x),y\rangle
+\lambda^2\|y\|_{\mathcal H}^2\phi(x)
\int_0^1(1-s)e^{\lambda sM}\,\mathrm{d}s\\
&=\phi(x)+\langle\nabla\phi(x),y\rangle
+\frac{\|y\|_{\mathcal H}^2}{M^2}
(e^{\lambda M}-1-\lambda M)\phi(x).
\end{aligned}
\]
Let $S_0=0$ and $S_k=\sum_{i=1}^kX_i$ for each integer $k\in[n]$. Since $X_k$ is centered and independent of $S_{k-1}$, taking conditional expectations gives
\[
\mathbb E[\phi(S_k)\mid S_{k-1}]
\le\left(1+\frac{\mathbb E\|X_k\|_{\mathcal H}^2}{M^2}(e^{\lambda M}-1-\lambda M)\right)\phi(S_{k-1}).
\]
Iterating this inequality and using $1+z\le e^z$ for every $z\ge0$, together with $\phi(S_0)=1$, we obtain
\[
\mathbb E\phi(S_n)\le\exp\left(\frac{\sigma^2}{M^2}(e^{\lambda M}-1-\lambda M)\right).
\]

For every $z\in[0,3)$, the power series for the exponential function and the inequality $k!\ge2\cdot3^{k-2}$ for every integer $k\ge2$ give $e^z-1-z\le z^2/[2(1-z/3)]$. Combining this estimate with Markov's inequality and $\cosh(\lambda u)\ge e^{\lambda u}/2$, we find that, for every $u>0$ and every $\lambda\in(0,3/M)$,
\[
\mathbb P\left\{\|S_n\|_{\mathcal H}>u\right\}
\le2\exp\left(-\lambda u+\frac{\sigma^2\lambda^2}{2(1-\lambda M/3)}\right).
\]
Choose $\lambda=u/(\sigma^2+Mu/3)$, which belongs to $(0,3/M)$. Substitution gives
\[
\mathbb P\left\{\|S_n\|_{\mathcal H}>u\right\}
\le2\exp\left(-\frac{u^2}{2(\sigma^2+Mu/3)}\right).
\]
Finally, for $u=2\sigma\sqrt{t}+2Mt$, we have $u^2\ge2t(\sigma^2+Mu/3)$. The preceding inequality therefore yields \eqref{eq:bernstein}.
\end{proof}

\subsection{Gradient bounds and constrained domination}
We verify sufficient conditions for variance growth and constrained gradient domination used in Section~\ref{subsec:gradient-risk}.

Throughout this subsection, let $\mathcal W\subseteq\mathbb R^d$ be nonempty, closed, and convex, with the differentiability and differentiation-under-expectation assumptions of Section~\ref{sec:gradient}. Fix a population minimizer $w^\star$, set $R^\star=R(w^\star)$, and suppose $v_\star=\Var_Z(\nabla\ell(w^\star,Z))<\infty$.

\begin{lemma}\label{lem:optimization-facts}
We establish the variance-growth condition~\eqref{eq:gradient-growth} in either of the following settings:
\begin{enumerate}
\item Each loss is nonnegative and $L$-smooth on $\mathbb R^d$, with $L>0$. We can take $v_0=2LR^\star$ and $v_1=2L$.
\item The expected smoothness condition~\eqref{eq:expected-smoothness} holds with $L_{\mathrm{ES}}>0$. We can take $v_0=2v_\star$ and $v_1=4L_{\mathrm{ES}}$.
\end{enumerate}
\end{lemma}
\begin{proof}
Under nonnegative smoothness, fix $z\in\mathcal Z$ and $w\in\mathbb R^d$. Smoothness at the point $w-(1/L)\nabla\ell(w,z)$ gives
\[
0\le\ell\left(w-\frac1L\nabla\ell(w,z),z\right)
\le\ell(w,z)-\frac1{2L}\|\nabla\ell(w,z)\|_2^2.
\]
Rearranging proves~\eqref{eq:self-bound}. Taking expectation over $Z$ gives $\Var_Z(\nabla\ell(w,Z))\le\mathbb E_Z\|\nabla\ell(w,Z)\|_2^2\le2LR(w)$, which yields the stated choices of $v_0$ and $v_1$. The step may leave $\mathcal W$, which explains the assumption on the loss over all of $\mathbb R^d$.

Under expected smoothness, center the decomposition of $\nabla\ell(w,Z)$ into its increment from $w^\star$ and its value at $w^\star$. The inequality $\|a+b\|_2^2\le2\|a\|_2^2+2\|b\|_2^2$, the bound of a variance by its raw second moment, and~\eqref{eq:expected-smoothness} give
\begin{align*}
\Var_Z(\nabla\ell(w,Z))&\le2\Var_Z(\nabla\ell(w,Z)-\nabla\ell(w^\star,Z))+2v_\star\\
&\le2\mathbb E_Z\|\nabla\ell(w,Z)-\nabla\ell(w^\star,Z)\|_2^2+2v_\star\\
&\le4L_{\mathrm{ES}}[R(w)-R^\star]+2v_\star.
\end{align*}
These are the stated choices of $v_0$ and $v_1$.
\end{proof}

For the second route, it is enough that every loss be convex and $L$-smooth on $\mathbb R^d$: in that case~\eqref{eq:expected-smoothness} holds with $L_{\mathrm{ES}}=L$~\citep{srebro2010,gower2019}. The argument uses the self-bounding inequality after subtracting a supporting affine function, and does not require the population gradient to vanish at a constrained minimizer.

To verify this, fix $z\in\mathcal Z$ and define the function $\psi_z:\mathbb R^d\to\mathbb R$ by
\[
\psi_z(w)=\ell(w,z)-\ell(w^\star,z)-\langle\nabla\ell(w^\star,z),w-w^\star\rangle.
\]
Convexity makes $\psi_z$ nonnegative on $\mathbb R^d$, and subtracting an affine function preserves $L$-smoothness. The self-bounding argument above, applied to $\psi_z$, therefore gives
\[
\|\nabla\ell(w,z)-\nabla\ell(w^\star,z)\|_2^2
\le2L\left[\ell(w,z)-\ell(w^\star,z)-\langle\nabla\ell(w^\star,z),w-w^\star\rangle\right].
\]
Taking expectation over $Z$ replaces the loss difference by $R(w)-R^\star$ and the linear term by $\langle\nabla R(w^\star),w-w^\star\rangle$. Since $w^\star$ minimizes the differentiable function $R$ over the convex set $\mathcal W$, the directional derivative along the segment from $w^\star$ to each $w\in\mathcal W$ is nonnegative. Thus $\langle\nabla R(w^\star),w-w^\star\rangle\ge0$, and dropping the subtracted linear term proves~\eqref{eq:expected-smoothness} with $L_{\mathrm{ES}}=L$.

\begin{lemma}\label{lem:strong-convexity-pl}
If $R$ is $\mu$-strongly convex on $\mathcal W$ for some $\mu>0$, then the constrained PL condition~\eqref{eq:constrained-pl} holds with the same $\mu$.
\end{lemma}
\begin{proof}
Fix $w\in\mathcal W$ and $u\in\mathcal{N}_{\mathcal W}(w)$. The definition of the normal cone gives $\langle u,w-w^\star\rangle\ge0$. Using strong convexity and this inequality, then completing the square, gives
\[
\begin{aligned}
R(w)-R^\star
&\le\langle\nabla R(w),w-w^\star\rangle-\frac\mu2\|w-w^\star\|_2^2\\
&\le\langle\nabla R(w)+u,w-w^\star\rangle-\frac\mu2\|w-w^\star\|_2^2\\
&\le\frac{\|\nabla R(w)+u\|_2^2}{2\mu}.
\end{aligned}
\]
Taking the infimum over $u\in\mathcal{N}_{\mathcal W}(w)$ proves~\eqref{eq:constrained-pl}. Unlike the smoothness arguments above, this argument uses only points in $\mathcal W$.
\end{proof}

\section{Summary of statistical bounds and assumptions}\label{app:comparison}
Throughout this appendix, $\lesssim$ hides universal numerical constants, and high-probability bounds hold with probability at least $1-\delta$ over $S$ unless stated otherwise. We take $n\ge2$ and $0<\delta\le1/e$ when comparing logarithmic factors, and write $w_\xi=A(S,\xi)$ for the output of a randomized algorithm.

\subsection{Generalization bounds under stability}\label{app:comparison-stability}
In Table~\ref{tab:applications-comparison}, we compare moment and deviation bounds for generalization gaps and excess-risk bounds under stability assumptions. 
\begingroup
\footnotesize
\setlength{\tabcolsep}{4pt}
\renewcommand{\arraystretch}{1.15}
\setlength{\LTpre}{6pt}
\setlength{\LTpost}{6pt}
\begin{longtable}{@{}>{\raggedright\arraybackslash}p{0.47\textwidth}@{\hspace{0.025\textwidth}}>{\raggedright\arraybackslash}p{0.47\textwidth}@{}}
\caption{Comparison of generalization bounds (moments and deviations) and excess-risk bounds under stability assumptions.}\label{tab:applications-comparison}\\
\hline
\textbf{Prior work} & \textbf{This paper}\\
\hline
\endfirsthead
\multicolumn{2}{@{}l}{\emph{Table~\thetable, continued}}\\
\hline
\textbf{Prior work} & \textbf{This paper}\\
\hline
\endhead
\hline
\multicolumn{2}{r@{}}{\emph{Continued on next page}}\\
\endfoot
\hline
\endlastfoot
\multicolumn{2}{@{}l}{\textbf{Hilbert-valued statistics}}\\*
\citet[Theorem~1]{fanlei2024}: bounded Hilbert-valued $F$, coordinatewise uniform stability $\gamma_j$;
\[
\begin{aligned}
\|\Delta(S)\|_{\mathcal H}
&\lesssim\gamma_{\mathrm{rms}}\log n\,\log(1/\delta)+M\sqrt{\frac{\log(1/\delta}{n}}.
\end{aligned}
\]
&
Corollary~\ref{cor:statistic}: bounded measurable Hilbert-valued $F$, coordinatewise uniform stability~\eqref{eq:statistic-assumptions};
\[
\begin{aligned}
\|\Delta(S)\|_{\mathcal H}
&\lesssim\gamma_{\mathrm{rms}}\log(e^2/\delta)\\
&\quad+\sqrt{\frac{\log(e^2/\delta)\Var_Z(F(S,Z))}{n}}
+\frac{M\log(e^2/\delta)}{n}.
\end{aligned}
\]
\\[2pt]
\hline

\multicolumn{2}{@{}l}{\textbf{Randomized algorithms}}\\*
\citet[Theorem~1]{yuanli2023}: $[0,M]$-valued scalar loss, $L_2$-uniform stability;
\[
\begin{aligned}
\mathbb E_\xi|\Delta(S,\xi)|
&\lesssim\gamma\log n\,\log(1/\delta)+M\sqrt{\frac{\log(1/\delta)}{n}}.
\end{aligned}
\]
&
Equation~\eqref{eq:random-statistic}: strongly measurable $F(s,z)=\Upsilon(s,z,\cdot)$ in a separable subspace of $L_2(\mathrm Q;\mathcal K)$, $\|F(s,z)\|_{L_2(\mathrm Q;\mathcal K)}\le M$, mean square stability~\eqref{eq:mean-square-stability};
\[
\begin{aligned}
(\mathbb E_\xi\|\Delta(S,\xi)\|_{\mathcal K}^2)^{1/2}
&\lesssim\gamma \log(e^2/\delta)\\
&\quad+\sqrt{\frac{\log(e^2/\delta)}{n}
\mathbb E_\xi\Var_Z(\Upsilon(S,Z,\xi))}\\
&\quad+\frac{M\log(e^2/\delta)}{n}.
\end{aligned}
\]
\\[2pt]
\hline

\multicolumn{2}{@{}l}{\textbf{Generalization of bounded losses}}\\*
\citet[Section~3.3, Eq.~(12)]{klochkov2021}: deterministic $\gamma$-stable algorithm, $0\le\ell\le M$;
\[
\begin{aligned}
\Delta(S)
&\lesssim\gamma\log n\,\log(1/\delta)\\
&\quad+\sqrt{\frac{MR(A(S))\log(1/\delta)}{n}}\\
&\quad+\frac{M\log(1/\delta)}{n}.
\end{aligned}
\]
&
Corollary~\ref{cor:uniform}: randomized algorithm, $0\le\ell\le M$, common-seed mean square loss stability;
\[
\begin{aligned}
(\mathbb E_\xi|\Delta(S,\xi)|^2)^{1/2}
&\lesssim\gamma \log(e^2/\delta)\\
&\quad+\sqrt{\frac{\log(e^2/\delta)}{n}\mathbb E_\xi\Var_Z(\ell(w_\xi,Z))}\\
&\quad+\frac{M\log(e^2/\delta)}{n}.
\end{aligned}
\]
\\[2pt]
\hline

\multicolumn{2}{@{}l}{\textbf{Worst-case bounded-loss moments}}\\*
\citet[Corollary~8]{bkz2020}: deterministic $\gamma$-stable algorithm, $0\le\ell\le M$, $p\ge2$;
\[
\|\Delta\|_{L_p}
\lesssim p\gamma\log n
+M\sqrt{\frac pn}.
\]
&
Corollary~\ref{cor:learning-sharp-order}: deterministic $\gamma$-stable algorithm, $0\le\ell\le M$, $p\ge1$;
\[
\begin{aligned}
&\sup_{\mathrm P,\ell,A}\|\Delta\|_{L_p}\asymp\min\left\{M,\gamma p +M\sqrt{\frac pn}\right\}.
\end{aligned}
\]
\\[2pt]
\hline

\multicolumn{2}{@{}l}{\textbf{Generalized Bernstein excess risk}}\\*
\citet[Theorem~1.1]{klochkov2021}: bounded loss, deterministic uniform stability, generalized Bernstein condition with constant $B$, fixed $\eta>0$;
\[
\begin{aligned}
R(A(S))-R^\star
&\lesssim\Delta_{\mathrm{opt}}(S)+\eta\mathbb E_S\Delta_{\mathrm{opt}}(S)\\
&\quad+(1+1/\eta)\left(\gamma\log n+\frac{M+B}{n}\right)\log(1/\delta).
\end{aligned}
\]
&
Corollary~\ref{cor:generalized-bernstein-risk}: bounded loss, deterministic uniform stability, centered Bernstein condition~\eqref{eq:generalized-bernstein}, $\mathcal W^\star\ne\varnothing$, strong $L_2(\mathrm P)$ measurability of the learned loss, all $\eta>0$;
\[
\begin{aligned}
R(A(S))-R^\star
&\lesssim\Delta_{\mathrm{opt}}(S)+\eta\mathbb E_S\Delta_{\mathrm{opt}}(S)\\
&\quad+(1+1/\eta)
\left(\gamma+\frac{M+B}{n}\right)\log(e^2/\delta).
\end{aligned}
\]
\\[2pt]
\hline
\end{longtable}
\endgroup

\subsection{Gradient generalization and excess risk}\label{app:comparison-gradient}
In Table~\ref{tab:gradient-comparison}, we compare bounds for gradient generalization, population gradient norms, and excess risk under gradient domination. We write $D_g(S,\xi)=\nabla R(w_\xi)-\nabla R_{\mathrm{emp}}(w_\xi)$ and $V_g(S)=\mathbb E_\xi\Var_Z(\nabla\ell(w_\xi,Z))$. For deterministic algorithms, we write $D_g(S)$.

\begingroup
\footnotesize
\setlength{\tabcolsep}{4pt}
\renewcommand{\arraystretch}{1.15}
\setlength{\LTpre}{6pt}
\setlength{\LTpost}{6pt}
\begin{longtable}{@{}>{\raggedright\arraybackslash}p{0.47\textwidth}@{\hspace{0.025\textwidth}}>{\raggedright\arraybackslash}p{0.47\textwidth}@{}}
\caption{Comparison of bounds for gradient generalization, population gradient norms, and excess risk under gradient domination.}\label{tab:gradient-comparison}\\
\hline
\textbf{Prior work} & \textbf{This paper}\\
\hline
\endfirsthead
\multicolumn{2}{@{}l}{\emph{Table~\thetable, continued}}\\
\hline
\textbf{Prior work} & \textbf{This paper}\\
\hline
\endhead
\hline
\multicolumn{2}{r@{}}{\emph{Continued on next page}}\\
\endfoot
\hline
\endlastfoot
\multicolumn{2}{@{}l}{\textbf{Gradient generalization}}\\*
\citet[Theorem~3]{fanlei2024}: deterministic algorithm, bounded gradients, coordinatewise gradient stability;
\[
\begin{aligned}
\|D_g(S)\|_2
&\lesssim\gamma_{\mathrm{rms}}\log n\,\log(1/\delta)+L_\ell\sqrt{\log(1/\delta)/n}.
\end{aligned}
\]
\citet[Theorem~2]{zhu2025}: mean square gradient stability, bounded gradients, randomized algorithm;
\[
\begin{aligned}
\mathbb E_\xi\|D_g(S,\xi)\|_2
&\lesssim\gamma\log n\,\log(e/\delta)\\
&\quad+\sqrt{\frac{\log(e/\delta)}{n}\mathbb E_{\xi,Z}\|\nabla\ell(w_\xi,Z)\|_2^2}\\
&\quad+\frac{L_\ell\log(e/\delta)}{n}.
\end{aligned}
\]
&
Equation~\eqref{eq:gradient-refined}: gradients bounded at outputs, coordinatewise mean square gradient stability;
\[
\begin{aligned}
(\mathbb E_\xi\|D_g(S,\xi)\|_2^2)^{1/2}
&\lesssim\gamma_{\mathrm{rms}}\log(e^2/\delta)\\
&\quad+\sqrt{\log(e^2/\delta)V_g(S)/n}\\
&\quad+\frac{L_\ell\log(e^2/\delta)}{n}.
\end{aligned}
\]
\\[2pt]
\hline

\multicolumn{2}{@{}l}{\textbf{Population gradient}}\\*
\citet[Proposition~1]{zhu2025}: strong growth with constant $1+\kappa$, bounded gradients, mean square gradient stability, fixed $\theta>0$;
\[
\begin{aligned}
\mathbb E_\xi\|\nabla R(w_\xi)\|_2
&\lesssim(1+\theta)
\mathbb E_\xi\|\nabla R_{\mathrm{emp}}(w_\xi)\|_2\\
&\quad+\frac{1+\theta}{\theta}\log(1/\delta)\left(\frac{(1+\kappa)L_\ell}{n}+\gamma\log n\right).
\end{aligned}
\]
&
Equation~\eqref{eq:strong-growth-mean}: gradients bounded at outputs, mean square gradient stability, variance growth~\eqref{eq:gradient-norm-growth}, $\sigma_g=0$, fixed $\theta>0$;
\[
\begin{aligned}
\mathbb E_\xi\|\nabla R(w_\xi)\|_2 &\le(1+\theta)\mathbb E_\xi\|\nabla R_{\mathrm{emp}}(w_\xi)\|_2\\
&\quad+84(1+\theta)\gamma\log(e^2/\delta)\\
&\quad+\left(4(1+\theta)+\frac{\kappa(1+\theta)^2}{\theta}\right)\frac{L_\ell\log(e^2/\delta)}n.
\end{aligned}
\]
Equation~\eqref{eq:strong-growth}: gradients bounded at outputs, mean square gradient stability, variance growth~\eqref{eq:gradient-norm-growth}, $n\ge16\kappa\log(e^2/\delta)$;
\[
\begin{aligned}
(\mathbb E_\xi\|\nabla R(w_\xi)\|_2^2)^{1/2}
&\lesssim(\mathbb E_\xi\|\nabla R_{\mathrm{emp}}(w_\xi)\|_2^2)^{1/2}\\
&\quad+\gamma \log(e^2/\delta)
+\sigma_g\sqrt{\log(e^2/\delta)/n} \\
&\quad +\frac{L_\ell \log(e^2/\delta)}{n}.
\end{aligned}
\]
\\[2pt]
\hline

\multicolumn{2}{@{}l}{\textbf{Excess risk from gradient domination}}\\*
\citet[Theorem~3]{zhu2025} (unconstrained formulation): nonnegative $L$-smooth, $L_\ell$-Lipschitz losses, population PL, projection onto population minimizer set, mean square gradient stability, $n\ge32(L/\mu)^2\log(6/\delta)$;
\[
\begin{aligned}
\mathbb E_\xi[R(w_\xi)-R^\star]
&\lesssim\frac{\mathbb E_\xi\|\nabla R_{\mathrm{emp}}(w_\xi)\|_2^2}{\mu}\\
&\quad+\frac{LR^\star \log(1/\delta)}{\mu n}
+\frac{L_\ell^2\log(1/\delta)^2}{\mu n^2}\\
&\quad+\frac{\gamma^2\log^2n\,\log(1/\delta)^2}{\mu}.
\end{aligned}
\]
&
Corollary~\ref{cor:risk}: gradients bounded at outputs, mean square gradient stability, population constrained PL, variance growth $(v_0,v_1)$ at outputs, $n\ge8v_1\log(e^2/\delta)/\mu$;
\[
\begin{aligned}
\mathbb E_\xi[R(w_\xi)-R^\star]
&\lesssim\frac{\mathbb E_\xi\mathcal R_{R_{\mathrm{emp}}}(w_\xi)^2}{\mu}\\
&\quad+\frac{v_0\log(e^2/\delta)}{\mu n}
+\frac{L_\ell^2\log(e^2/\delta)^2}{\mu n^2}\\
&\quad+\frac{\gamma^2\log(e^2/\delta)^2}{\mu}.
\end{aligned}
\]
\\[2pt]
\hline
\end{longtable}
\endgroup

\subsection{Empirical minimization and gradient methods}\label{app:comparison-optimization}
In Table~\ref{tab:optimization-comparison}, we compare excess-risk bounds for exact and approximate empirical risk minimization and projected subgradient descent. We write $v_\star=\Var_Z(\nabla\ell(w^\star,Z))$.

\begingroup
\footnotesize
\setlength{\tabcolsep}{4pt}
\renewcommand{\arraystretch}{1.15}
\setlength{\LTpre}{6pt}
\setlength{\LTpost}{6pt}
\begin{longtable}{@{}>{\raggedright\arraybackslash}p{0.47\textwidth}@{\hspace{0.025\textwidth}}>{\raggedright\arraybackslash}p{0.47\textwidth}@{}}
\caption{Comparison of excess-risk bounds for exact and approximate empirical risk minimization and projected subgradient descent.}\label{tab:optimization-comparison}\\
\hline
\textbf{Prior work} & \textbf{This paper}\\
\hline
\endfirsthead
\multicolumn{2}{@{}l}{\emph{Table~\thetable, continued}}\\
\hline
\textbf{Prior work} & \textbf{This paper}\\
\hline
\endhead
\hline
\multicolumn{2}{r@{}}{\emph{Continued on next page}}\\
\endfoot
\hline
\endlastfoot
\multicolumn{2}{@{}l}{\textbf{Approximate ERM: excess risk}}\\*
\citet[Proposition~2.1]{klochkov2021}: $L_\ell$-Lipschitz, $\lambda$-strongly convex losses; optimization error at most $\epsilon_{\mathrm{opt}}$;
\[
\begin{aligned}
&R(A(S))-R^\star\\
&\lesssim\epsilon_{\mathrm{opt}}+\left(\frac{L_\ell^2}{\lambda n}
+L_\ell\sqrt{\frac{\epsilon_{\mathrm{opt}}}{\lambda}}\right)
\log n\,\log(1/\delta).
\end{aligned}
\]
&
Corollary~\ref{cor:approximate}: $L_\ell$-Lipschitz losses, empirical quadratic growth~\eqref{eq:qg-pop-emp}, $\mathcal M_s\ne\varnothing$, integrable loss at a fixed parameter, measurable empirical optimum;
\[
\begin{aligned}
R(A(S))-R^\star
&\le\frac{1310L_\ell^2\log(e^2/\delta)}{\lambda n}+L_\ell\sqrt{\frac{2\Delta_{\mathrm{opt}}(S)}{\lambda}}.
\end{aligned}
\]
\\[2pt]
\hline

\multicolumn{2}{@{}l}{\textbf{Projected subgradient descent}}\\*
\citet[Section~2.3]{klochkov2021}: $L_\ell$-Lipschitz, $\lambda$-strongly convex losses, weighted projected subgradient method;
\[
\begin{aligned}
R(\bar w_{S,T})-R^\star
&\lesssim\frac{L_\ell^2}{\lambda}
\left(\frac1n+\frac1{\sqrt T}\right)
\log n\,\log(1/\delta).
\end{aligned}
\]
&
Corollary~\ref{cor:approximate}: $L_\ell$-Lipschitz, $\lambda$-strongly convex losses, weighted projected subgradient method;
\[
R(\bar w_{S,T})-R^\star\le\frac{L_\ell^2}{\lambda}\left(\frac{1310\log(e^2/\delta)}n+\frac{2}{\sqrt{T+1}}\right).
\]
\\[2pt]
\hline

\multicolumn{2}{@{}l}{\textbf{Smooth ERM: fast rates in stochastic optimization}}\\*
\citet[Theorems~3 and~5]{zhang2017}: nonnegative $L$-smooth losses, $\lambda$-strongly convex population risk, $\kappa=L/\lambda$; $n=\tilde{\Omega}(\kappa d)$ for convex individual losses (Theorem~3), $n=\tilde{\Omega}(\kappa^2d)$ otherwise (Theorem~5); regularized generalized linear models in Hilbert space, $n=\tilde{\Omega}(\kappa^2)$ (Theorem~7);
\[
R(\widehat w)-R^\star
\lesssim\frac1{\lambda n^2}
+\frac{\kappa R^\star}{n},
\]
&
Proposition~\ref{prop:smooth-erm}: nonnegative $L$-smooth losses on a separable Hilbert space, empirical PL almost surely, $\|\nabla\ell(w^\star,z)\|_{\mathcal H}\le G$, $n\ge4L/\mu$;
\[
\begin{aligned}
R(\widehat w)-R^\star
&\lesssim\frac{Lv_\star\log(6/\delta)}{\mu^2n}+\frac{LG^2\log^2(6/\delta)}{\mu^2n^2}.
\end{aligned}
\]
\\[2pt]
\hline

\multicolumn{2}{@{}l}{\textbf{Constrained smooth ERM}}\\*
\citet[Theorem~4]{zhu2025}: false under the stated assumptions on a constrained domain; see Appendix~\ref{app:smooth-erm-counterexamples}.
&
Proposition~\ref{prop:smooth-erm-constrained}: closed convex domain in a separable Hilbert space, nonnegative $L$-smooth losses, empirical normal-cone PL almost surely, self-bounding, $0\le\ell(w^\star,Z)\le B$ almost surely, $n\ge4L/\mu$;
\[
\begin{aligned}
R(\widehat w)-R^\star
&\lesssim\frac{L^2}{\mu^2}
\left(\frac{R^\star\log(3/\delta)}{n}\right.\left.+\frac{B\log^2(3/\delta)}{n^2}\right).
\end{aligned}
\]
\\[2pt]
\end{longtable}
\endgroup

\end{document}